\documentclass[12pt]{article} 
\usepackage{graphicx} 
\usepackage{amsmath}  
\usepackage{geometry} 
\usepackage{subcaption}
\usepackage{amssymb}
\usepackage{authblk}
\usepackage{amsfonts} 
\usepackage{fancyhdr} 
\usepackage{lipsum} 
\usepackage{enumitem}         
\usepackage{xcolor} 
\usepackage{threeparttable} 
\usepackage{booktabs} 
\usepackage{amsthm} 
\theoremstyle{remark}

\newcommand{\keyword}[1]{\textbf{Keywords:} #1}
\usepackage[linesnumbered,ruled,vlined]{algorithm2e}
\usepackage[round]{natbib} 
\usepackage[bookmarks=true,bookmarksopen=true,bookmarksnumbered=true]{hyperref}
\title{A general physics-constrained  method for the modelling of equation's closure terms with sparse data}

\author[1,2]{Tian Chen}
\author[1,3]{Shengping Liu}
\author[1,3]{Li Liu\thanks{Corresponding author: Li Liu (liu\_li@iapcm.ac.cn)}} 
\author[1,3]{Heng Yong}

\affil[1]{\small Institute of Applied Physics and Computational Mathematics, Beijing}
\affil[2]{\small Graduate School of China Academy of Engineering Physics, Beijing}
\affil[3]{\small National Key Laboratory of Computational Physics}
\date{}

\begin{document}
\maketitle
\vspace{-20pt} 
\begin{abstract}
    Accurate modeling of closure terms is a critical challenge in engineering and scientific research, particularly when data is sparse (scarse or incomplete), making widely applicable models difficult to develop. This study proposes a novel approach for constructing closure models in such challenging scenarios. We introduce a Series-Parallel Multi-Network Architecture that integrates Physics-Informed Neural Networks (PINNs) to incorporate physical constraints and heterogeneous data from multiple initial and boundary conditions, while employing dedicated subnetworks to independently model unknown closure terms, enhancing generalizability across diverse problems. These closure models are integrated into an accurate Partial Differential Equation (PDE) solver, enabling robust solutions to complex predictive simulations in engineering applications. To assess the method's accuracy and robustness under sparse data conditions, we evaluate it on two representative benchmark problems: a PDE system with an unknown source term and a compressible Euler fluid dynamics system with an unknown equation of state. Results demonstrate that the proposed method constructs accurate closure models with robust generalizability, even with limited data. This study offers an innovative framework for unified closure modeling in multiscale or nonlinear systems, facilitating advancements in related engineering applications.

    
\end{abstract}
\keyword{Closure model, Sparse data, Physics-Informed Neural Networks, generalizability}

\section{Introduction}
In the formulation of governing equations for complex engineering systems, non-closure issues frequently arise due to cross-scale effects—specifically, the impact of smaller-scale phenomena on overall system behavior. A prominent example of this challenge is observed in Inertial Confinement Fusion (ICF), where simulations of high-impact dynamics rely on equations of state~\cite{Traditional_EOS_1,Traditional_EOS_2,AI_EOS_3,AI_EOS_4} and constitutive relations~\cite{Solids_1,Solids_2} inherently defined at microscopic or mesoscopic scales. Similar dependencies are evident in detonation reaction models~\cite{detonation_1,detonation_2,detonation_3} and turbulence modeling~\cite{Traditional_turbulent_1,Traditional_turbulent_2,AI_turbulent_3,AI_turbulent_4,AI_turbulent_5}, where unresolved small-scale processes critically influence macroscopic dynamics.

This persistent scale disparity results in underdetermined macroscopic variables, manifesting as the fundamental closure problem in equation systems. To address this incompleteness, closure models must effectively capture the effects of inaccessible microscale phenomena.

To tackle these challenges, closure models are known by various names across domains~\cite{Closure_name}, such as moment closure~\cite{Closure_name_moment_closure1,Closure_name_moment_closure2}, parameterization~\cite{Closure_name_Parametrization1, Closure_name_Parametrization2}, coarse graining~\cite{Closure_name_coarse_graining1,Closure_name_coarse_graining2}, and subgrid-scale (SGS) models~\cite{SGS1, SGS2}. From a unified mathematical perspective, these approaches share the common abstraction of closure processes. However, developing such models for multiscale systems remains challenging. Conventional approaches, for instance, derive semi-empirical macroscopic models by reducing microscopic equations—a process requiring deep physical insight and extensive calibration against experimental or simulation data~\cite{Traditional_turbulent_1,Traditional_turbulent_2,Traditional_EOS_1,Traditional_EOS_2}. Though less reliant on data, this method is labor-intensive and prone to oversimplification, often producing overconstrained systems with limited adaptability across scenarios.

In contrast, data-driven approaches have demonstrated transformative potential in recent years, particularly in capturing high-dimensional nonlinear relationships. Deep neural networks (DNNs), with their exceptional nonlinear mapping capabilities, have significantly advanced closure modeling research. In turbulence modeling, machine learning techniques have markedly improved closure model accuracy~\cite{AI_turbulent_3,AI_turbulent_4,AI_turbulent_5}. A key strength of neural network-based methods is their ability to learn from large datasets, enabling unified models with consistent model parameters across diverse conditions. Complementary techniques—such as dynamic mode decomposition~\cite{dong_tai_mo_fen_jie}, feature-space regression~\cite{te_zheng_hui_gui_1,te_zheng_hui_gui_2}, and operator inference~\cite{suan_zi_tui_li}—enhance the extraction of patterns from high-fidelity simulations or experiments.

Despite their strengths, data-driven approaches face significant challenges. A principal limitation is their reliance on large, high-quality datasets, which are frequently unavailable in engineering applications. In such contexts, data is often scarce due to financial constraints or limitations in measurement techniques. Furthermore, integrating physical constraints into these models poses a substantial difficulty, commonly leading to breaches of foundational principles such as conservation laws or symmetries that underpin traditional modeling frameworks.

To address these deficiencies, Physics-Informed Neural Networks (PINNs)~\cite{origin_PINN_zheng_fan_problem} have been introduced as an innovative solution. PINNs incorporate partial differential equations (PDEs) and boundary conditions directly into the neural network training process, establishing a hybrid methodology that merges data-driven learning with physically grounded principles. This physics-informed approach diminishes the reliance on extensive datasets and supports the resolution of inverse problems, including the estimation of parameters in governing equations~\cite{PINNs_fan_problem1, PINNs_fan_problem2}. As a result, PINNs have proven effective across diverse disciplines, such as fluid dynamics, materials science, and geophysics~\cite{PINN_ying_yong_ling_yu}.

Nevertheless, PINNs exhibit a notable limitation: their generalization capability is often constrained to specific scenarios. This case-dependent performance restricts the development of broadly applicable closure models, presenting an ongoing challenge for their wider adoption.

Building on these foundations, we propose a two-step methodology comprising a \textbf{Constructor} and an \textbf{Applicator}. In the first stage, the Constructor employs a novel \textbf{Series-Parallel Multi-Network Architecture} that integrates data from multiple cases and incorporates known physical constraints while preserving the independence of closure terms. This design enhances the model's generalizability across diverse problems, extending even beyond the dimensionality of the training cases. In the subsequent stage, the Applicator embeds the trained closure model into traditional numerical frameworks. This synergistic approach leverages the nonlinear modeling capabilities of deep neural networks (DNNs) while retaining the accuracy and convergence robustness of established numerical methods, which remain indispensable for solving forward problems. To validate this framework, we applied it to two distinct equation systems. The numerical results confirm the method's generalizability across different problem types and its precision in capturing multiscale interactions, thereby establishing a promising pathway for the development of closure models.

The structure and strategy of this work are as follows. In Section 2, we present an abstract mathematical description of the closure term model. In Section 3, we explain the working principles and theoretical foundations of Physics-Informed Neural Networks (PINNs) and traditional numerical methods for solving partial differential equations (PDEs). In Section 4, we elaborate on the core methods and technical details involved in constructing the closure surrogate model using the new framework. In Section 5, we introduce two distinct equation systems and conduct numerical experiments to validate the effectiveness and practicality of the proposed method. Finally, conclusions are drawn in Section 6. Additionally, Appendices A, B, and C provide detailed information on the experimental data and supplementary materials, enabling readers to gain a deeper understanding of the research context and implementation specifics.

\section{Mathematical Problem Statement\label{sec2}}
As emphasized earlier, closure-term problems are pervasive in engineering applications. In this study, we formulate the problem using the following general equation system:
\begin{subequations}
\begin{align}
&\text{The Known term } (\mathbb{K}): \; F(t,\mathbf{X},\mathbf{U}, \partial_{t}{\mathbf{U}},\nabla \mathbf{U}, \ldots)=0, \label{eq:Kterm}\\
&\text{The Unknown term } (\mathbb{U}): \; G(\mathbf{U_1},\mathbf{U_2},\partial_\mathbf{U_1}{\mathbf{U_2}} , \ldots)=0,  \label{eq:Uterm}
\end{align}
\label{eq:ProblemStatement}
\end{subequations}

Here, $\mathbf{U}(t, \mathbf{X}) = (\mathbf{U_1}, \mathbf{U_2}) = (U_1, U_2, \dots, U_n)$ denotes an $n$-dimensional vector defined over a $d$-dimensional spatial domain, with $\mathbf{U_1}$ and $\mathbf{U_2}$ representing two disjoint subsets of $\mathbf{U}$. The known term, denoted $\mathbb{K}$, encapsulates the well-established components of the system. This term incorporates various mathematical constructs, such as time derivatives $\partial_t \mathbf{U}$, spatial gradients $\nabla \mathbf{U}$, and potentially complex nonlinear relationships among these derivatives. These relationships typically reflect underlying physical or mathematical principles, such as conservation laws or constitutive relations, governing the system's behavior.

In contrast, the unknown term, $\mathbb{U}$, represents the component requiring closure modeling. It generally comprises unknown functions of the vector $\mathbf{U}$ and its internal derivatives (e.g., $\partial_{\mathbf{U_1}} \mathbf{U_2}$, denoting partial derivatives of $\mathbf{U_2}$ with respect to components of $\mathbf{U_1}$). The primary challenge in modeling this term stems from its inherent complexity, which may arise from nonlinear interactions or  mathematical and statistical relationships of smaller-scale physics.

The modeling process is constrained by limited additional information, including:
\begin{itemize}
    \item $n$ distinct cases, each accompanied by corresponding initial and boundary conditions:
        \begin{equation}
            \mathbf{U^i}(t=0, \mathbf{X}) = \mathbf{U^i_0}(\mathbf{X}), \quad \mathbf{X} \in \Omega,
        \end{equation}
        \begin{equation}
            \mathbf{U^i}(t, \mathbf{X}) = \mathbf{g^i}(t, \mathbf{X}), \quad (t, \mathbf{X}) \in [0, T] \times \partial \Omega,
        \end{equation}
        where $\mathbf{U^i_0}$ denotes the initial condition and $\mathbf{g^i}$ represents the boundary condition for the $i$-th case.
    \item Data derived from these cases, expressed as:
        \begin{equation}
            \mathbf{U^i}(t, \mathbf{X}) = \mathbf{d^i}(t, \mathbf{X}),
        \end{equation}
        where $\mathbf{d^i}$ denotes the observed system data for the $i$-th case.
\end{itemize}

However, the available data is subject to the following limitations that can be collectively referred to as sparse data problems:
\begin{itemize}
    \item \textbf{Scarce Sample Cases}: Due to experimental costs and engineering constraints, the number of cases, $n$, is often severely limited.
    \item \textbf{Incomplete Data}: Data incompleteness arises when certain components of $\mathbf{U}$ (e.g., $\mathbf{U_2}$) are more difficult to measure than others (e.g., $\mathbf{U_1}$), resulting in datasets with scarce or missing $\mathbf{U_2}$ values.
    \item \textbf{Selection Bias}: The dataset is predominantly collected from simplified, low-dimensional problem configurations, whereas the intended applications involve complex, high-dimensional real-world scenarios with intricate interactions.
\end{itemize}

Consequently, this scenario presents an extreme modeling challenge, necessitating the development of a model with robust generalization capabilities despite sparse data—a task at which traditional modeling approaches often struggle. To address this, we must maximize the utilization of information embedded within the existing equations.

\section{Methods For Solving PDEs}
The newly developed method in Section~\ref{sec:Method} will combine PINNs and conventional high-precision numerical methods. Below, we introduce the PINNs-WE method, a modified PINNs approach designed to capture discontinuities~\cite{PINNs_loss_using}, and the high-accuracy shock-capturing finite difference approach~\cite{WENO_1,WENO_2} for solving Euler equations.

\subsection{PINNs-WE Method\label{PINNs-WE}}
Physics-Informed Neural Networks (PINNs)~\cite{origin_PINN_zheng_fan_problem,PINNs_zheng_problem1,PINNs_zheng_problem2,PINNs_zheng_problem3,PINNs_zheng_problem4} represent a novel paradigm for solving PDEs, particularly excelling in inverse problems with unknown parameters~\cite{origin_PINN_zheng_fan_problem,PINNs_fan_problem1,PINNs_fan_problem2,PINNs_fan_problem3}. However, challenges persist in handling discontinuities, resolving complex fields, ensuring theoretical convergence, and achieving numerical accuracy, as exemplified by Eq.~(\ref{eq:Compressible_Euler_control_equation}) in this work. The weighted equation physics-informed neural networks (PINNs-WE) method~\cite{PINNs_loss_using} is an enhanced PINNs approach designed to improve discontinuity capture in Euler equation systems.

\begin{figure}[!htb]
  \centering  \includegraphics[height=7.5cm]{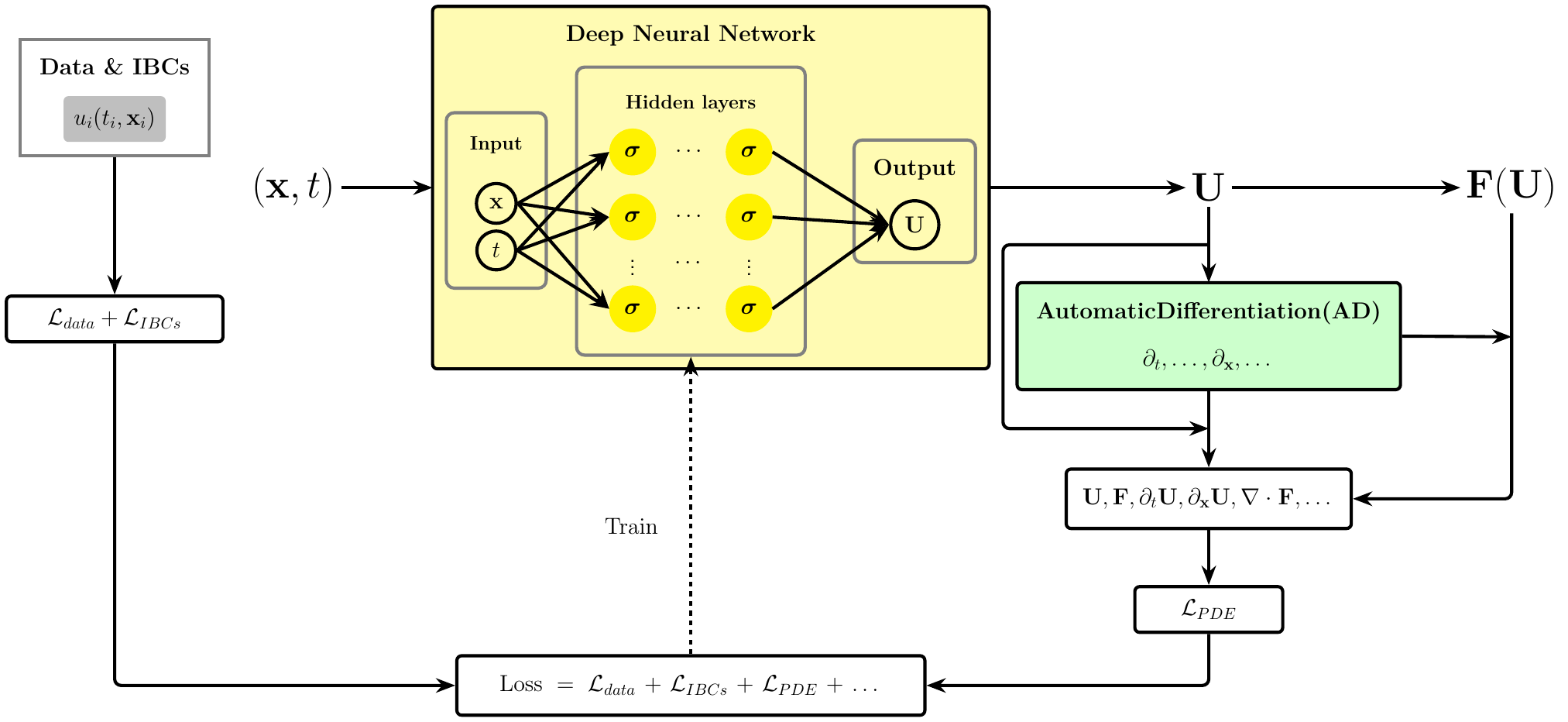}
  \renewcommand{\figurename}{Fig.}
  \caption{The Structure Diagram of PINNs}
  \label{fig:The structure diagram of PINN}
\end{figure}

As illustrated in Figure~\ref{fig:The structure diagram of PINN}, solving PDEs with PINNs typically involves two components. The first is a neural network $\hat{\mathbf{U}}(t,\mathbf{x};\theta)$ that approximates the relationship between $\mathbf{U}(t,\mathbf{x})$ and trainable parameters $\theta$. The second incorporates the governing equations, initial conditions, and boundary conditions to train the network. Nonlinear operators such as $\partial / \partial t$ and $\nabla$ are computed via {\textbf{automatic differentiation}}~\cite{Automatic_Automatic}.

The loss function of PINNs, used to train the neural network $\hat{\mathbf{U}}(t,\mathbf{x};\theta)$, comprises at least two terms: one reflecting the network's adherence to the PDEs, and another enforcing initial/boundary conditions (IBCs) and data constraints:
\begin{equation}
\begin{split}
\mathcal{L} = {} & 
\omega_{PDE} \mathcal{L}_{PDE} + \omega_{IBCs} \mathcal{L}_{IBCs} + \omega_{data} \mathcal{L}_{data}
\end{split}
\label{eq:Definition of the original PINN loss function}
\end{equation}
To define the loss, we select residual points within the domain $\Omega$ and on its boundary $\partial\Omega$, denoted as $S_{PDE}$, $S_{IBCs}$, and $S_{data}$, respectively:
\begin{subequations}
\begin{align}
\mathcal{L}_{PDE} = {} & 
\frac{1}{|S_{\text{PDE}}|} \sum_{t_i,\mathbf{x}_i\in S_{\text{PDE}}}|\mathbf{G}_i|^2 \\
\mathcal{L}_{IBCs} = {} &
\frac{1}{|S_{\text{IBCs}}|} \sum_{\mathbf{x}_i\in S_{\text{IBCs}}}| \hat{\mathbf{U}}_0(\mathbf{x}_i;\theta) - \mathbf{U}_0(\mathbf{x}_i) |^2 \\
\mathcal{L}_{data} = {} & 
\frac{1}{|S_{\text{data}}|} \sum_{t_i, \mathbf{x}_i\in S_{\text{data}}}| \hat{\mathbf{U}}(t_i,\mathbf{x}_i;\theta) - \mathbf{U}(t_i,\mathbf{x}_i) |^2 \\
\end{align}
\label{eq:The specific definition of the original PINN loss function}
\end{subequations}
Here, 
$$ \mathbf{G}_i := \lambda(\partial_t\hat{\mathbf{U}}(t_i,\mathbf{x}_i) - \nabla\cdot\mathbf{F}(\hat{\mathbf{U}}_i))$$
$$ \lambda = \frac{1}{k(|\nabla \cdot \vec{u}| - \nabla \cdot \vec{u}) + 1},\quad k = 0.2 $$
in PINNs-WE, and
$$ \mathbf{G}_i = 0 $$
 represents the PDEs that the network must satisfy at residual points \((t_i, \mathbf{x}_i) \in S_{\text{PDE}}\). Additionally, \(\mathbf{U_0}\) denotes the initial and boundary conditions that \(\hat{\mathbf{U}}(t_i, \mathbf{x}_i; \theta)\) must satisfy at points \((t_i, \mathbf{x}_i) \in S_{\text{IBCs}}\). The parameter \(\omega\) adjusts the strength of loss constraints for different terms~\cite{weight_w1, weight_w2, weight_w3}, typically assigning higher weights to initial and boundary points. In PINNs-WE, two additional loss functions, \(\mathcal{L}_{RH}\) and \(\mathcal{L}_{CONs}\), are also employed, with specific configurations detailed in Appendix A.

\subsection{High-order accuracy shock-capturing finite difference schemes}
Equation~(\ref{eq:Compressible_Euler_control_equation}) considered in this paper belongs to the class of hyperbolic conservation laws, which typically require high-accuracy shock-capturing schemes to compute numerical solutions in Applicator part of the new method. In this section, we briefly introduce these schemes, focusing on the one-dimensional scalar case, i.e., Eq.~(\ref{eq:one dimensional scalar case}):
\begin{equation}
\begin{split}
\frac{\partial{u}}{\partial{t}} + \frac{\partial{f}}{\partial{x}} = {} & 
0
\end{split}
\label{eq:one dimensional scalar case}
\end{equation}

Consider a uniform grid defined by points $x_i = i\Delta x, \quad i = 0, \dots, N$, also termed cell centers, where $\Delta x$ is the uniform grid spacing. The semi-discretized form of Eq.~(\ref{eq:Compressible_Euler_control_equation}), obtained via the method of lines, yields a system of ordinary differential equations:
\begin{equation}
\begin{split}
\frac{du_i(t)}{dt} = {} & 
-\frac{\partial f}{\partial x}|_{x=x_i}, \quad i = 0, \dots, N
\end{split}
\label{eq:The semi-discretized form of Euler equation}
\end{equation}
where $u_i(t)$ is a numerical approximation to the point value $u(x_i, t)$. This can be approximated by a conservative finite difference formula:
\begin{equation}
\begin{split}
\frac{du_i(t)}{dt} = {} & 
-\frac{\hat{f}_{i+\frac{1}{2}} - \hat{f}_{i-\frac{1}{2}}}{\Delta x}
\end{split}
\label{eq:finite difference formula for Euler}
\end{equation}
where $\hat{f}_{i\pm\frac{1}{2}}$ denotes the numerical flux.

Here, we employ spatially discretized WENO schemes~\cite{WENO_Z, WENO_Z_for_p, WENO_1, WENO_2, WENO_3}, a high-resolution numerical method tailored for fluid dynamics problems involving shocks and discontinuities. For temporal discretization, we use third-order Runge-Kutta (R-K) schemes~\cite{RK}, an explicit time integration method offering high accuracy and robust stability.

\section{Methodology\label{sec:Method}}
PINNs offer the advantage of integrating equation relationships with data; however, they lack generalization across different cases. To address the general problem outlined in Eq.~\ref{eq:ProblemStatement}, we propose a novel DNN-based closure modeling method that fully leverages the flexibility of PINNs while seamlessly incorporating the strengths of traditional high-order schemes for solving complex forward problems. Compared to standard PINNs, this new approach achieves models with cross-problem generalization capabilities using only limited, sparse data alongside partial known equation information.

\begin{figure}[htbp]
    \centering
    \includegraphics[width=\textwidth]{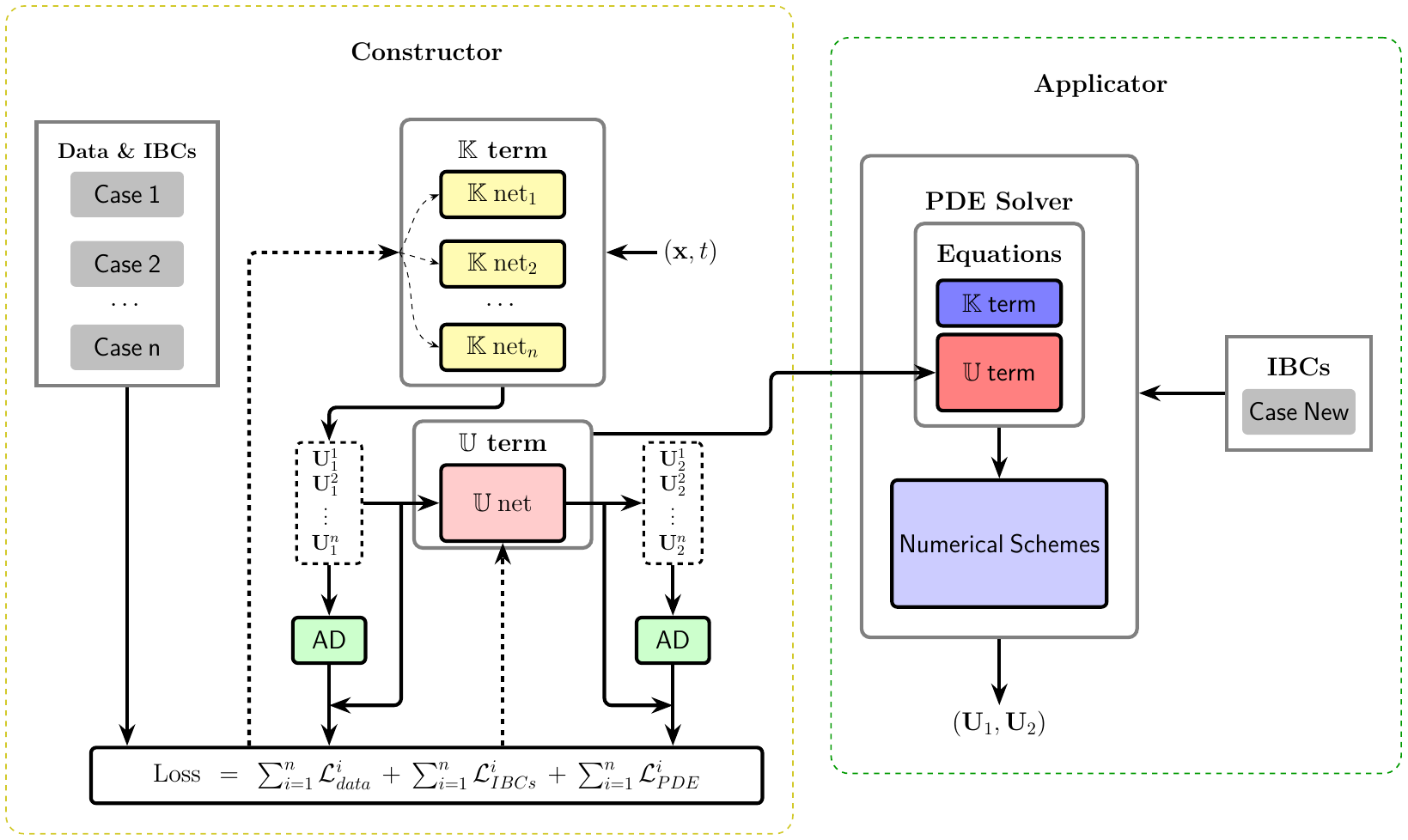}
    \caption{Network Framework Diagram of New Method}
    \label{fig:surrogate model of the closure term}
\end{figure}

The overall workflow of the method is illustrated in Fig.~\ref{fig:surrogate model of the closure term} and comprises two main components: the model \textbf{Constructor} and the model \textbf{Applicator}. In the constructor, we embed the known terms of the equation into the PINNs loss function to effectively constrain the learning process. For each specific problem, we design an independent network architecture to handle the known terms, overcoming the generalization limitations of PINNs. For the unknown terms, we construct a universal DNN as a shared module, ensuring its independence from specific problems. In the applicator, recognizing that PINNs do not offer significant advantages over traditional numerical methods for complex forward problems, we integrate the developed DNN model into conventional high-accuracy numerical solvers, coupling algorithms to address new, complex challenges. For the compressible fluid problems considered in this paper, the traditional method employs the WENO-Z finite difference numerical scheme~\cite{WENO_Z} combined with the third-order Runge-Kutta (R-K) time advancement method~\cite{RK}.

\subsection{Constructor}
\begin{figure}[htbp]
    \centering
    \includegraphics[width=\textwidth]{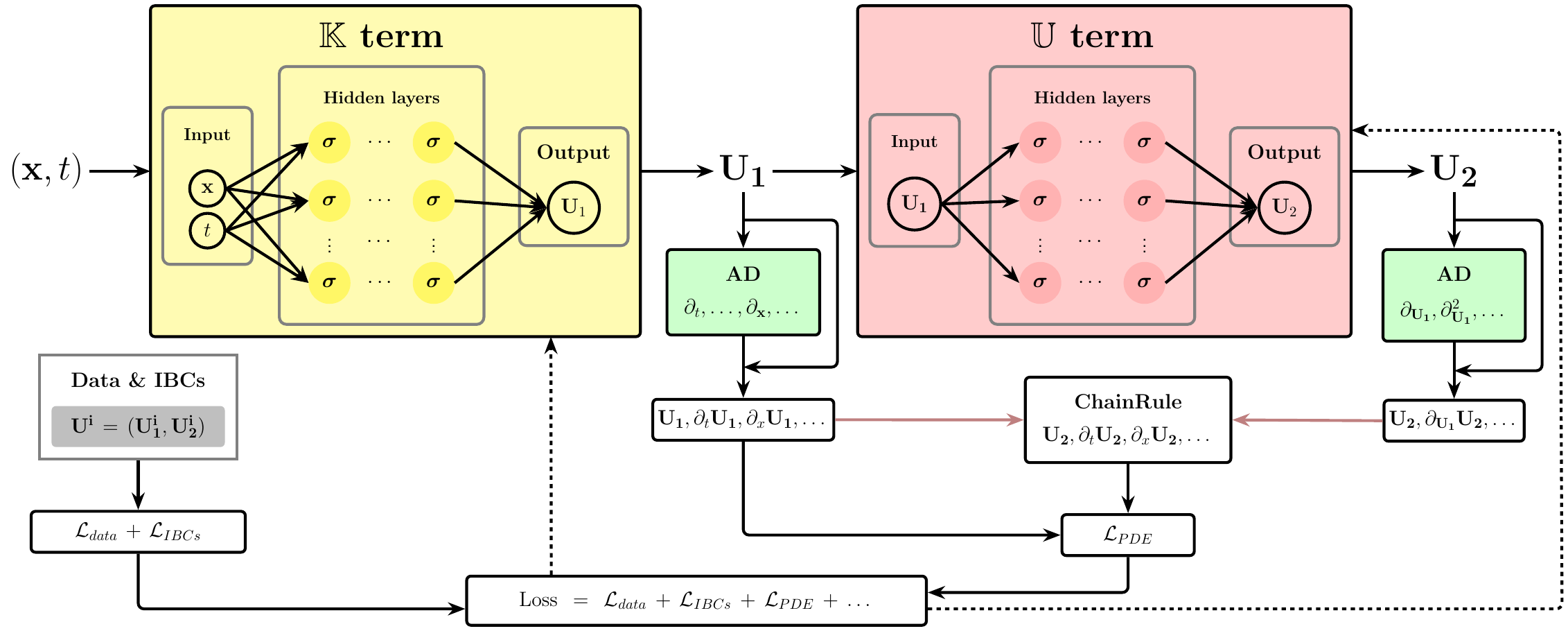}
    \caption{Network Framework Diagram of Single Model}
    \label{fig:Single model of the closure term}
\end{figure}

In the construction phase of the closure model, we adopt a \textbf{series-parallel multi-network architecture}, where the \(\mathbb{K}\) and \(\mathbb{U}\) terms are modeled independently in the series part, and the \(\mathbb{K}\) term is further decomposed into a parallel multi-network based on different cases. Below, we explain the rationale behind this architectural design:
\begin{enumerate}
    \item The \textbf{series structure} is designed to model the unknown equations independently, preserving their case-independent generalization properties while enabling the collective computation of output quantities to form the final loss function. Figure~\ref{fig:Single model of the closure term} illustrates the detailed internal working mechanism for a single case. As shown, in terms of PDE differentiation, the two networks transmit derivative information via the chain rule, collaboratively constructing the final residual loss.

    \item The \textbf{parallel structure} leverages the strengths of PINNs, maintaining their independence across different cases. This enables the integration of governing equations and data from multiple cases to collectively support the modeling of the \(\mathbb{U}\)-network. The single-case configuration depicted in Fig.~\ref{fig:Single model of the closure term} alone is insufficient to achieve this.
\end{enumerate}

\subsection{Applicator}
It is well-known that machine learning PDE methods face significant challenges in handling complex forward problems especially with fine or high-frequency structures~\cite{PINN_challenge_for_Euler,PINNs_loss_using}. 
In these complex problems, for example, problems within compressible Euler equations system, traditional methods, such as the finite volume or finite difference methods, remain indispensable in terms of accuracy, efficiency, and convergence.
To enable the integration of the newly developed closure model into traditional code, we propose an algorithmic framework based on finite difference methods for handling Euler equations as an example, as detailed in Algorithm~\ref{algorithm:The coupled algorithm procedure for solving 2D Euler equations using finite differences method}. 

\begin{algorithm}[H]
    \SetAlgoLined
    \textbf{Init}: $t_0=0$, $\mathbf{U}_0$: $\rho_0$, $u_0$, $v_0$, $e_0$, $p_0$\;    
    \textbf{Boundary}: Periodic boundaries, extrapolation boundaries, etc.\;
    \While{$t_k < T$}{
        $t_k \leftarrow t_k + \Delta t$\;
        \textbf{Compute Flux1}:\quad $\mathbf{F}_1 = \mathbf{F}_1^+ + \mathbf{F}_1^- \leftarrow$ Using Global L-F flux splitting\;
        \textbf{Reconstruct $\mathbf{F}_1^+$ and $\mathbf{F}_1^-$}:\quad Using WENO schemes\;
        \textbf{Compute Flux2}:\quad $\mathbf{F}_2 = \mathbf{F}_2^+ + \mathbf{F}_2^- \leftarrow$ Using Global L-F flux splitting\;
        \textbf{Reconstruct $\mathbf{F}_2^+$ and $\mathbf{F}_2^-$}:\quad Using WENO schemes\;
        \textbf{Compute $\mathbf{U}_{t_{k+1}}$}:\quad Using the third-order RK method\;
        \textbf{Compute physical quantities at $t_k$}:\quad $\rho_{t_k}$, $u_{t_k}$, $v_{t_k}$, $e_{t_k}$, {$p_{t_k} \leftarrow$ Using the closure model}\;
        \textbf{Boundary}: Periodic boundaries, extrapolation boundaries, etc.\;
    }
    \caption{The coupling algorithm procedure for solving 2D hyperbolic equations using the finite difference method}
    \label{algorithm:The coupled algorithm procedure for solving 2D Euler equations using finite differences method}
\end{algorithm}

\section{Applications and Results\label{Numerical_Examples}}
Here, we demonstrate the effectiveness of the proposed method by applying it to two distinct equation systems.

\subsection{A Simple PDE System with Unknown Source Term\label{A Simple PDE System}}
\subsubsection{Problem Description}
We first consider a simple equation involving the modeling of an unknown source term:
\begin{equation}
  \frac{\partial u}{\partial t} + s(u) = 0,
\end{equation}
where $s(u)$ is an unknown source term. This type of modeling problem is commonly encountered in reaction problems. It can be expressed in the form of Eq.~(\ref{eq:ProblemStatement}):
\begin{subequations}
\begin{align}
&\mathbb{K}:\;\quad \frac{\partial u_1}{\partial t} + u_2 = 0, \label{eq:Kterm_for_simple_model}\\
&\mathbb{U}:\;\quad u_2 = f(u_1), \label{eq:Uterm_for_simple_model}\\
& u_1(t_0)=u_0,
\end{align}
\label{eq:ProblemStatementForSimpleModel}
\end{subequations}

Here, we can only obtain data for $u_1$, and when the data is sufficiently dense, constructing an accurate model for the term $\mathbb{U}$ is straightforward. However, when the data is sparse, it becomes a challenging problem.

\subsubsection{Model Construction}
We assume the target model is defined by:
\begin{equation}
\begin{aligned}
u_2(u_1) &= u_1\sqrt{1-u_1^2}
\end{aligned}
\end{equation}
Then, the expression for $u_1$ can be calculated through the equation(\ref{eq:ProblemStatementForSimpleModel}):
\begin{equation}
\begin{aligned}
u_1(t) &= \frac{2c_0e^t}{c_0^2e^{2t}+1}
\end{aligned}
\end{equation}

Here, we constructed closure models for $u_1$ and $u_2$ using three examples with $c_0 = \frac{1}{2}, 1, 2$. For $c_0 = \frac{1}{2}$, $t \in [2, 5]$, and the networks consist of $Net_1$ with $5$ hidden layers($15$ neurons each), with $31$ uniformly sampled points of $u_1$ provided. For $c_0 = 1$, $t \in [\frac{1}{2}, 3]$, the same network architectures are used, but $51$ uniformly sampled points of $u_1$ are provided. For $c_0 = 2$, $t \in [1, 4]$, the network configurations remain unchanged, with $31$ uniformly sampled points of $u_1$ provided. 

The $Net_2$ shared among the three cases uses a network with $3$ hidden layers, each containing $15$ neurons. The generalized $L^2$ error is $0.61\%$, as shown in Table \ref{table:the $L^2$ error of toy model}. Since the data for $u_2$ is unknown, it is not possible to establish a model using a data-driven approach. Therefore, no comparative experiments are provided here.
\begin{table}[!htb]
    \centering
     \renewcommand{\tablename}{Table}
    \caption{The $L^2$ error of the closure model for the ideal equation of state}
      \begin{threeparttable}
      \begin{tabular}{cc}
          \toprule
           &Generalization error of the new method \\
          \midrule
          $L^2$ error & 0.61\% \\
          \bottomrule
      \end{tabular}
    \end{threeparttable}
\label{table:the $L^2$ error of toy model}
\end{table}

\subsubsection{Numerical Results}
\begin{figure}[!htb]
  \centering
  \includegraphics[height=6cm]{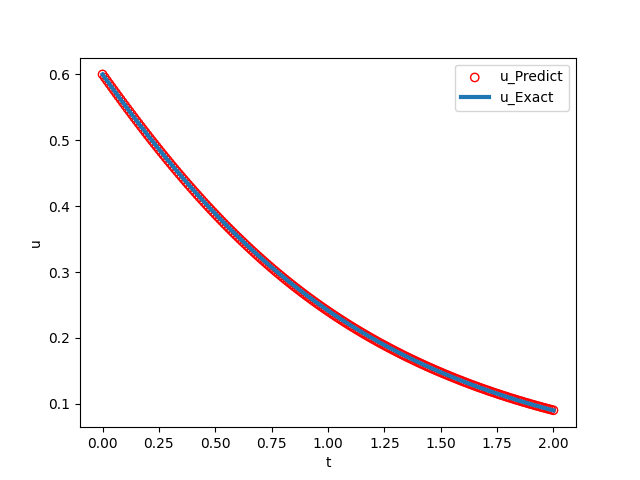}  
  \includegraphics[height=6cm]{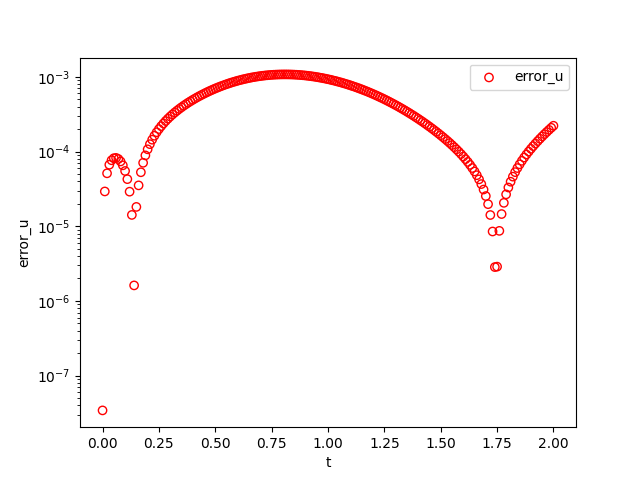}
  \renewcommand{\figurename}{Fig.}
  \caption{The result images of the test case in the simple model for the closure of $u_2(u_1)$}
  \label{fig:test_for_toy_model}
\end{figure}

To test the performance of the model, we use the established model to calculate a new case:
\begin{equation}
\begin{aligned}
u_1(t) &= \frac{6e^t}{9e^{2t}+1}
\end{aligned}
\end{equation}
Here, $t \in [0, 2]$, and the value of $u_1(t)$ is given at $t = 0$. The input of $Net_1$ is $t$, and its output is $\hat{u_1}$, with a network structure consisting of $5$ hidden layers, each containing $10$ neurons. The closure model $Net_2$ is used to output $u_2$ in the equation. The result can be shown in Fig.\ref{fig:test_for_toy_model}. From the plot of absolute error in Fig.\ref{fig:test_for_toy_model}, it can be seen that the error is generally controlled within $10^{-3}$. This result fully demonstrates that the closure model constructed using the method proposed in this paper exhibits excellent generalization performance, validating the effectiveness and potential value of the proposed approach. It sets a solid foundation for future research and practical applications.

\subsection{Compressible Euler Equations with Unknown EOS}
\subsubsection{Problem Description}
Next, we consider the compressible Euler equations with the modeling of the equation of state (EOS):
\begin{subequations}
\begin{align}
&\mathbb{K}:\;\quad \partial_{t}{\mathbf{U}}+\nabla \cdot \mathbf{F}(\mathbf{U})=0, \label{eq:Kterm_for_Euler_model}\\
&\mathbb{U}:\;\quad p=p(\rho,e), \\
& {\rm ICs}: \quad \mathbf{U}(t=0,\mathbf{X})=\mathbf{U_0}(\mathbf{X}), \quad \mathbf{X} \in \Omega,\\
& {\rm BCs}: \quad \mathbf{U}(t,\mathbf{X})=\mathbf{g}(\mathbf{X}),\quad \mathbf{X} \in \partial \Omega,
\label{eq:Uterm_for_Euler_model}
\end{align}
\label{eq:Compressible_Euler_control_equation}
\end{subequations}
For the 1D case:
\begin{equation}
\begin{alignedat}{4}
\mathbf{U} &= \left[ \begin{array}{c} \rho \\ \rho u  \\ E \end{array} \right],\quad & \mathbf{F} &= \left[ \begin{array}{c} \rho u \\ \rho u^2 + p \\  u(E+p) \end{array}\right]
\end{alignedat}
\label{eq:1D_Euler_control_equation}
\end{equation}
And for the 2D case, $\mathbf{F} = (\mathbf{F_1}, \mathbf{F_2})$, expressed as:
\begin{equation}
\begin{alignedat}{4}
\mathbf{U} &= \left[ \begin{array}{c} \rho \\ \rho u \\ \rho v \\ E \end{array} \right],\quad & \mathbf{F_1} &= \left[ \begin{array}{c} \rho u \\ \rho u^2 + p \\ \rho uv \\ u(E+p) \end{array} \right],\quad & \mathbf{F_2} &= \left[ \begin{array}{c} \rho v\\ \rho uv \\ \rho v^2 + p \\  v(E+p) \end{array} \right]
\end{alignedat}
\label{eq:2D Euler control equation}
\end{equation}
Here, $E=\rho e+\frac{1}{2}\rho(u^2+v^2)$ represents the total energy.

The modeling of EOS constitutes a broad field of study. For instance, under extreme conditions—such as those in ICF and weapons physics—complex challenges emerge, including single-medium EOS modeling or mixture EOS modeling for multi-medium scenarios. Here, we focus solely on modeling the relationship $p(\rho, e)$, where $p$ denotes pressure, $\rho$ denotes density, and $e$ represents specific internal energy, corresponding to $\mathbf{U_2}$ in Eq.~(\ref{eq:ProblemStatement}). In the 1D case, per Eq.~(\ref{eq:ProblemStatement}), $\mathbf{U_1}$ comprises $(\rho, u, e)$.

\subsubsection{Model Construction}
Here, we select the Noble-Abel equation of state (N-A) shown below~\cite{surrogate_eos2} as a test case to evaluate the performance of the method in handling more complex physical scenarios. A simpler equation of state can be found in Appendix B.
\begin{equation}
p=\frac{e\rho (\gamma -1)}{(1.0-b\rho)}, \quad \gamma = 1.4, \quad b=0.075
\end{equation}
Here, we used five groups of sparse experimental data with the same equation of state to train the closure model of the equation of state. In the jointly constructed $Net_2$, we employed a fully connected neural network with $3$ hidden layer, where each hidden layer contained $20$ neurons. To effectively prevent overfitting, an $L^2$ regularization term \cite{zheng_ze_hua} was added during the training process to constrain the network weights, thereby enhancing the model's generalization ability. The design of the five independent $Net_1$ networks and the detailed conditions of the five data groups are presented in Appendix C.

As shown in Table \ref{table:the $L^2$ error of eos2}, the $L^2$ generalization error of the closure model for the equation of state obtained by combining these five training datasets is $0.27\%$. In contrast, under the same data conditions, the $L^2$ generalization error of the closure model obtained using a data-driven method is $1.46\%$. This result demonstrates that the method proposed in this paper can still maintain high accuracy and generalization ability even in more complex scenarios, showcasing significant advantages.
\begin{table}[!htb]
    \centering
     \renewcommand{\tablename}{Table}
    \caption{The $L^2$ error of the closure model for the N-A equation of state}
      \begin{threeparttable}
      \begin{tabular}{ccc}
          \toprule
           &Generalization error of the new method&Generalization error of data-driven \\
          \midrule
          $L^2$ error & 0.27\%&1.46\% \\
          \bottomrule
      \end{tabular}
    \end{threeparttable}
\label{table:the $L^2$ error of eos2}
\end{table}

Similarly, the constructed closure model of the equation of state is combined with traditional numerical methods to solve flow field problems with corresponding initial and boundary value conditions. By comparing the obtained results with the reference solution and the numerical solution computed by the target model under the same experimental settings, the feasibility and accuracy of the closure model in practical applications are evaluated.

\subsubsection{Numerical Results}
We evaluate the model through four different test scenarios.

\paragraph{Test Case1:}
The first example is 1D smooth periodic case, with the initial conditions given as follows:
\begin{equation}
\left\{
\begin{aligned}
u(x,0)  &= 0.3\sin(\frac{2\pi x}{L}),\quad L=5,\\
\rho(x,0) &= (1+\frac{(\gamma-1)u(x,0)}{2c})^\frac{2}{\gamma-1},\quad c=\frac{\sqrt{\gamma}}{\epsilon}, \gamma=2, \epsilon=1.0,\\
p(x,0)&= \rho(x,0)^\gamma
\end{aligned}
\right.
\end{equation}
The computational domain is $\Omega = [-2.5, 2.5]$, and the final time is $T = 0.3$. The number of grid points is chosen as $N = 1000$, and the fifth-order WENO-Z scheme is used for computation. The reference solution is obtained using a fifth-order WENO-Z spatial discretization and third-order Runge-Kutta (RK) time integration with $N = 25000$ grid points.

\begin{figure}[!htb]
  \centering
  \subfloat[Comparison of the numerical solution and absolute error of density obtained by two models at $t = 0.3$]{
      \centering  
      \includegraphics[height=6.5cm]{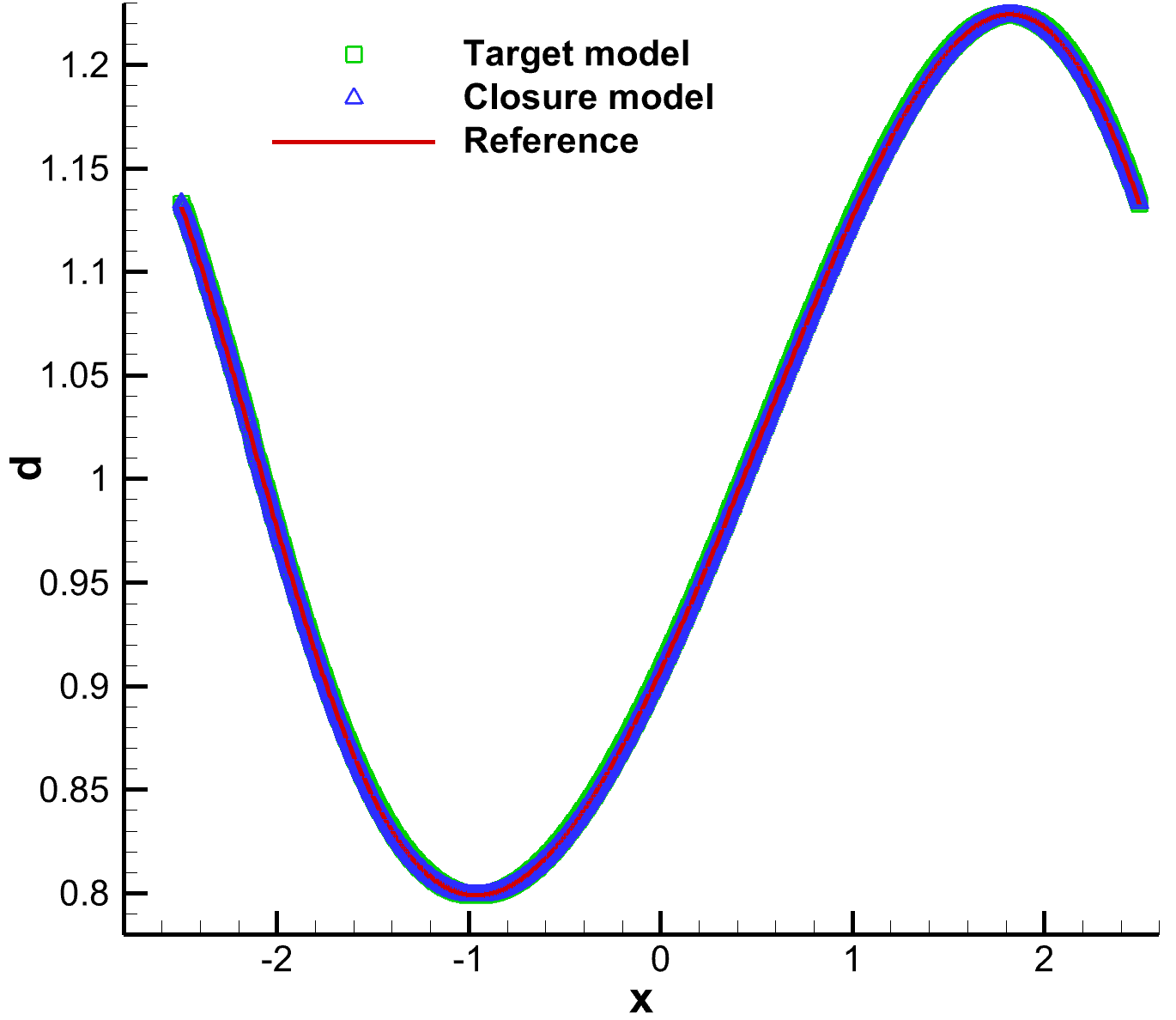}
      \label{fig:the numerical solutions of density} 
      \includegraphics[height=6.5cm]{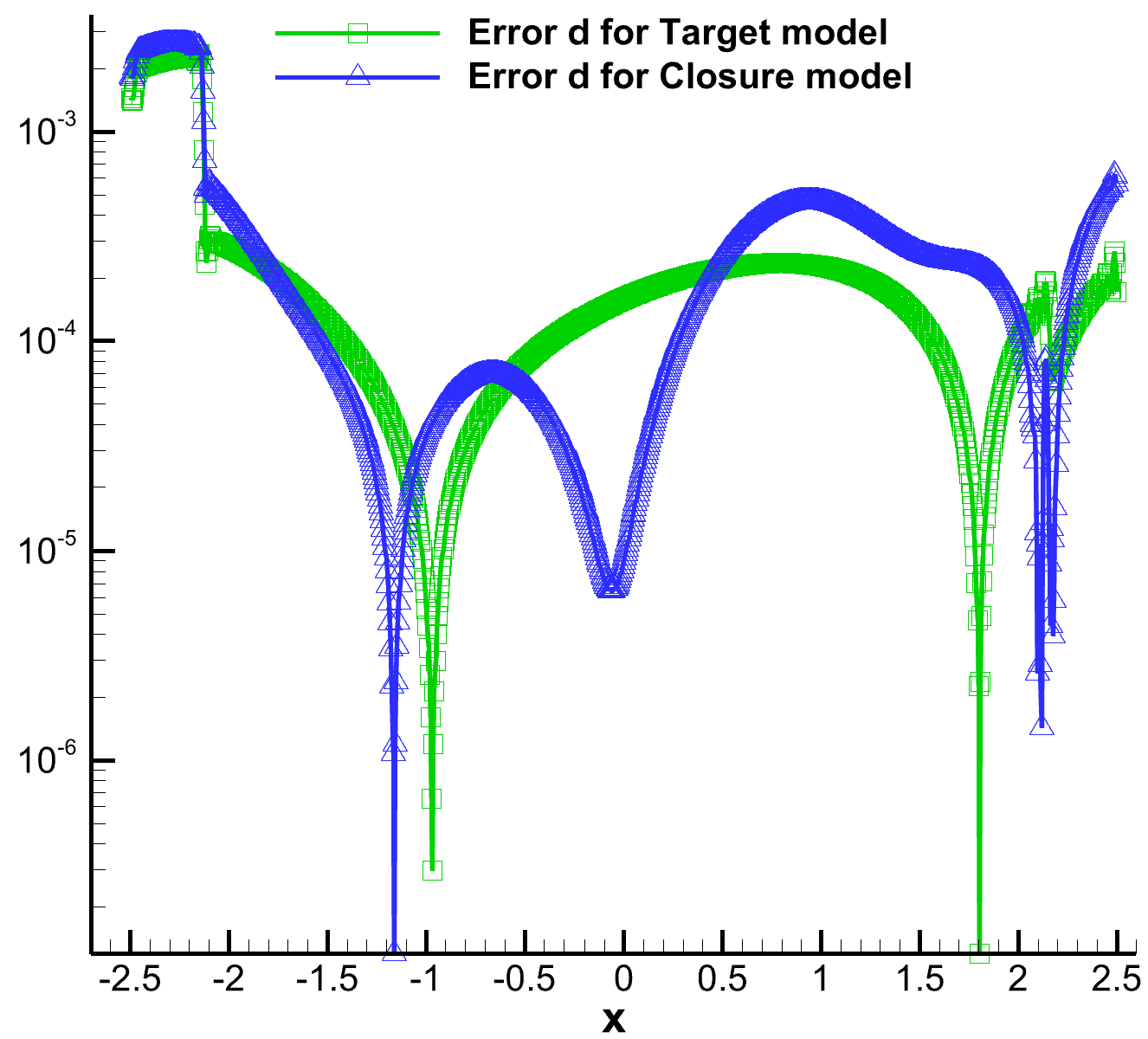}
      \label{fig:the absolute error of density}
  }
  \quad
  \subfloat[Comparison of the numerical solution and absolute error of velocity obtained by two models at $t = 0.3$]{
      \centering
       \includegraphics[height=6.5cm]{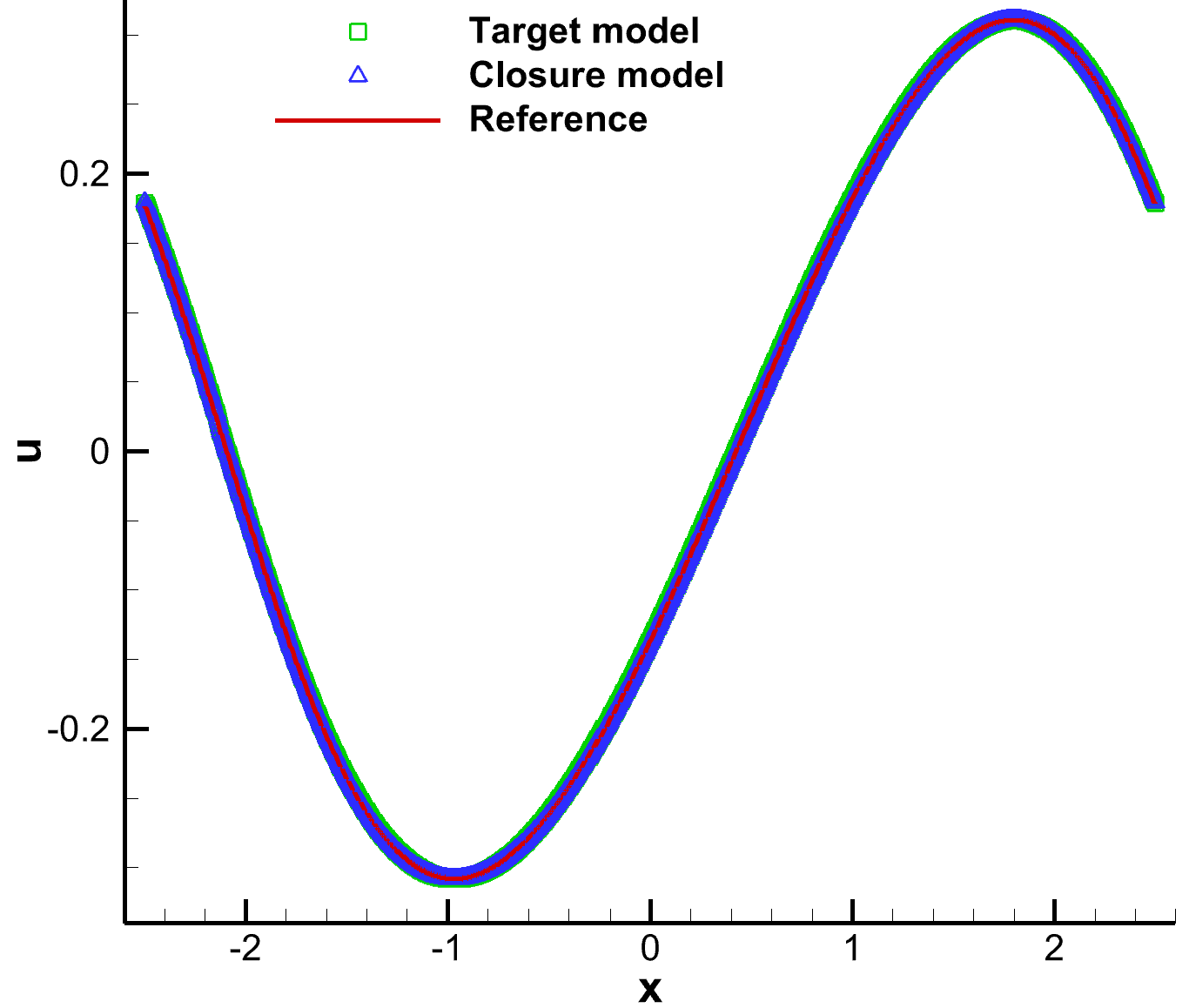}
      \label{fig:the numerical solutions of velocity}
      \includegraphics[height=6.5cm]{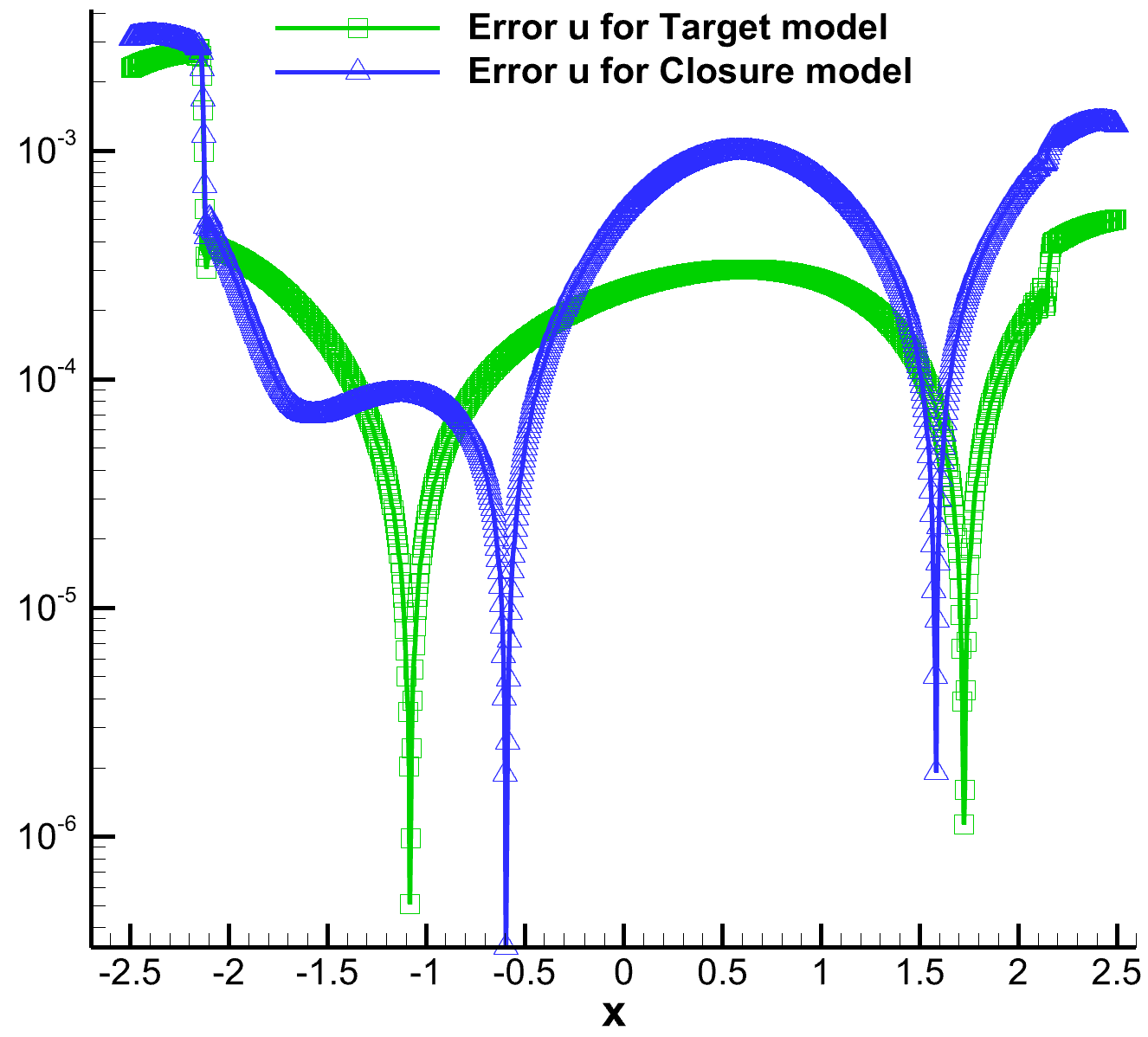}
      \label{fig:the absolute error of velocity}
  }
  \quad
  \subfloat[Comparison of the numerical solution and absolute error of pressure obtained by two models at $t = 0.3$]{
      \centering  
      \includegraphics[height=6.5cm]{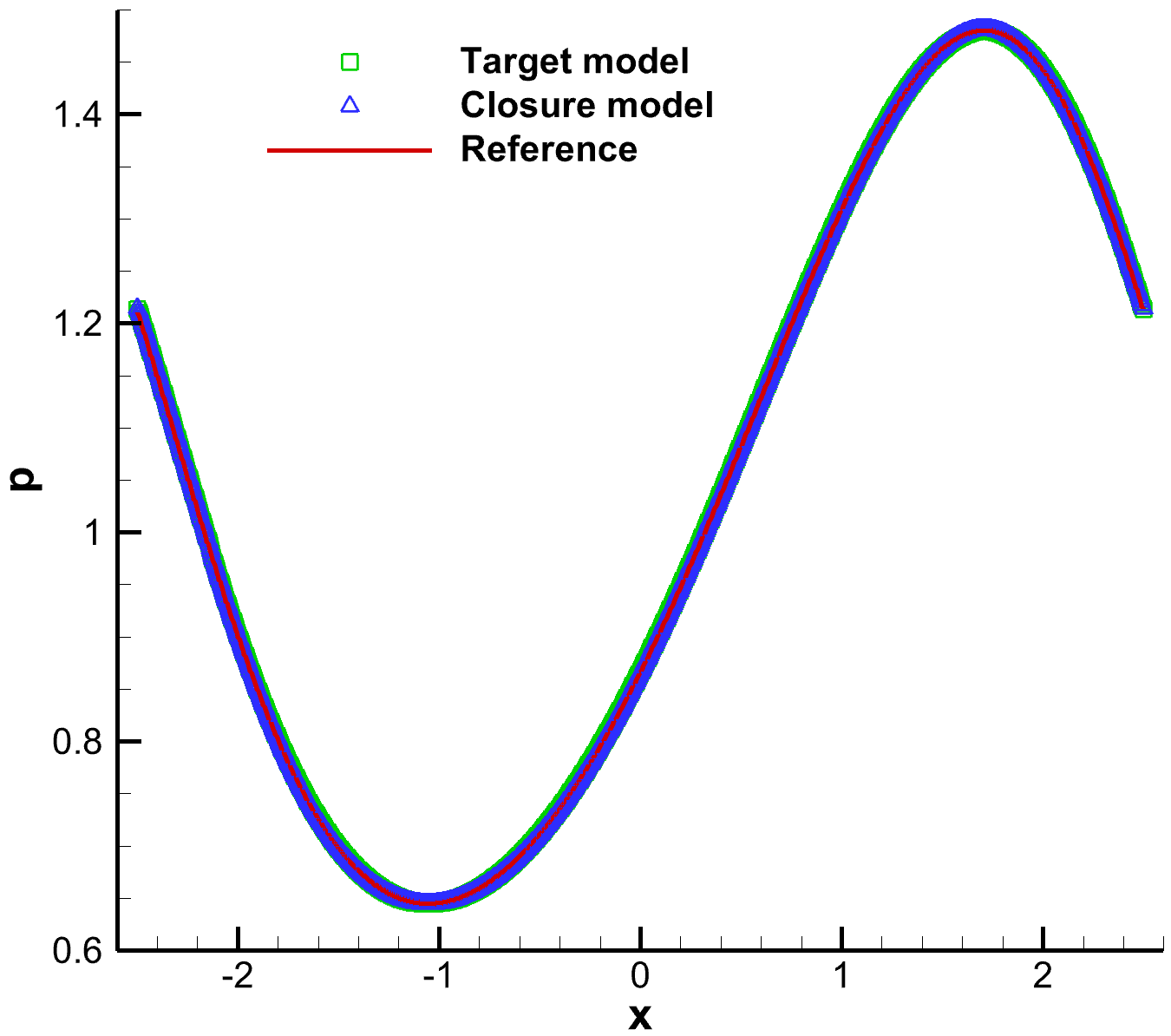}
      \label{fig:the numerical solutions of pressure}   
      \includegraphics[height=6.5cm]{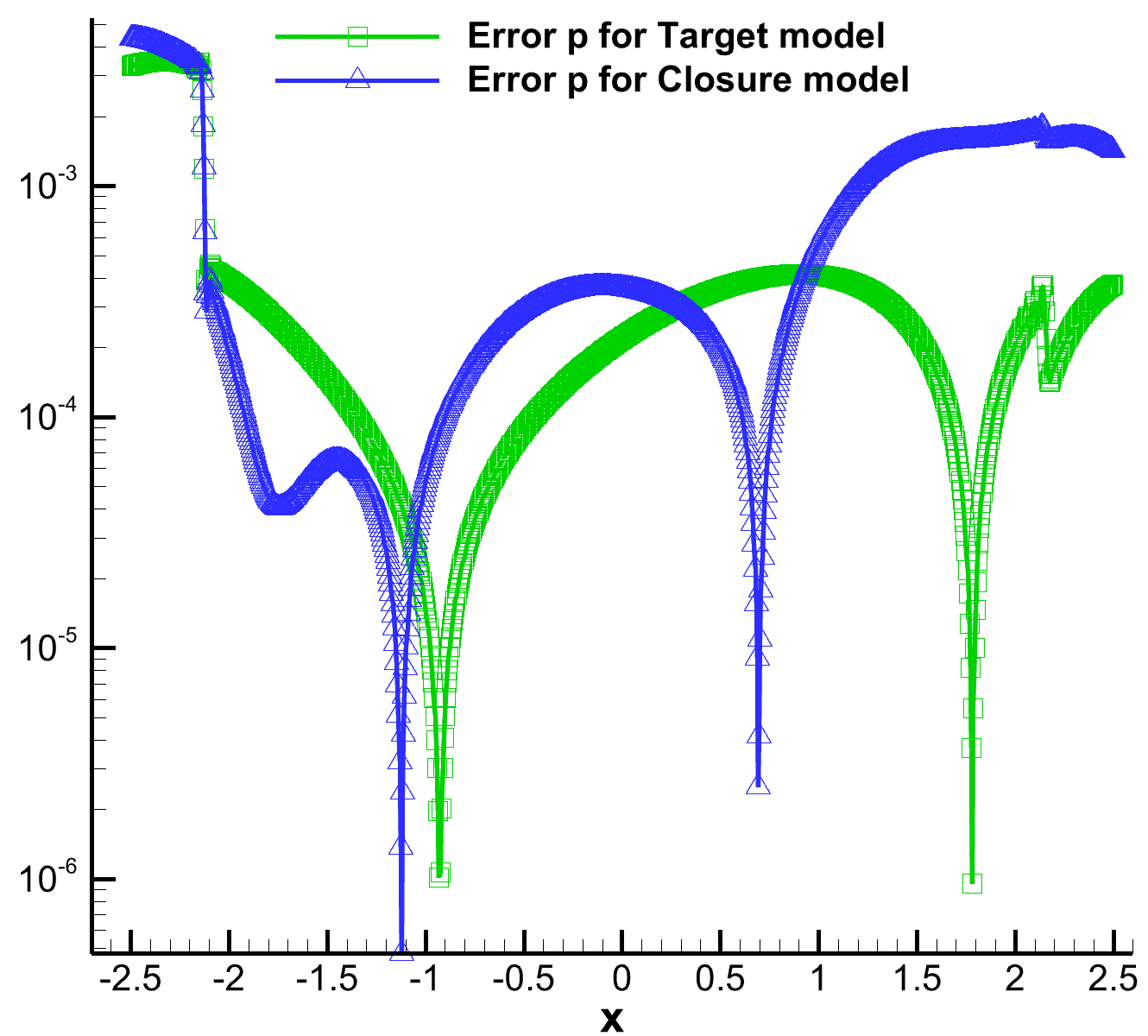}
      \label{fig:the absolute error of pressure}
  }
  \renewcommand{\figurename}{Fig.}
  \caption{At $t = 0.3$, the comparison of the predicted solutions for each physical quantity and their corresponding absolute errors between the closure model of the N-A equation of state and the target model in the 1D smooth periodic test case}
  \label{fig:Comparison image of the predicted solution of eos2 test example 1}
\end{figure}

Figure \ref{fig:Comparison image of the predicted solution of eos2 test example 1} shows the comparison of numerical solutions for $\rho$, $u$, and $p$ at $t = 0.3$ under the two models. From the error analysis diagram, it is clearly evident that the numerical results obtained using the closure model and coupling algorithm developed in this method are highly consistent with those of the target model when solving new problems. This fully demonstrates that the closure model constructed from sparse data can accurately capture the intrinsic laws of complex equations, even under complicated working conditions. Moreover, the coupled algorithm maintains numerical accuracy and stability when solving new problems.

\paragraph{Test Case2:}
The Sod problem is a classical 1D Riemann problem that has been extensively studied. Here, we utilize this problem to evaluate the closure model, and its formulation is as follows:
\begin{equation}
(\rho, u, p) = 
\begin{cases} 
(1, 0, 1), & \text{if }\quad  x \leq 0.5, \\
(0.125, 0, 0.1), & \text{if}\quad 0.5 < x 
\end{cases}
\end{equation}
The computational domain is $\Omega = [0, 1]$, and the final time is $T = 0.2$. The number of grid points is chosen as $N = 200$, and the third-order WENO-Z scheme is used for computation. The reference solution is obtained using a third-order WENO-Z spatial discretization and third-order Runge-Kutta (RK) time integration with $N = 5000$ grid points.

\begin{figure}[!htb]
  \centering
  \includegraphics[height=7cm]{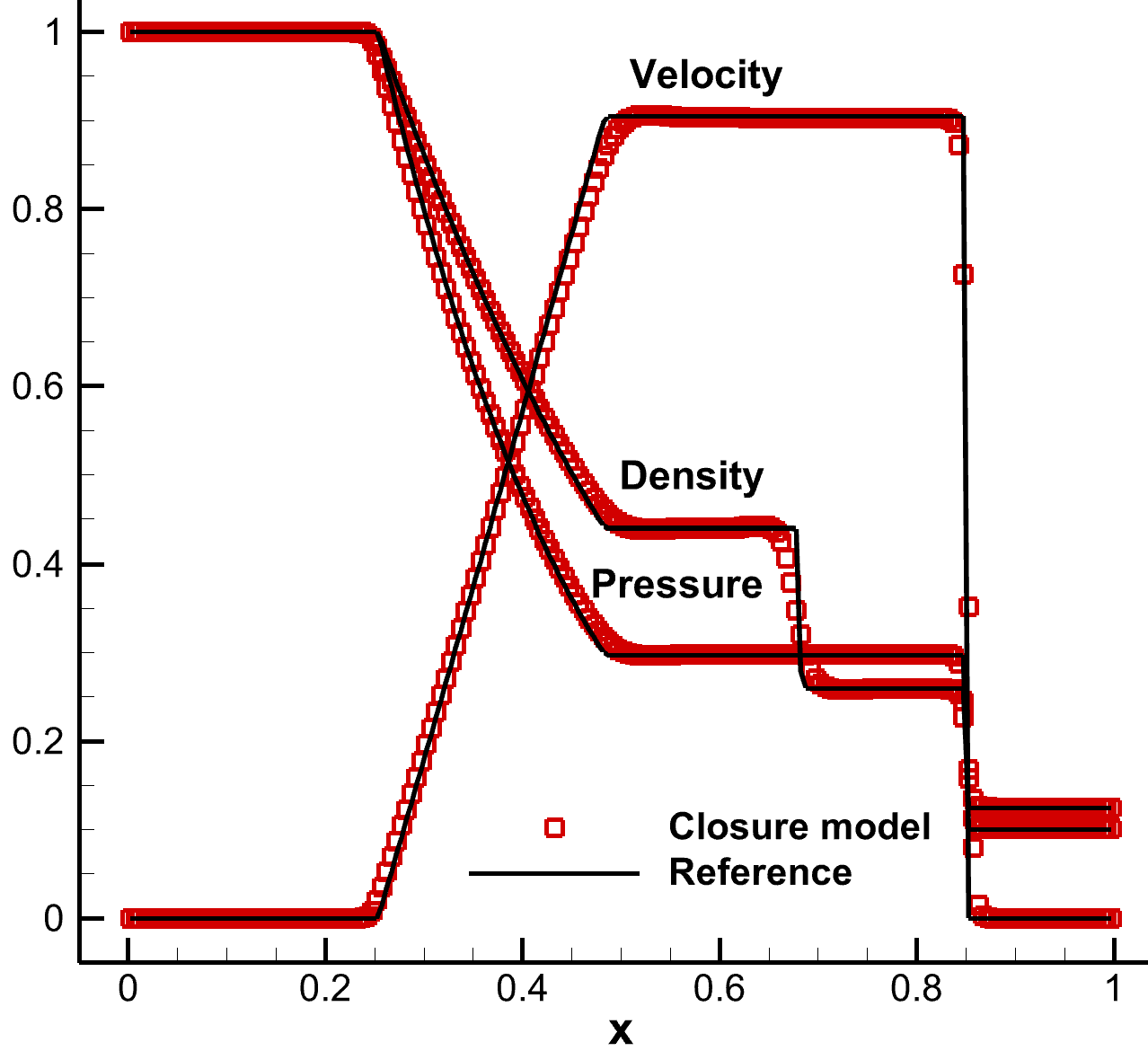}  
  \includegraphics[height=7cm]{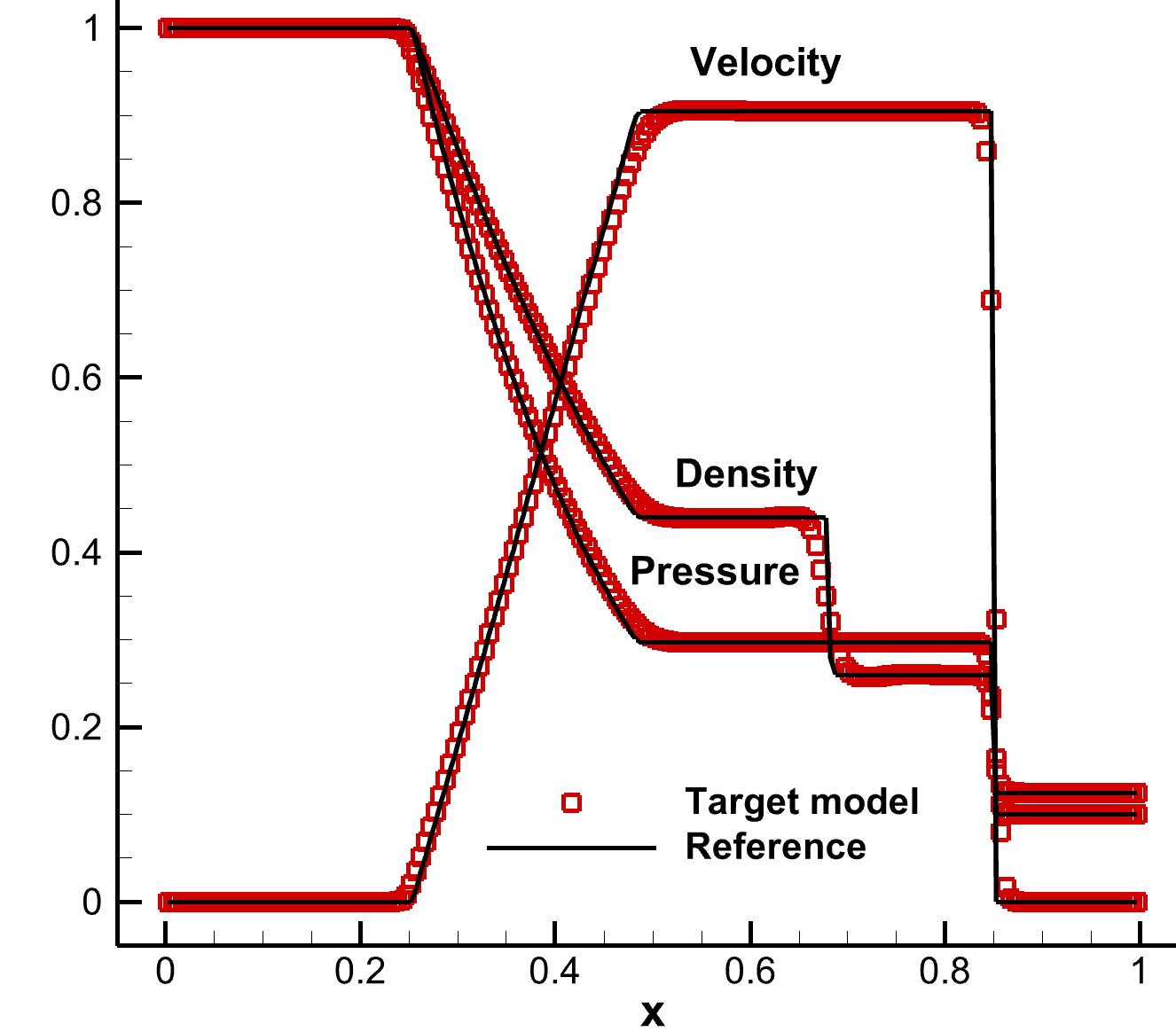}
  \renewcommand{\figurename}{Fig.}
  \caption{At $t = 0.2$, the comparison of predicted solutions for each physical quantity between the closure model and the target model using the N-A equation of state in the 1D Riemann problem}
  \label{fig:Comparison chart of 1D Discontinuous problem}
\end{figure}

Figure \ref{fig:Comparison chart of 1D Discontinuous problem} shows the comparison of numerical solutions for $\rho$, $u$, and $p$ at $t = 0.2$ under the two models. From the figure, the closure model and coupling algorithm constructed using the proposed method exhibit an extremely high degree of agreement with the target model in terms of numerical results when solving new problems. This indicates that the closure model developed in this study is capable of accurately capturing the underlying patterns of complex equations. It also demonstrates that the coupled algorithm proposed in this paper maintains numerical accuracy and stability in new scenarios, showcasing its reliability. 

\paragraph{Test Case3:}
Although the closure model is constructed based on one-dimensional data, we also hope to apply it to higher-dimensional cases. To this end, we test it on a 2D periodic case with the initial conditions set as follows:
\begin{equation}
\left\{
\begin{aligned}
\rho(x,0) &= 0.5,\\
u(x,0)  &= 0.3\sin(2\pi x),\\
v(x,0)  &= 0.3\sin(2\pi y),\\
p(x,0)&= 0.5
\end{aligned}
\right.
\end{equation}
The computational domain is $\Omega = [0, 1] \times [0, 1]$, and the final time is $T = 0.1$. The grid points are chosen as $N \times N = 400 \times 400$, and the fifth-order WENO-Z scheme is used for the computation.

\begin{figure}[!htb]
  \centering
  \subfloat[Density (numerical solution given by the closure model)]{
      \centering
      \includegraphics[height=6.5cm]{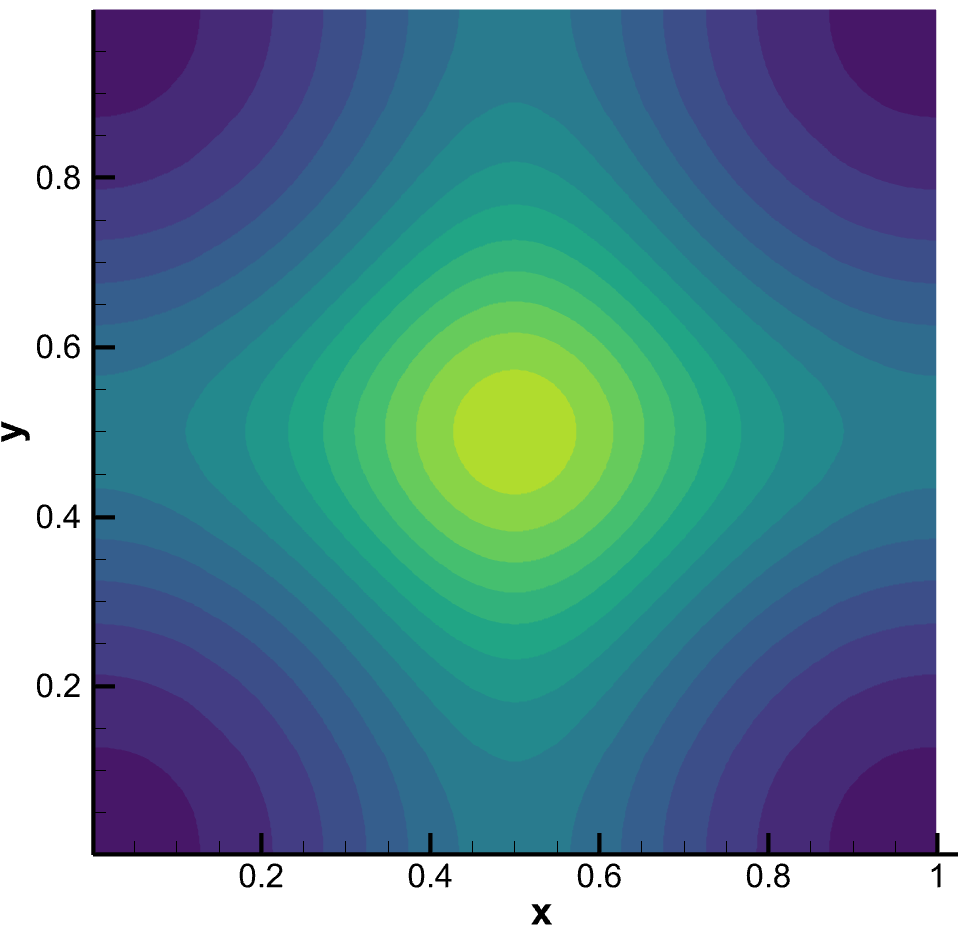}
  }
  \quad
  \subfloat[Density (numerical solution given by the target model)]{
      \centering
       \includegraphics[height=6.5cm]{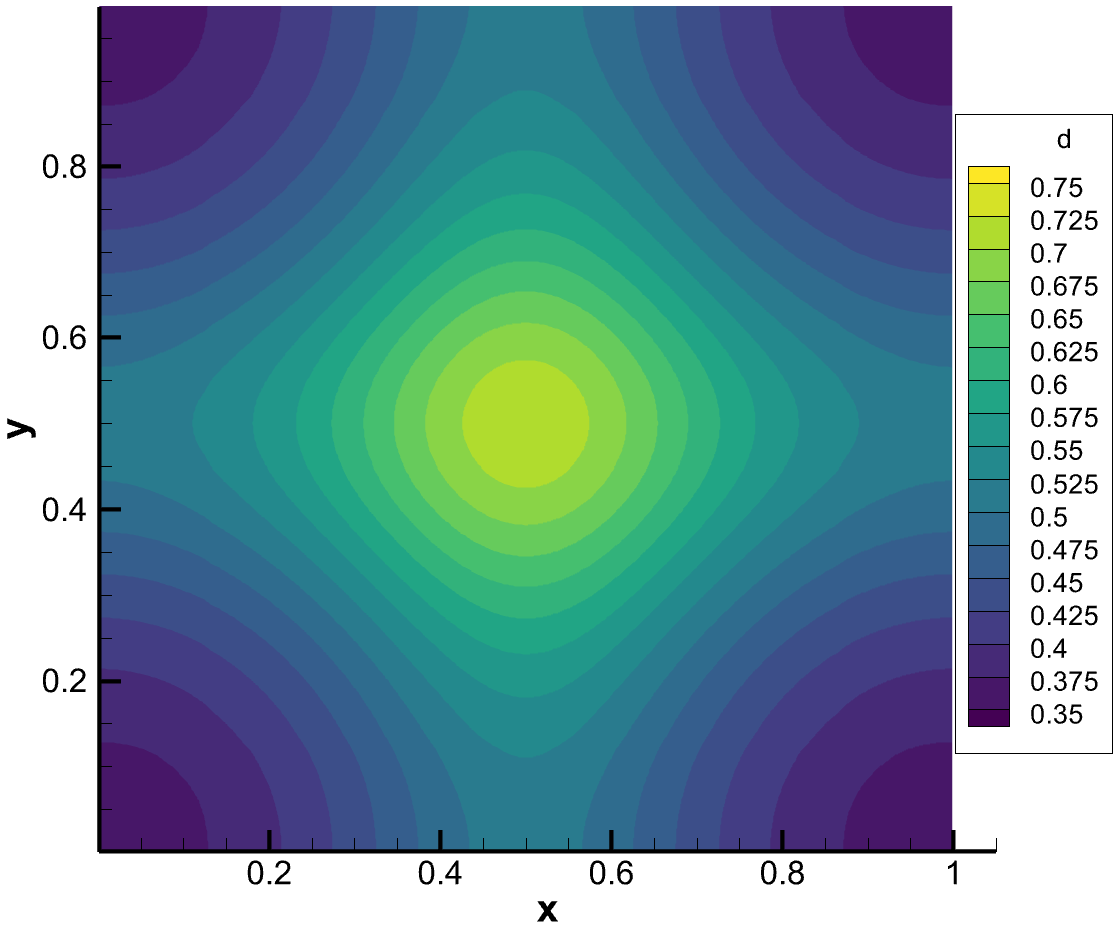}
  }
  \quad
  \subfloat[Velocity u (numerical solution given by the closure model)]{
      \centering
       \includegraphics[height=6.5cm]{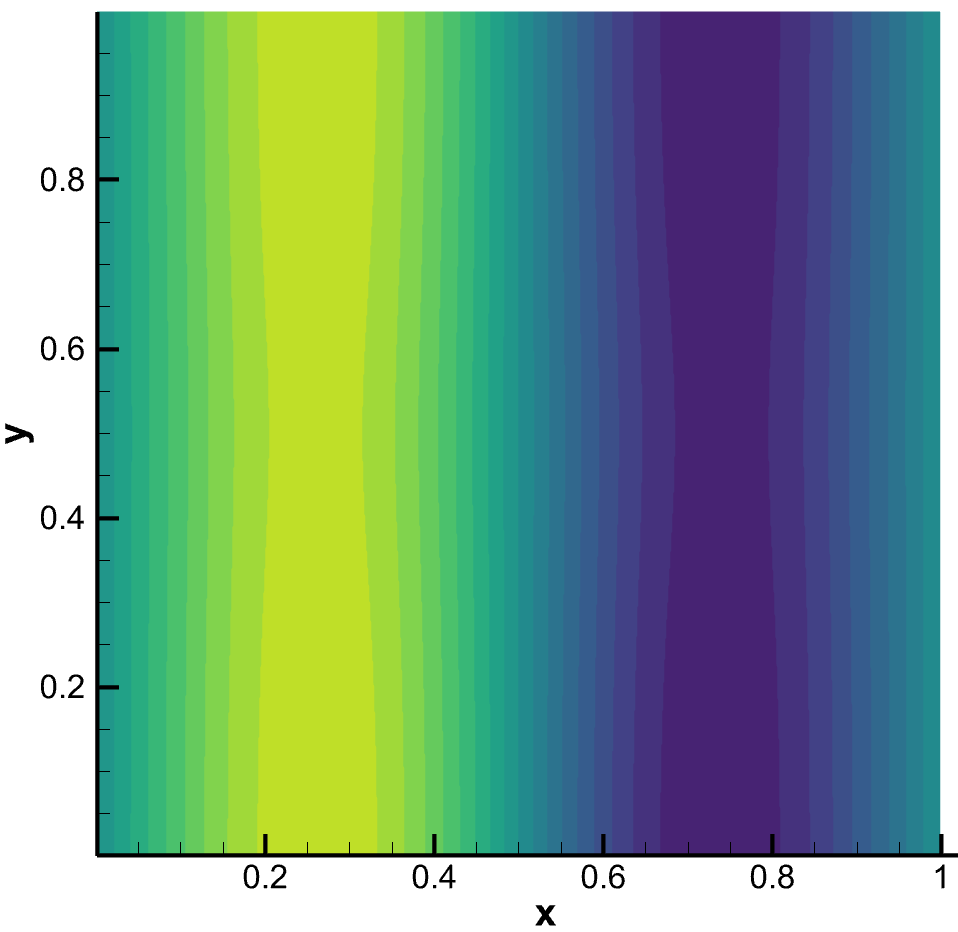}
  }
  \quad
  \subfloat[Velocity u (numerical solution given by the target model)]{
      \centering
      \includegraphics[height=6.5cm]{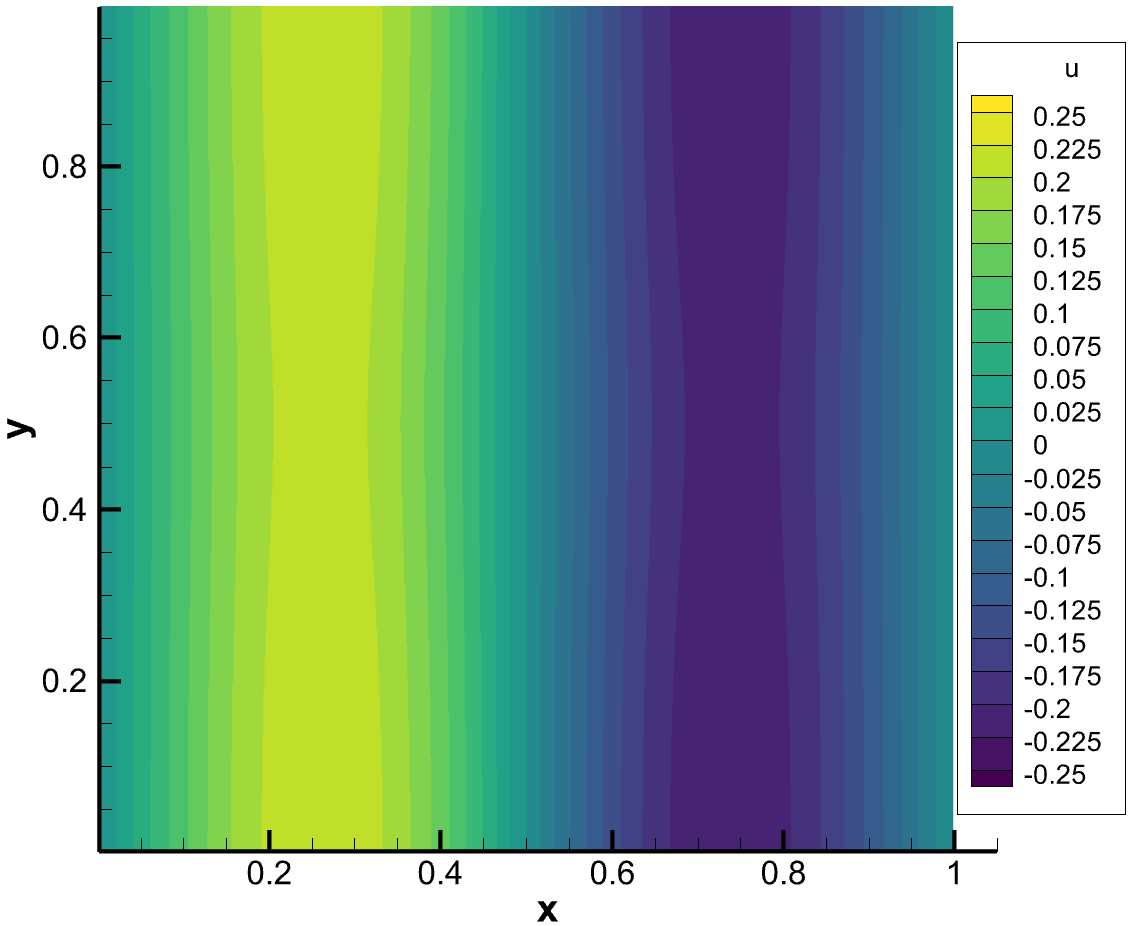}
  }
  \renewcommand{\figurename}{Fig.}
  \caption{At $t = 0.1$, the comparison of the predicted solutions for density and velocity u between the closure model of the N-A equation of state and the target model in the 2D smooth periodic test case(part 1)}
  \label{fig:eos2 2D smooth the predicted solutions for rho and u}
\end{figure}

\begin{figure}[!htb]
  \centering
  \subfloat[Velocity v (numerical solution given by the closure model)]{
      \centering
      \includegraphics[height=6.5cm]{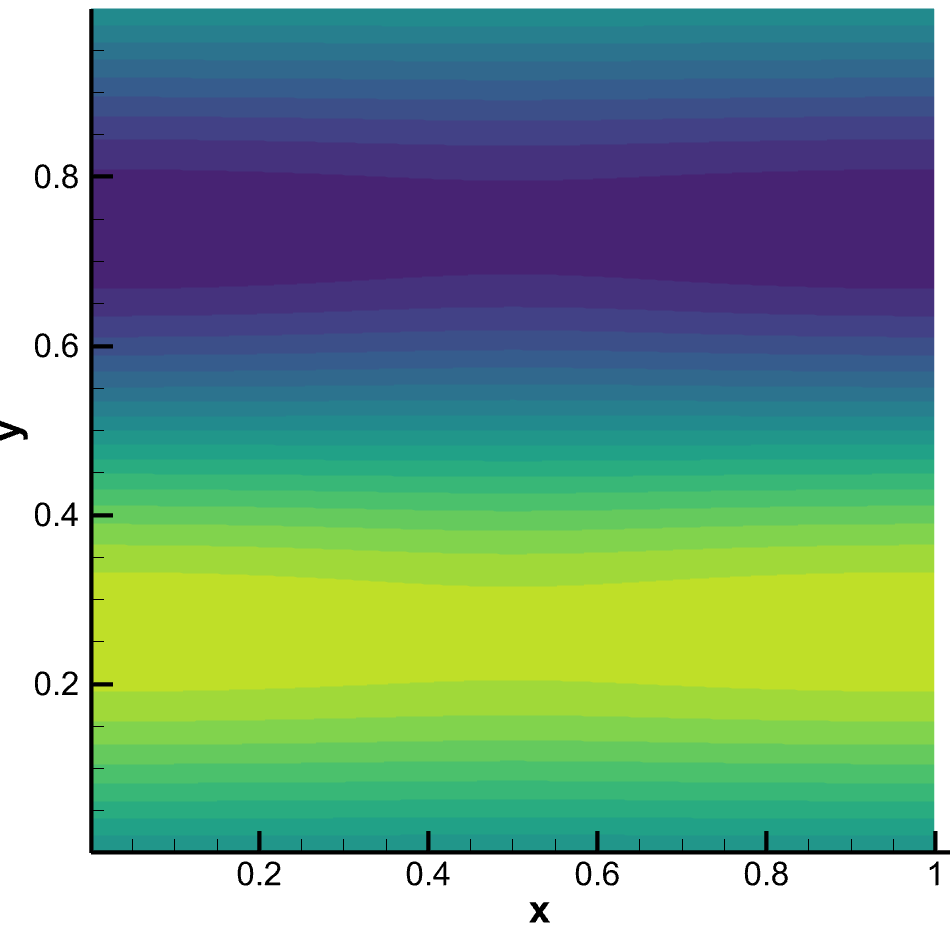}
  }
  \quad
  \subfloat[Velocity v (numerical solution given by the target model)]{
      \centering
       \includegraphics[height=6.5cm]{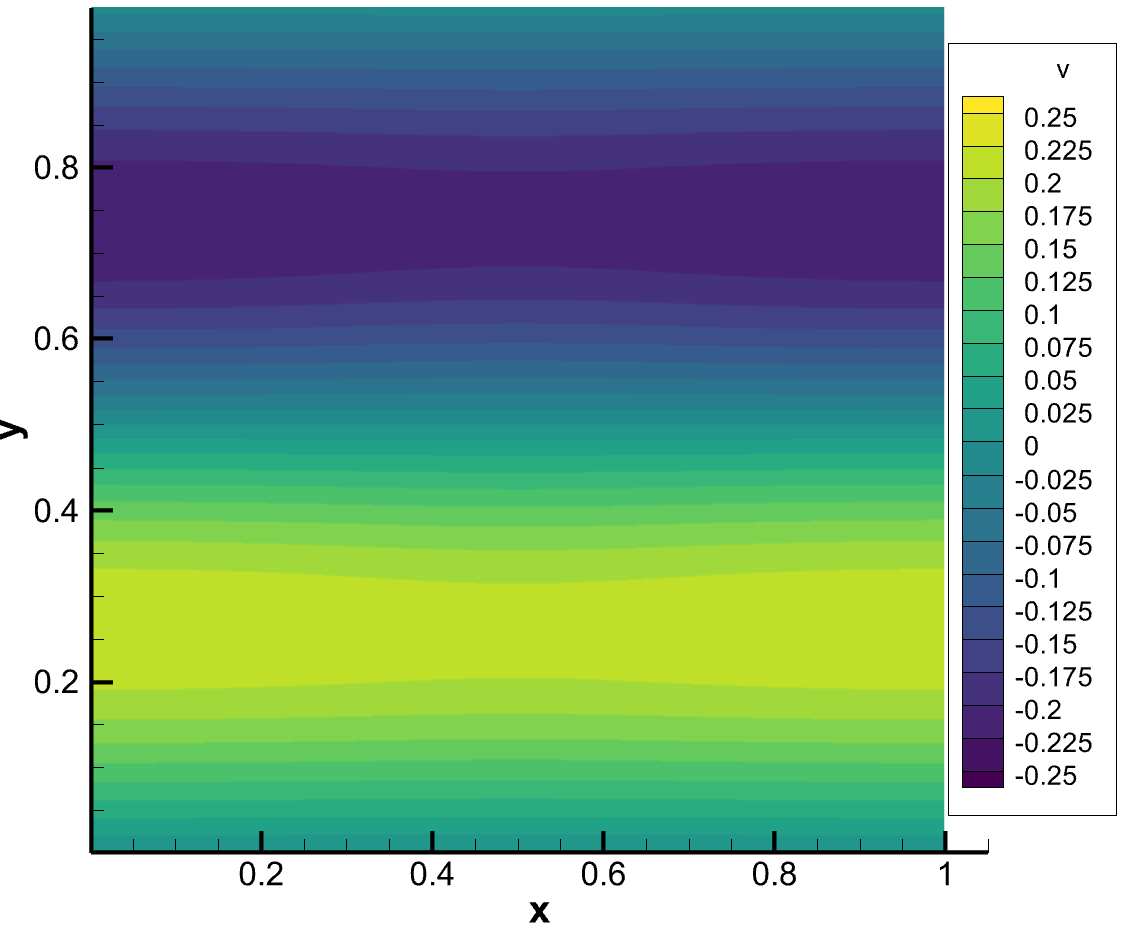}
  }
  \quad
  \subfloat[Pressure (numerical solution given by the closure model)]{
      \centering
       \includegraphics[height=6.5cm]{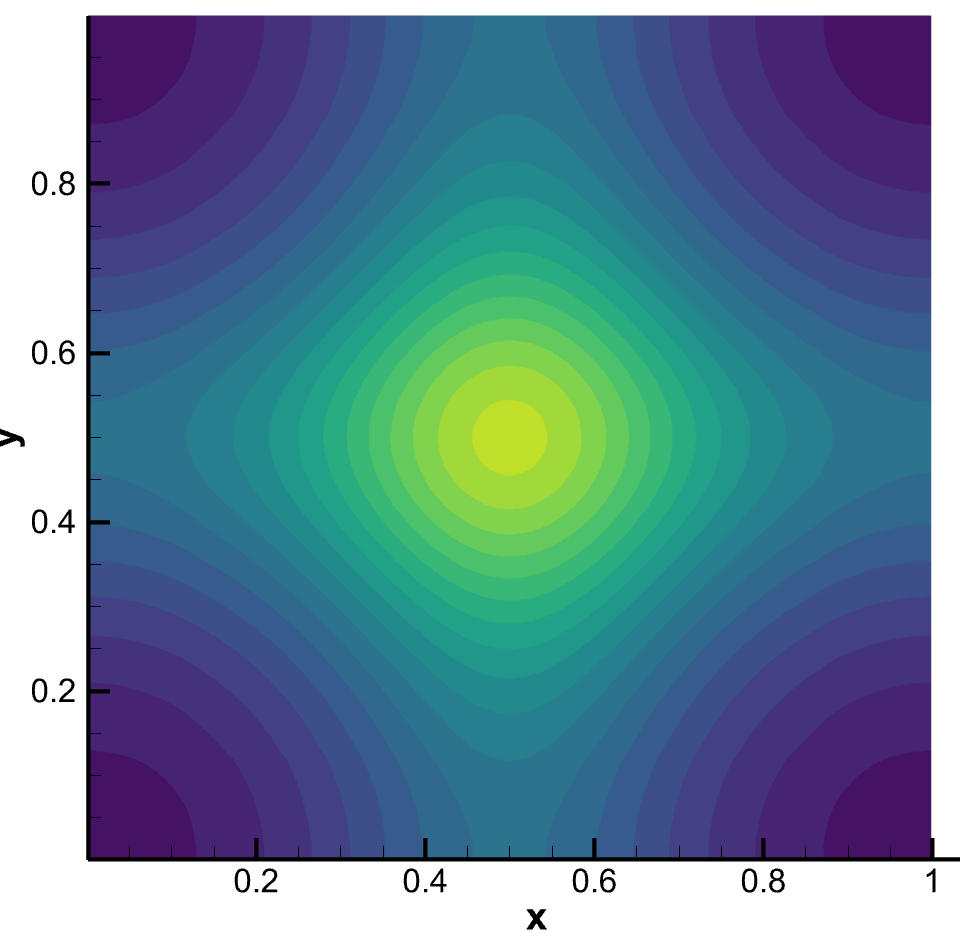}
  }
  \quad
  \subfloat[Pressure (numerical solution given by the target model)]{
      \centering
      \includegraphics[height=6.5cm]{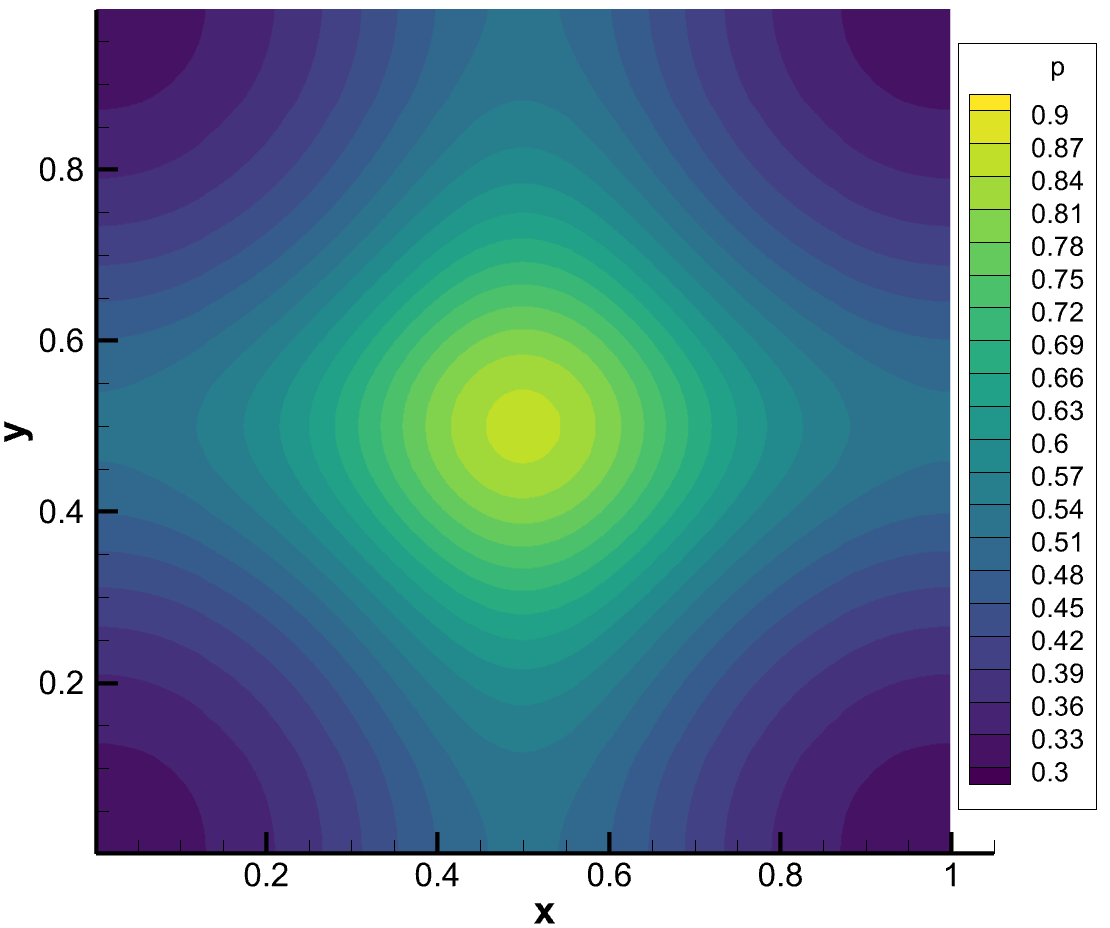}
  }
  \renewcommand{\figurename}{Fig.}
  \caption{At $t = 0.1$, the comparison of the predicted solutions for velocity v and pressure between the closure model of the N-A equation of state and the target model in the 2D smooth periodic test case(part 2)}
  \label{fig:eos2 2D smooth the predicted solutions for velocity v and pressure}
\end{figure}

Figures \ref{fig:eos2 2D smooth the predicted solutions for rho and u} and \ref{fig:eos2 2D smooth the predicted solutions for velocity v and pressure} show the comparison of numerical solutions for $\rho$, $u$, $v$, and $p$ at $t = 0.1$ under the two models. As can be seen from the figure, the closure model and coupling algorithm developed using our method achieve a high degree of consistency with the target model in terms of numerical performance. This implies that the closure model constructed using the proposed method for complex equations of state, even when built upon 1D data, can still be successfully applied to 2D test cases. In addition, the coupling algorithm demonstrates excellent numerical accuracy and stability by maintaining a high level of consistency with the target model in numerical solutions.

\paragraph{Test Case4:}
To further verify the applicability and accuracy of the closure model, a 2D Riemann problem with the following initial values is considered for testing:
\begin{equation}
(\rho, u, v, p) = 
\begin{cases} 
(0.5, -0.5, 0.35, 0.5) & \text{if }\quad 0 \leq x \leq 0.5,\quad 0 \leq y \leq 0.5, \\
(1.0, 0.5, 0.35, 0.5) & \text{if }\quad 0 \leq x \leq 0.5, \quad 0.5 < y \leq 1, \\
(1.5, -0.5, -0.35, 0.5) & \text{if }\quad 0.5 < x \leq 1, \quad 0 \leq y \leq 0.5, \\
(0.5, 0.5, -0.35, 0.5) & \text{if }\quad 0.5 \leq x \leq 1, \quad 0.5 < y \leq 1.
\end{cases}
\end{equation}
The computational domain is $\Omega = [0, 1] \times [0, 1]$, and the final time is $T = 0.2$. The grid points are chosen as $N \times N = 400 \times 400$, and the third-order WENO-Z scheme is used for computation.

\begin{figure}[!htb]
  \centering
  \subfloat[Density (numerical solution given by the closure model)]{
      \centering
      \includegraphics[height=6.5cm]{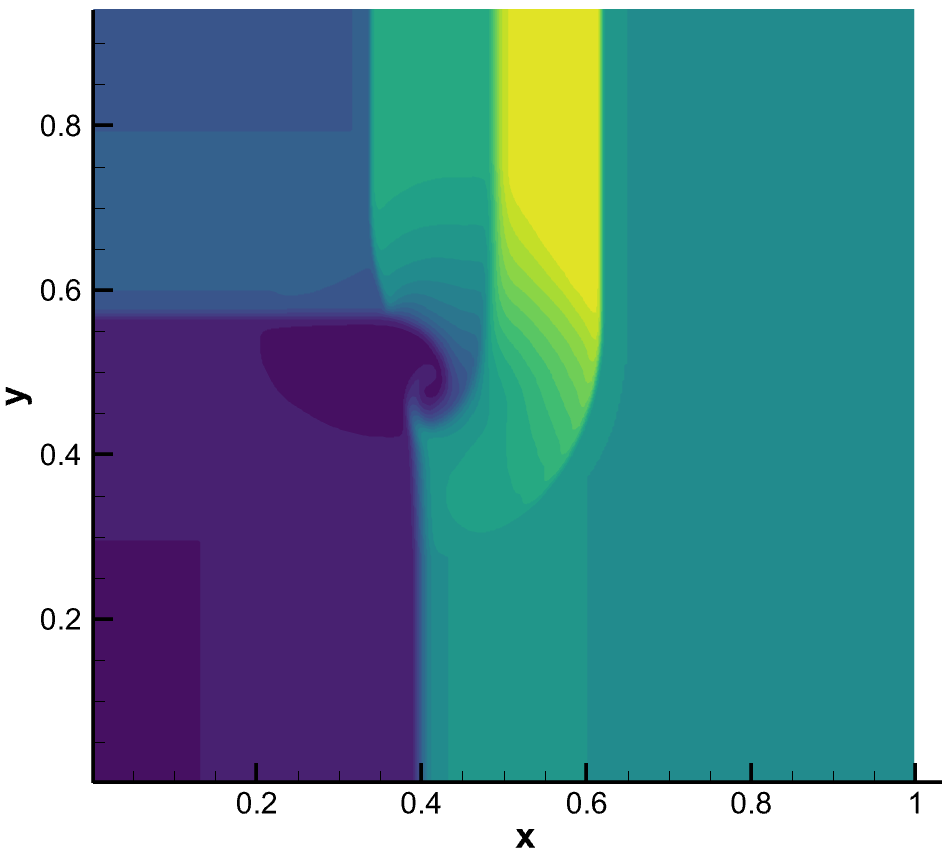}
  }
  \quad
  \subfloat[[Density (numerical solution given by the target model)]{
      \centering
       \includegraphics[height=6.5cm]{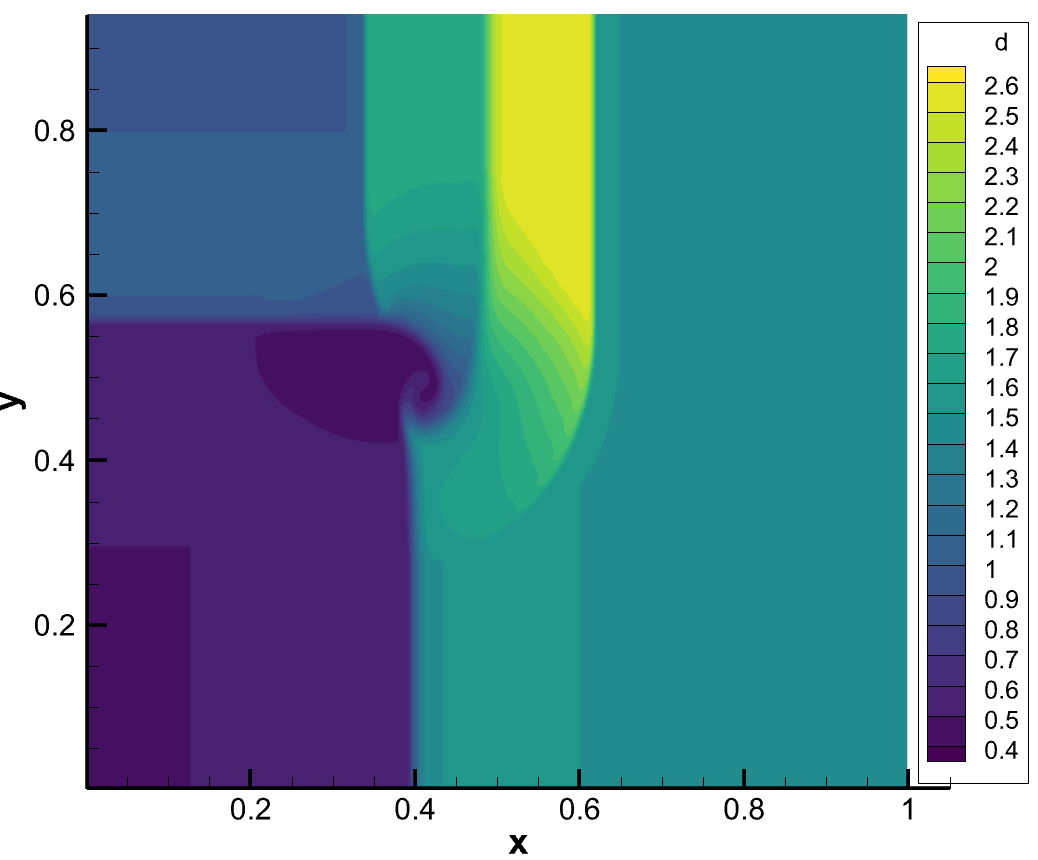}
  }
  \quad
  \subfloat[Velocity u (numerical solution given by the closure model)]{
      \centering
       \includegraphics[height=6.5cm]{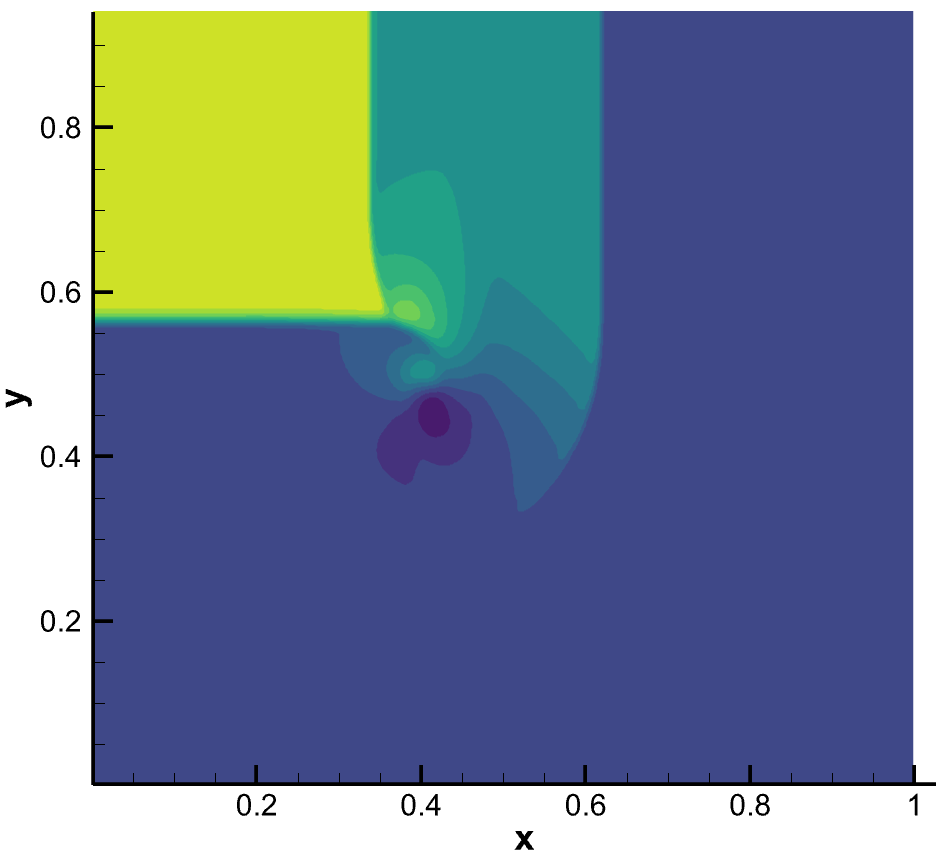}
  }
  \quad
  \subfloat[Velocity u (numerical solution given by the target model)]{
      \centering
      \includegraphics[height=6.5cm]{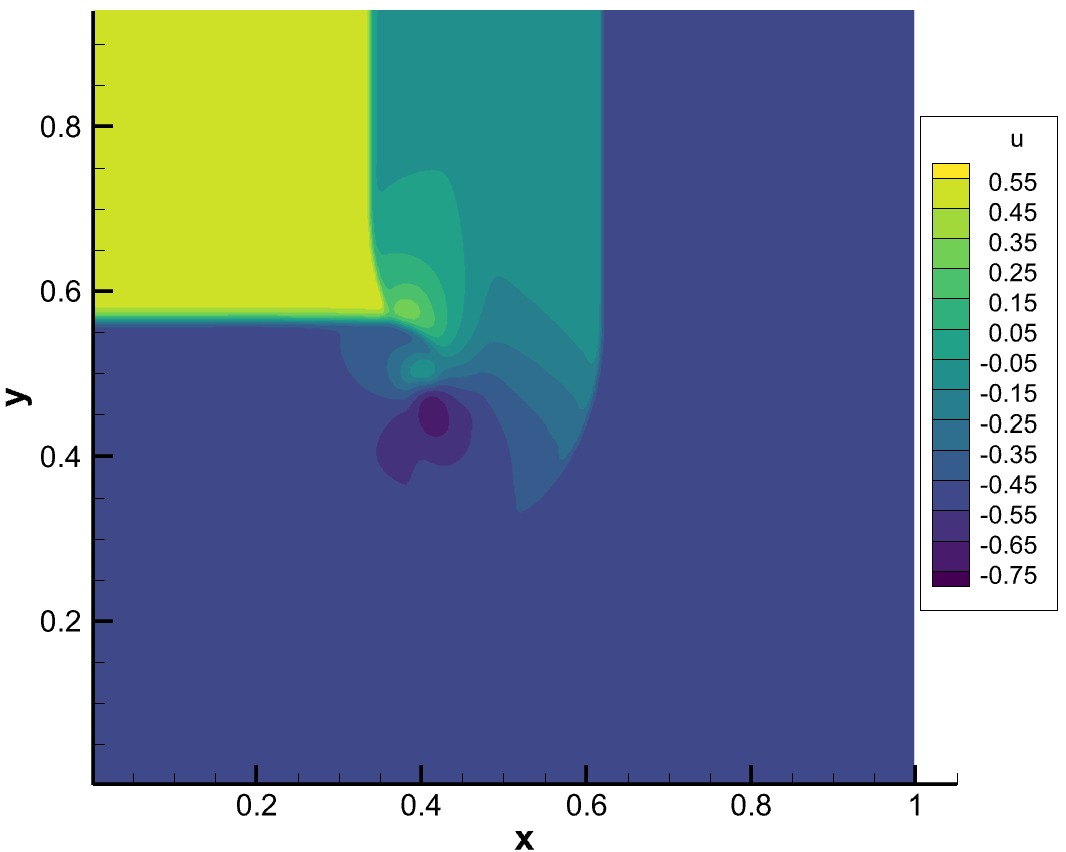}
  }
  \renewcommand{\figurename}{Fig.}
  \caption{At $t = 0.2$, the comparison of the predicted solutions for density and velocity u between the closure model of the N-A equation of state and the target model in the 2D Riemann problem(part 1)}
  \label{fig:eos2 2D Riemann the predicted solutions for density and velocity u}
\end{figure}

\begin{figure}[!htb]
  \centering
  \subfloat[Velocity v (numerical solution given by the closure model)]{
      \centering
      \includegraphics[height=6.5cm]{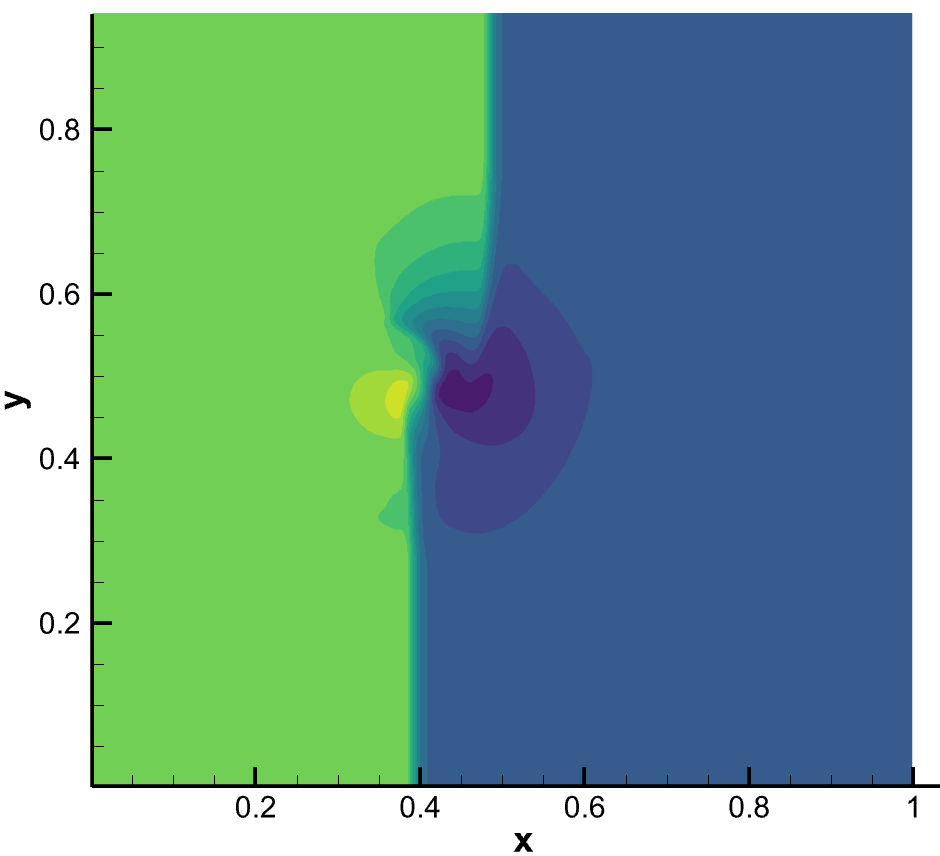}
  }
  \quad
  \subfloat[Velocity v (numerical solution given by the target model)]{
      \centering
       \includegraphics[height=6.5cm]{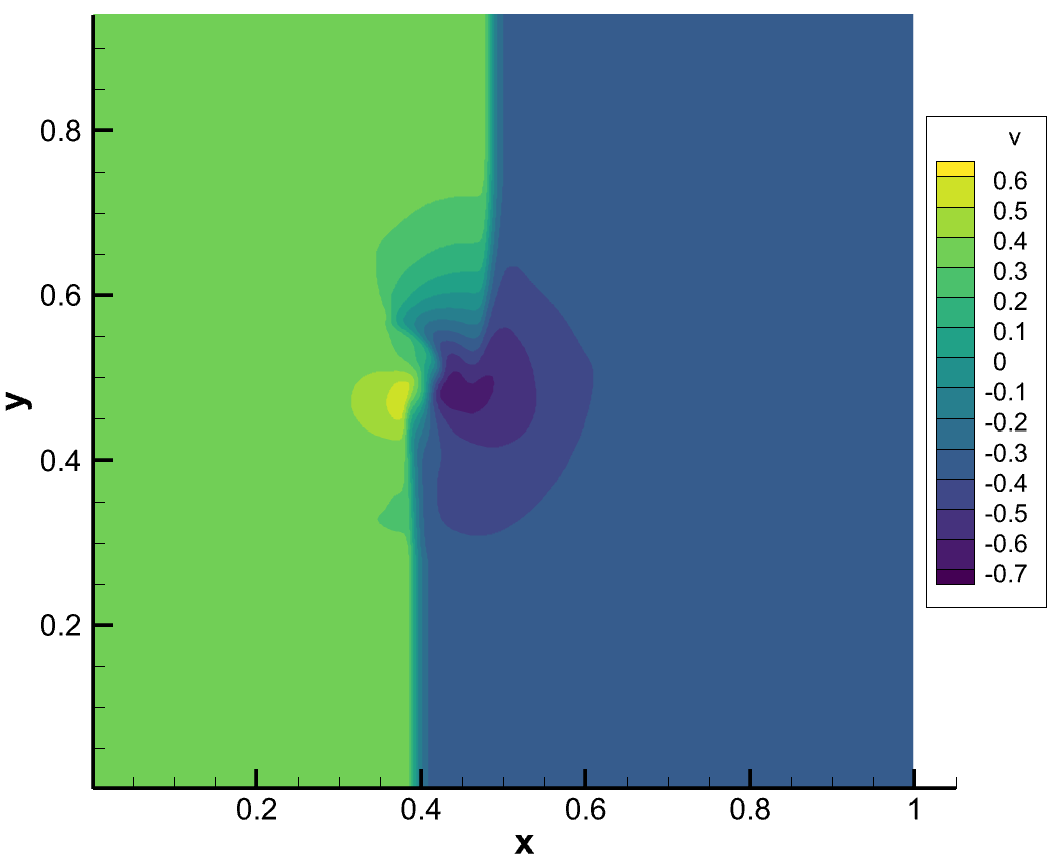}
  }
  \quad
  \subfloat[Pressure (numerical solution given by the closure model)]{
      \centering
       \includegraphics[height=6.5cm]{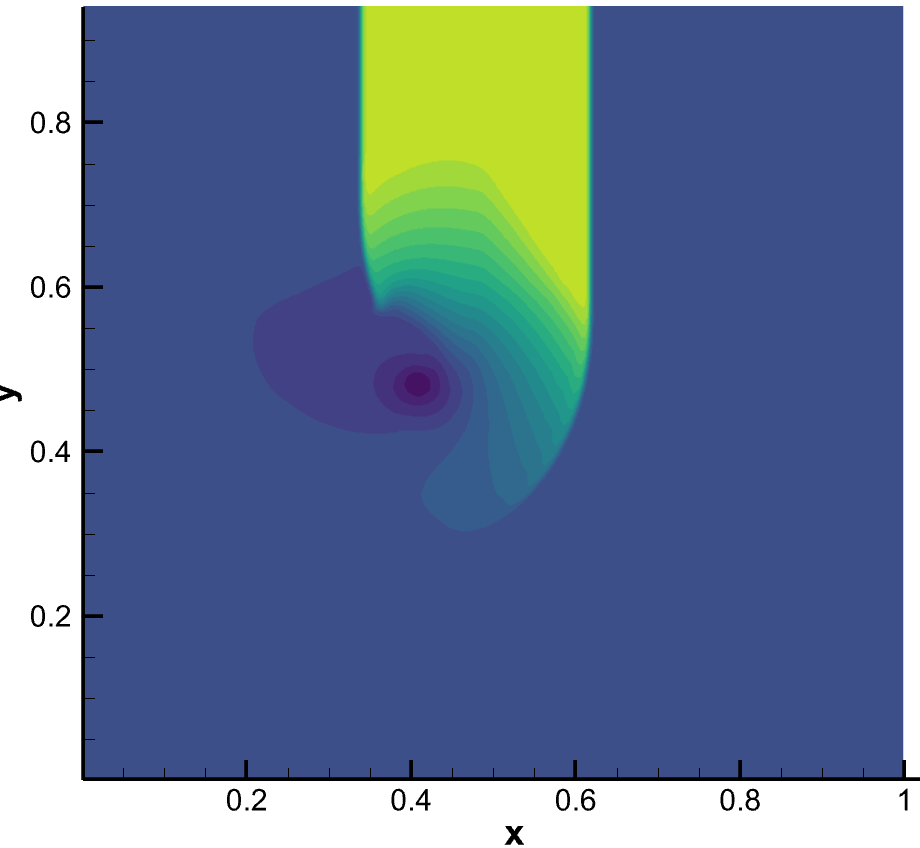}
  }
  \quad
  \subfloat[Pressure (numerical solution given by the target model)]{
      \centering
      \includegraphics[height=6.5cm]{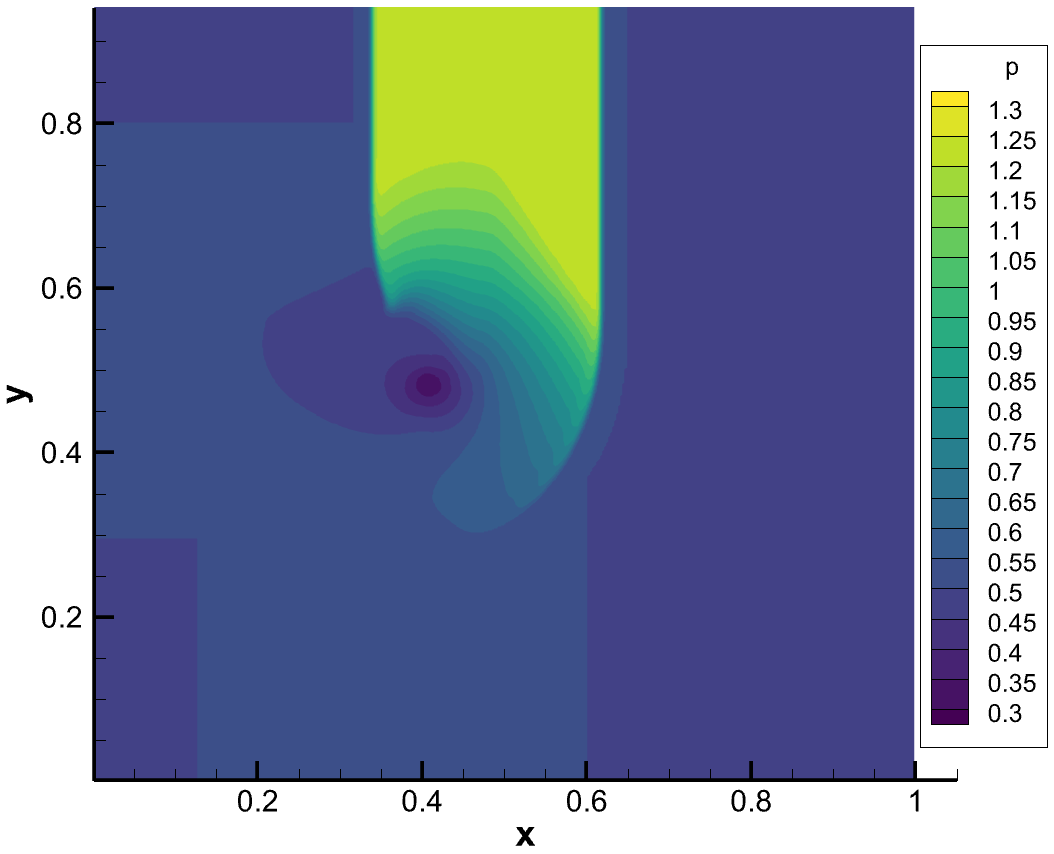}
  }
  \renewcommand{\figurename}{Fig.}
  \caption{At $t = 0.2$, the comparison of the predicted solutions for velocity v and pressure between the closure model of the N-A equation of state and the target model in the 2D Riemann problem(part 2)}
  \label{fig:eos2 2D Riemann the predicted solutions for velocity v and pressure}
\end{figure}

Figures \ref{fig:eos2 2D Riemann the predicted solutions for density and velocity u} and \ref{fig:eos2 2D Riemann the predicted solutions for velocity v and pressure} show the comparison of numerical solutions for $\rho$, $u$, $v$, and $p$ at $t = 0.2$ under the two models. From the figures, although there are slight differences in complex regions such as shock waves and contact discontinuities, the overall performance of the closure model and coupling algorithm is highly consistent with the target model. This fully demonstrates the reliability and accuracy of the proposed method in capturing complex flow field characteristics.

\section{Conclusion\label{Conclusion}}
In this paper, we introduce a general method for constructing closure models with sparse data. This approach leverages a series-parallel multi-network architecture within the PINNs framework to overcome limitations posed by sparse data, resulting in a closure model with robust generalization capabilities. Additionally, we integrate the newly developed closure model into traditional PDE-solving codes to address the high-precision demands of complex engineering problems. The method's effectiveness is validated through two distinct equation systems. Given its universal closure modeling capability and minimal data dependency, this approach holds significant promise as a valuable tool for practical engineering applications.


\setcounter{secnumdepth}{0}
\section{Acknowledgments}
The authors would like to thank all the members of YONG Heng's team "AI++" for their help and fruitful discussions. This work is partially supported by the National Natural Science Foundation of China(NSFC) under Grant Nos.(12331010) and the National Safety Academic Fund(NSAF) under Grant No.U2230208.

\setcounter{secnumdepth}{2}

\appendix
\renewcommand{\thesection}{Appendix \Alph{section}:}
\renewcommand{\thesubsection}{Appendix \Alph{section}.\arabic{subsection}:}

\section{Definition of Loss Function for PINNs-WE}
In Sec.~\ref{PINNs-WE}, we provide an introduction to $\mathcal{L}_{PDE}$, and here we present the introduction of $\mathcal{L}_{RH}$ and $\mathcal{L}_{CONs}$. And more details can be found in~\cite{PINNs_loss_using}.

\subsection{Definition of $\mathcal{L}_{RH}$}
Although solutions do not exist at discontinuities such as shock waves, the physical conservation laws still hold and must satisfy the Rankine–Hugoniot (RH) relations, which ensure compliance with conservation laws at these discontinuities. For the 1D Euler equations, according to the RH relations, the following RH constraints can be constructed:
\begin{equation}
\begin{split}
\mathcal{L}_{RH} = {} & 
\frac{1}{|S_{RH}|}\sum_{t_i,\mathbf{X}_i \in S_{RH}}|\lambda_2(\mathbf{U},\mathbf{U_L}) f_1(\mathbf{U},\mathbf{U_L}) |^2+\frac{1}{|S_{RH}|}\sum_{t_i,\mathbf{X}_i \in S_{RH}}|\lambda_2(\mathbf{U},\mathbf{U_L}) f_2(\mathbf{U},\mathbf{U_L})|^2
\end{split}
\label{eq:1D Euler's RH loss function}
\end{equation}
Here, $\mathbf{U_L}=\mathbf{U}(t,\textbf{x}+\Delta \textbf{x})$, and $f_1$ and $f_2$ are expressed as:
\begin{equation}
\begin{split}
f_1(\mathbf{U},\mathbf{U_L}) ={} & 
\rho\rho_L(u-u_L)^2 - (\rho-\rho_L)(p-p_L) \\
f_2(\mathbf{U},\mathbf{U_L}) ={} & 
\rho\rho_L(e-e_L)-\frac{1}{2}(\rho-\rho_L)(p+p_L) \\
\end{split}
\end{equation}
$\lambda_2$ is a filter, defined as in Equation \ref{eq:filter}. $\varepsilon_1 = \varepsilon_2 = 0.1$ are two parameters for detecting shocks, and their values are referenced from the article \cite{PINNs_loss_using}.
\begin{equation}
\begin{split}
\lambda_2(\mathbf{U},\mathbf{U_L}) =
    \begin{cases}
        |(p - p_L)(u - u_L)| & \text{if } \; |p - p_L| > \varepsilon_1 \quad \text{and} \quad |u - u_L| > \varepsilon_2, \\
        0 & \text{elsewhere}.
    \end{cases}
\end{split}
\label{eq:filter}
\end{equation}

\subsection{Definition of $\mathcal{L}_{CONs}$}
Although we can ensure energy conservation at the shock by using the RH relations, PINNs cannot achieve the theoretically total conservative property in the same way as finite difference or finite volume methods. Therefore, a soft global conservation constraint needs to be added, which is defined as:
\begin{equation}
\begin{aligned}
\mathcal{L}_{CONs} ={} & 
(\text{Mas}(t_2)-\text{Mas}(t_1)-\text{BD}_\text{Mas})^2+(\text{Mom}(t_2)-\text{Mom}(t_1)-\text{BD}_\text{Mom})^2\\
& +(\text{Ene}(t_2)-\text{Ene}(t_1)-\text{BD}_\text{Ene})^2
\end{aligned}
\label{eq:Definition of Conservation Loss}
\end{equation}
Here, the conservation approximation for the interior points is given by:
\begin{equation}
\begin{aligned}
\text{Mas}(t_k) &= \frac{V}{|S_{Con}(t_k)|} \sum_{x \in S_{Con}(t_k)} \rho(x), \\
\text{Mom}(t_k) &= \frac{V}{|S_{Con}(t_k)|} \sum_{x \in S_{Con}(t_k)} \rho(x)\vec{u}(x), \\
\text{Ene}(t_k) &= \frac{V}{|S_{Con}(t_k)|} \sum_{x \in S_{Con}(t_k)} E(x),
\end{aligned}
\label{eq:conservation approximation for the interior points}
\end{equation}
The approximation at the boundary is:
\begin{equation}
\begin{aligned}
\text{BD}_\text{Mas}(t_1,t_2)&=\frac{(t_2 - t_1)A}{|S_{BD}|} \sum_{x \in S_{BD}}\rho(x)(\vec{u}(x)\cdot \vec{n}_{\partial V}(x),\\
\text{BD}_\text{Mom}(t_1,t_2)&=\frac{(t_2 - t_1)A}{|S_{BD}|} \sum_{x \in S_{BD}}[\rho(x)\vec{u}(x)^2+p(x)]\cdot \vec{n}_{\partial V}(x),\\
\text{BD}_\text{Ene}(t_1,t_2)&=\frac{(t_2 - t_1)A}{|S_{BD}|} \sum_{x \in S_{BD}}[E(x)+p(x)]\vec{u}(x)\cdot \vec{n}_{\partial V}(x),
\end{aligned}
\label{eq:conservation approximation for the boundary}
\end{equation}
Where $S_{BD}$ represents the set of boundary points selected within the computation time $t_1 < t < t_2$, $A$ denotes the area of the boundary surface, and $\vec{n}_{\partial V}(x)$ is the unit normal vector of the boundary.

\section{Testing with Ideal Equation of State}
Here, we present the experimental procedure for the ideal equation of state. The ideal equation of state is expressed as:
\begin{equation}
p=e\rho (\gamma -1), \quad \gamma = 1.4
\end{equation}

Here, we also used five groups of sparse experimental data with the same equation of state to train the closure model for the equation of state. In the jointly constructed $Net_2$, we employed a fully connected neural network with 1 hidden layer, where each hidden layer contained 20 neurons. To effectively prevent overfitting, an $L^2$ regularization term \cite{zheng_ze_hua} was incorporated during the training process to constrain the network weights, thereby enhancing the model's generalization ability. The design of the five independent $Net_1$ networks and the detailed conditions of the five data groups are presented as follows:

\textbf{Case1:}
The case adopts periodic boundary conditions, with time $ t \in [0, 0.5] $ and spatial variable $ x \in [-1.0, 1.0] $. The initial values of the physical quantities are given as:
\begin{equation}
\left\{
\begin{aligned}
\rho(x,0) &= 0.35+0.25\sin(\pi x),\\
u(x,0)  &= 1.0,\\
p(x,0)&=0.154-0.03\sin(\pi x)-0.1\sin^2(\pi x)
\end{aligned}
\right.
\end{equation}
In the constructed $Net_1$, we used a neural network with $8$ hidden layers, each containing $50$ neurons. In the $T \times X$ space, $8000$ points were randomly sampled as residual points for the RH loss and equation loss. Additionally, we have a total of $300$ conservation points at both $t_1=0$ and $t_2=0.5$. The data included only a small amount of initial and final-state information.

\textbf{Case2:}
The case adopts periodic boundary conditions, with time $ t \in [0, 0.5] $ and spatial variable $ x \in [-1.0, 1.0] $. The initial values of the physical quantities are given as:
\begin{equation}
\left\{
\begin{aligned}
\rho(x,0) &= 0.75+0.5(1.0-\cos(\pi x)),\\
u(x,0)  &= 0.5\sqrt{1.4}(1.0-\cos(\pi x)),\\
p(x,0)&=0.825+0.55(1.0-\cos(\pi x))
\end{aligned}
\right.
\end{equation}
In the constructed $Net_1$, we used a neural network with $8$ hidden layers, each containing $50$ neurons. In the $T \times X$ space, $8000$ points were randomly sampled as residual points for the RH loss and equation loss. Additionally, we have a total of $400$ conservation points at both $t_1=0$ and $t_2=0.5$. The data included only a small amount of initial and final-state information.

\textbf{Case3:}
The case adopts periodic boundary conditions, with time $ t \in [0, 0.5] $ and spatial variable $ x \in [-1.0, 1.0] $. The initial values of the physical quantities are given as:
\begin{equation}
\left\{
\begin{aligned}
\rho(x,0) &= 0.6+0.5(1.0-\cos(\pi x)),\\
u(x,0)  &= -0.5\sqrt{1.4}(1.0-\cos(\pi x)),\\
p(x,0)&=0.348+0.29(1.0-\cos(\pi x))
\end{aligned}
\right.
\end{equation}
In the constructed $Net_1$, we used a neural network with $8$ hidden layers, each containing $50$ neurons. In the $T \times X$ space, $8000$ points were randomly sampled as residual points for the RH loss and equation loss. Additionally, we have a total of $300$ conservation points at both $t_1=0$ and $t_2=0.5$. The data included only a small amount of initial and final-state information.

\textbf{Case4:}
The case adopts periodic boundary conditions, with time $ t \in [0, 0.5] $ and spatial variable $ x \in [-1.0, 1.0] $. The initial values of the physical quantities are given as:
\begin{equation}
\left\{
\begin{aligned}
\rho(x,0) &= 0.75+0.5(1.0-\cos(\pi x)),\\
u(x,0)  &= 0.3\sqrt{1.4}(1.0-\cos(\pi x)),\\
p(x,0)&=0.3275-0.131\cos(\pi x)-0.1\sin(\pi x)+0.02\sin(2\pi x)
\end{aligned}
\right.
\end{equation}
In the constructed $Net_1$, we used a neural network with $7$ hidden layers, each containing $50$ neurons. In the $T \times X$ space, $8000$ points were randomly sampled as residual points for the RH loss and equation loss. Additionally, we have a total of $300$ conservation points at both $t_1=0$ and $t_2=0.5$. The data included only a small amount of initial and final-state information.

\textbf{Case5:}
The case adopts periodic boundary conditions, with time $ t \in [0, 0.4] $ and spatial variable $ x \in [-1.0, 1.0] $. The initial values of the physical quantities are given as:
\begin{equation}
\left\{
\begin{aligned}
\rho(x,0) &= 0.1+\exp(-5x^2),\\
u(x,0)  &= 0.0,\\
p(x,0)&= 0.1+0.75\exp(-5x^2)
\end{aligned}
\right.
\end{equation}
In the constructed $Net_1$, we used a neural network with $8$ hidden layers, each containing $50$ neurons. In the $T \times X$ space, $8000$ points were randomly sampled as residual points for the RH loss and equation loss. Additionally, we have a total of $300$ conservation points at both $t_1=0$ and $t_2=0.4$. The data included only a small amount of initial and final-state information.

As shown in Table \ref{table:the $L^2$ error of eos1}, the $L^2$ generalization error of the closure model for the equation of state obtained by combining these five groups of training data is $0.24\%$. Under the same data conditions, the $L^2$ generalization error of the closure model obtained using data-driven methods is $0.0092$. This indicates that the method proposed in this paper indeed performs better.
\begin{table}[!htb]
    \centering
     \renewcommand{\tablename}{Table}
    \caption{The $L^2$ error of the closure model for the ideal equation of state}
      \begin{threeparttable}
      \begin{tabular}{ccc}
          \toprule
           &Generalization error of the new method&Generalization error of data-driven \\
          \midrule
          $L^2$ error & 0.24\%&0.92\% \\
          \bottomrule
      \end{tabular}
    \end{threeparttable}
\label{table:the $L^2$ error of eos1}
\end{table}

Now, we combine the constructed closure model of the equation of state with traditional numerical schemes to solve flow field problems with corresponding initial and boundary value conditions. By comparing the obtained results with the reference solution and the numerical solutions computed by the target model under the same experimental settings, we evaluate the feasibility and accuracy of ours closure model in practical applications.

\subsection{Test Case 1}
First, a 1D smooth periodic test case from \cite{1Dtest_case1} is used for verification. This test case is commonly employed to evaluate the accuracy of traditional numerical schemes, and here we apply it to assess the accuracy of the closure model. The initial conditions are set as follows:
\begin{equation}
\left\{
\begin{aligned}
u(x,0)  &= \sin(\frac{2\pi x}{L}),\quad L=5,\\
\rho(x,0) &= (1+\frac{(\gamma-1)u(x,0)}{2c})^\frac{2}{\gamma-1}, c=\frac{\sqrt{\gamma}}{\epsilon}, \gamma=2, \epsilon=0.3,\\
p(x,0)&= \rho(x,0)^\gamma
\end{aligned}
\right.
\end{equation}
The computational domain is $\Omega = [-2.5, 2.5]$, and the final time is $T = 0.3$. The number of grid points is chosen as $N = 1000$, and the fifth-order WENO-Z scheme is used for computation. The reference solution is obtained using a fifth-order WENO-Z spatial discretization and third-order Runge-Kutta (RK) time integration with $N = 25000$ grid points.

\begin{figure}[!htb]
  \centering
  \subfloat[Comparison of the numerical solution and absolute error of density obtained by two models at $t = 0.3$]{
      \centering  
      \includegraphics[height=6.5cm]{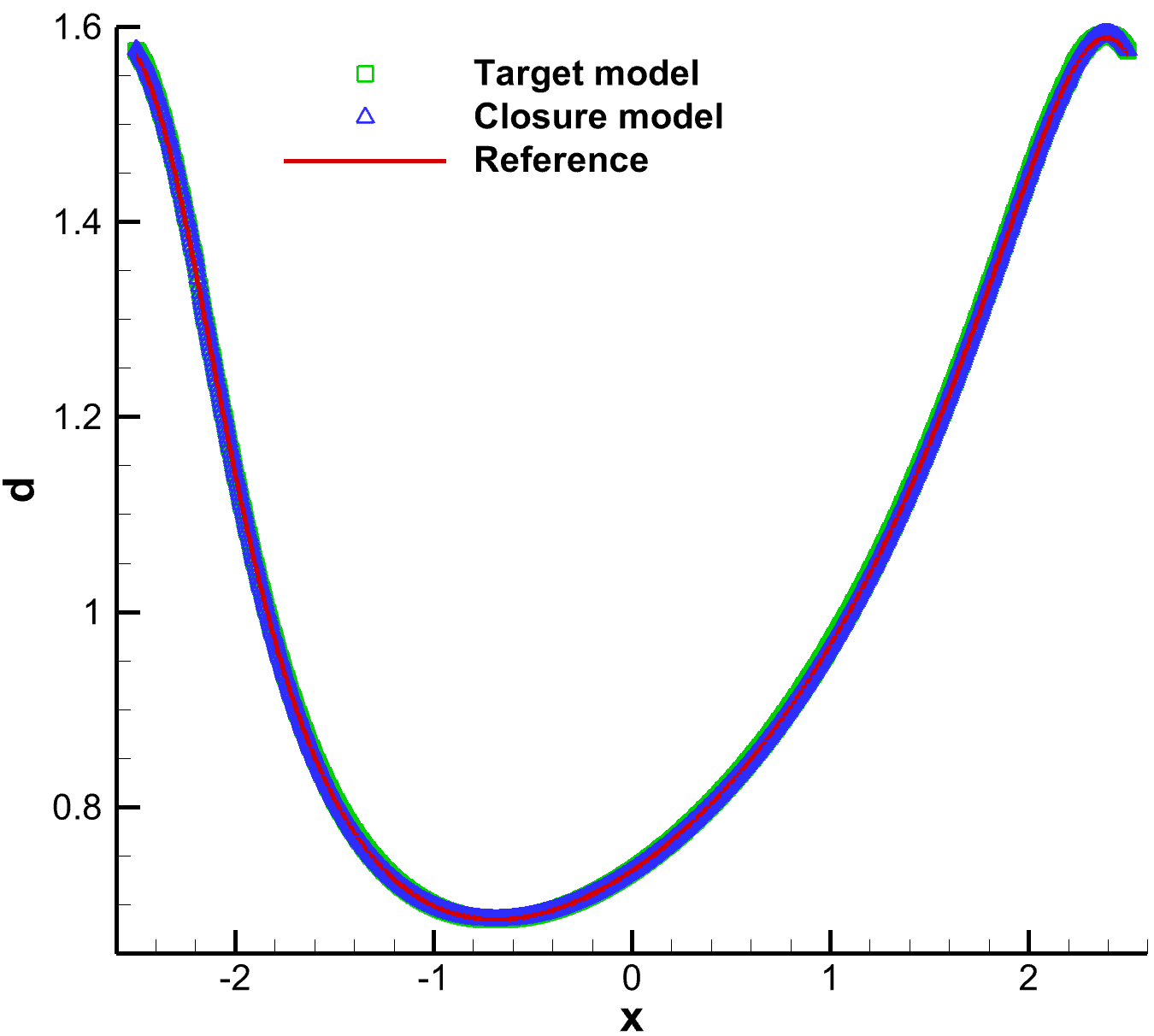}
      \label{fig:the numerical solutions of density} 
      \includegraphics[height=6.5cm]{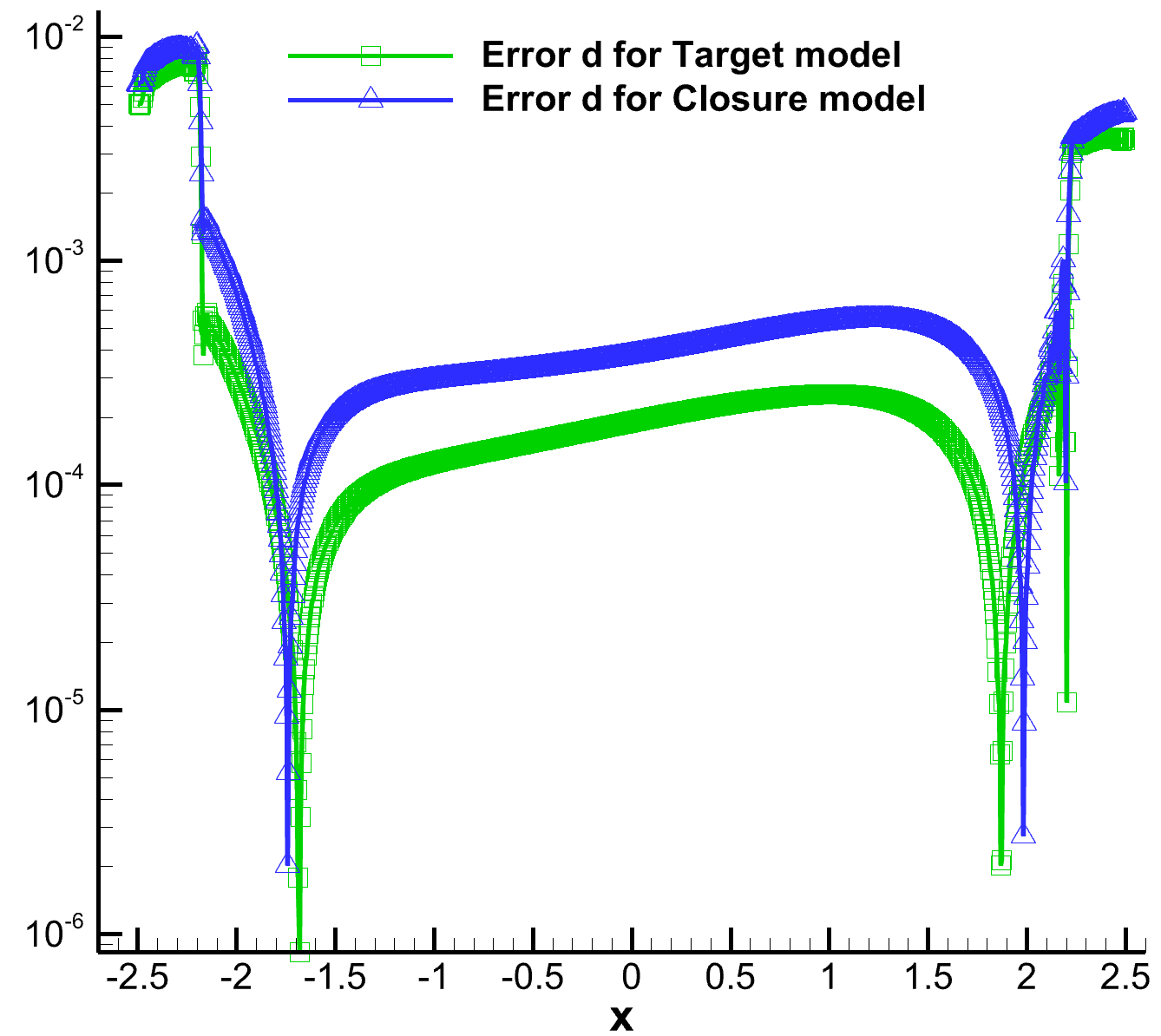}
      \label{fig:the absolute error of density}
  }
  \quad
  \subfloat[Comparison of the numerical solution and absolute error of velocity obtained by two models at $t = 0.3$]{
      \centering
       \includegraphics[height=6.5cm]{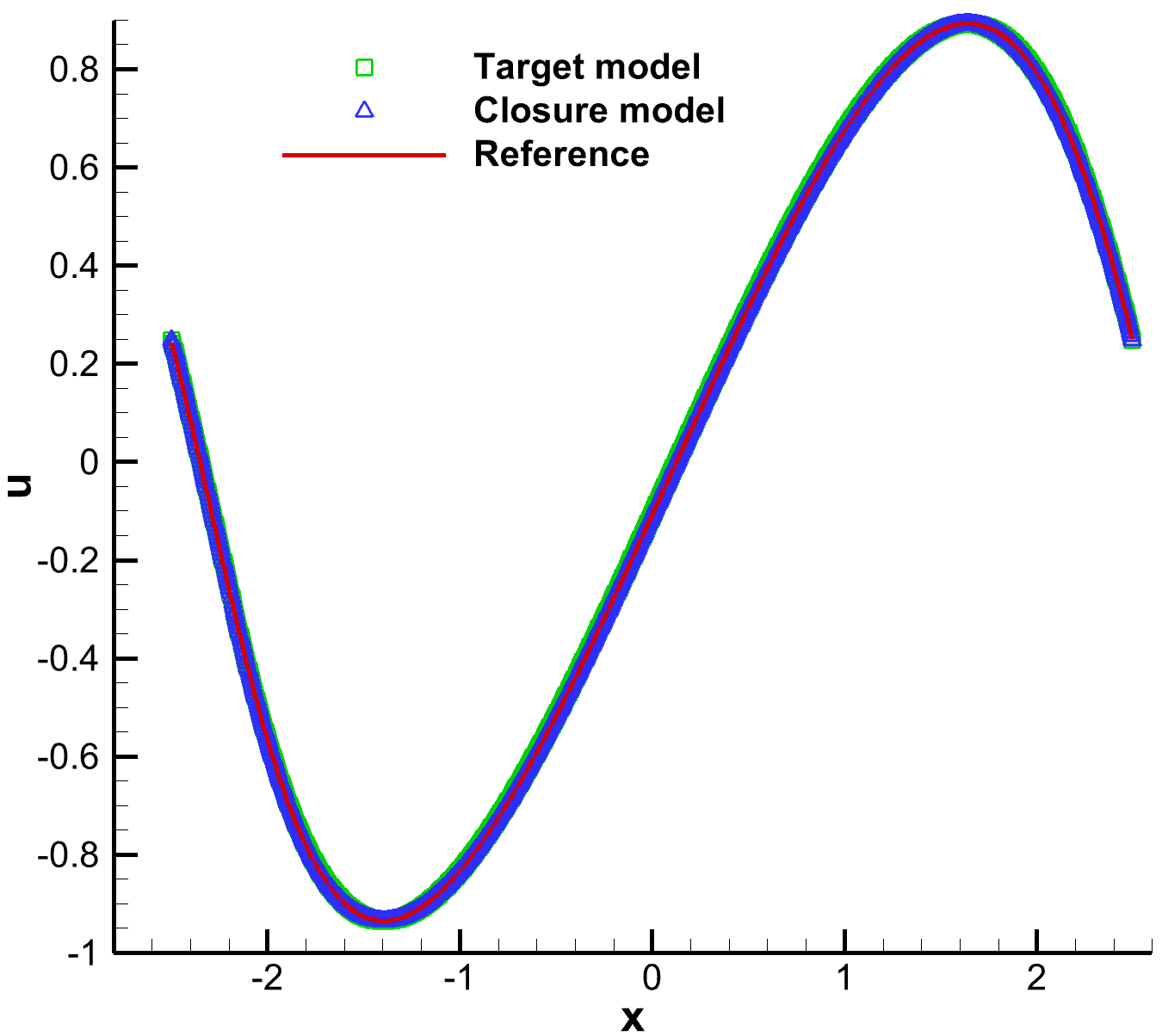}
      \label{fig:the numerical solutions of velocity}
      \includegraphics[height=6.5cm]{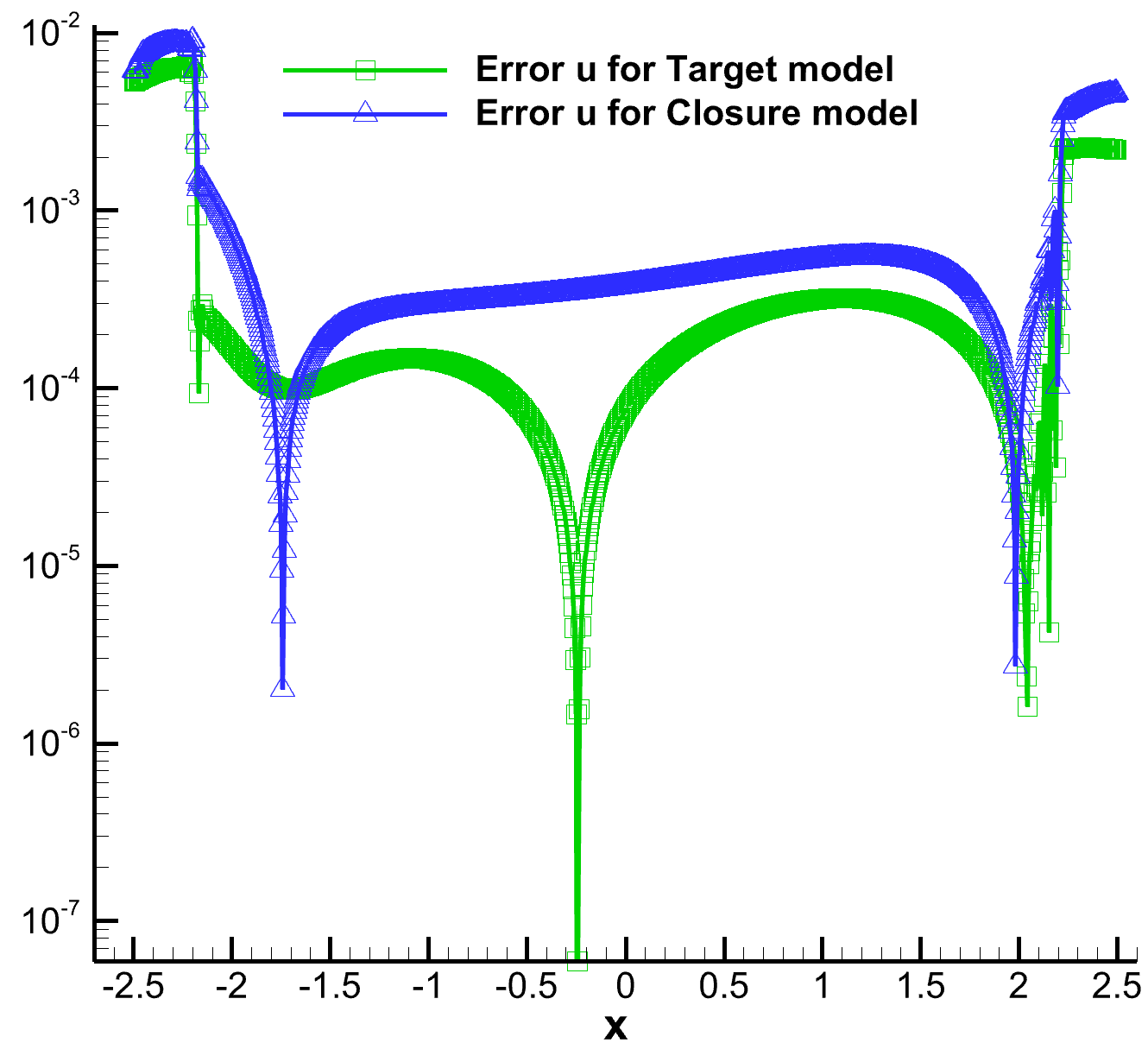}
      \label{fig:the absolute error of velocity}
  }
  \quad
  \subfloat[Comparison of the numerical solution and absolute error of pressure obtained by the models at $t = 0.3$]{
      \centering  
      \includegraphics[height=6.5cm]{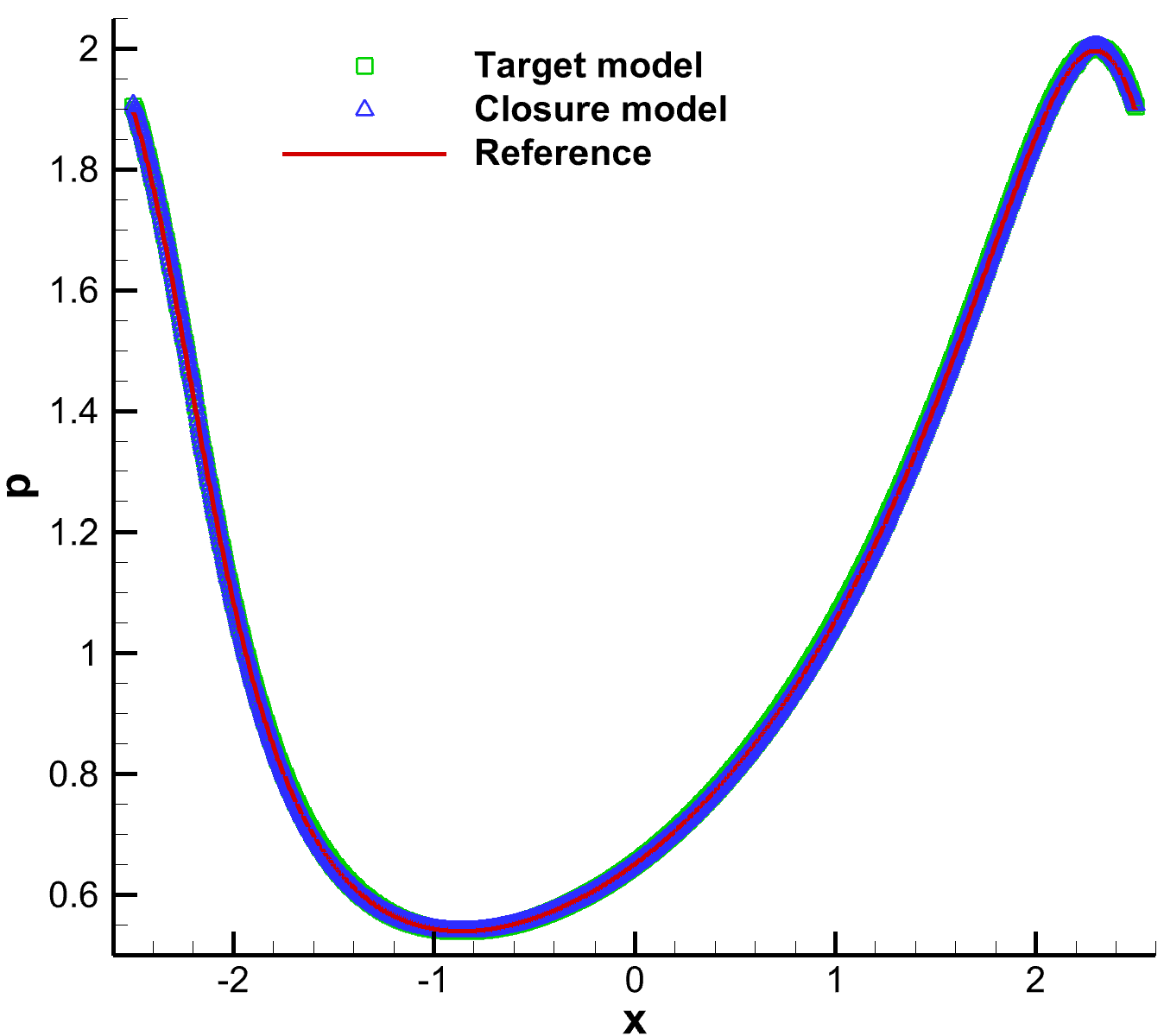}
      \label{fig:the numerical solutions of pressure}   
      \includegraphics[height=6.5cm]{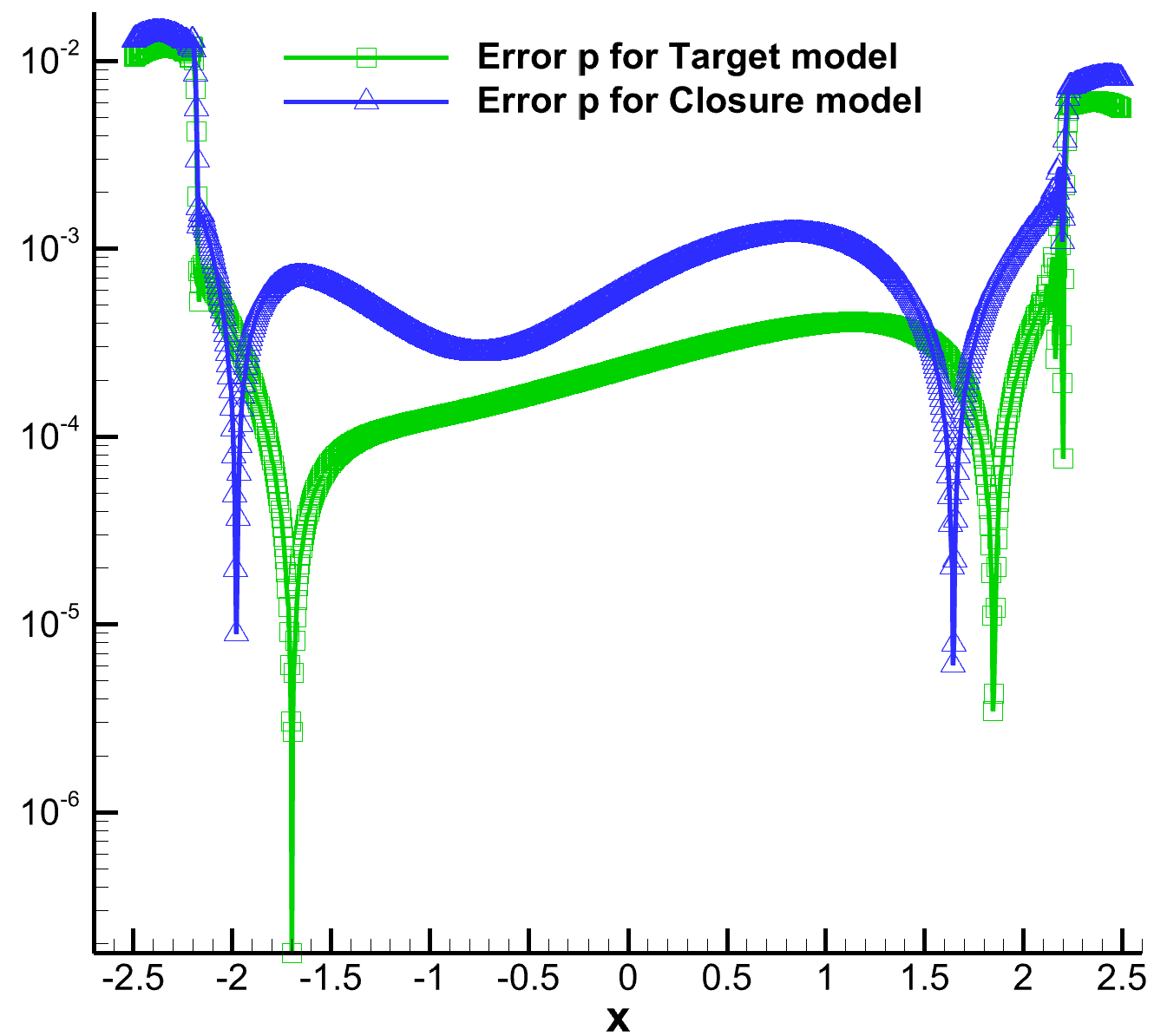}
      \label{fig:the absolute error of pressure}
  }
  \renewcommand{\figurename}{Fig.}
  \caption{At $t = 0.3$, the comparison of the predicted solutions for each physical quantity and their corresponding absolute errors between the closure model of the ideal equation of state and the target model in the 1D smooth periodic test case}
  \label{fig:Comparison image of the predicted solution of eos1 test example 1}
\end{figure}

Figure \ref{fig:Comparison image of the predicted solution of eos1 test example 1} shows the comparison of numerical solutions for $\rho$, $u$, and $p$ at time $t = 0.3$ under two different models. From the error analysis graph, it is evident that the closure model constructed using this method, along with the coupled algorithm, produces numerical results that are in strong agreement with the target model when addressing new problems. This demonstrates that the closure model, developed using sparse data, exhibits significant generalization and reliability, making it an effective alternative to traditional models for solving practical problems. Moreover, the coupled algorithm's strengths are clearly evident in the test cases. By leveraging the stability inherent in traditional numerical methods, the algorithm mitigates risks of numerical divergence or instability, which are common challenges in simulating complex flow fields.

\subsection{Test Case 2}
Similarly, we use the Sod problem to test the model, with the following initial setup:
\begin{equation}
(\rho, u, p) = 
\begin{cases} 
(1, 0, 1), & \text{if }\quad  x \leq 0.5, \\
(0.125, 0, 0.1), & \text{if}\quad 0.5 < x 
\end{cases}
\end{equation}
The computational domain is $\Omega = [0, 1]$, and the final time is $T = 0.2$. The number of grid points is chosen as $N = 200$, and the fifth-order WENO-Z scheme is used for computation. The reference solution is the exact solution obtained by Riemann solver.

\begin{figure}[!htb]
  \centering
  \includegraphics[height=7cm]{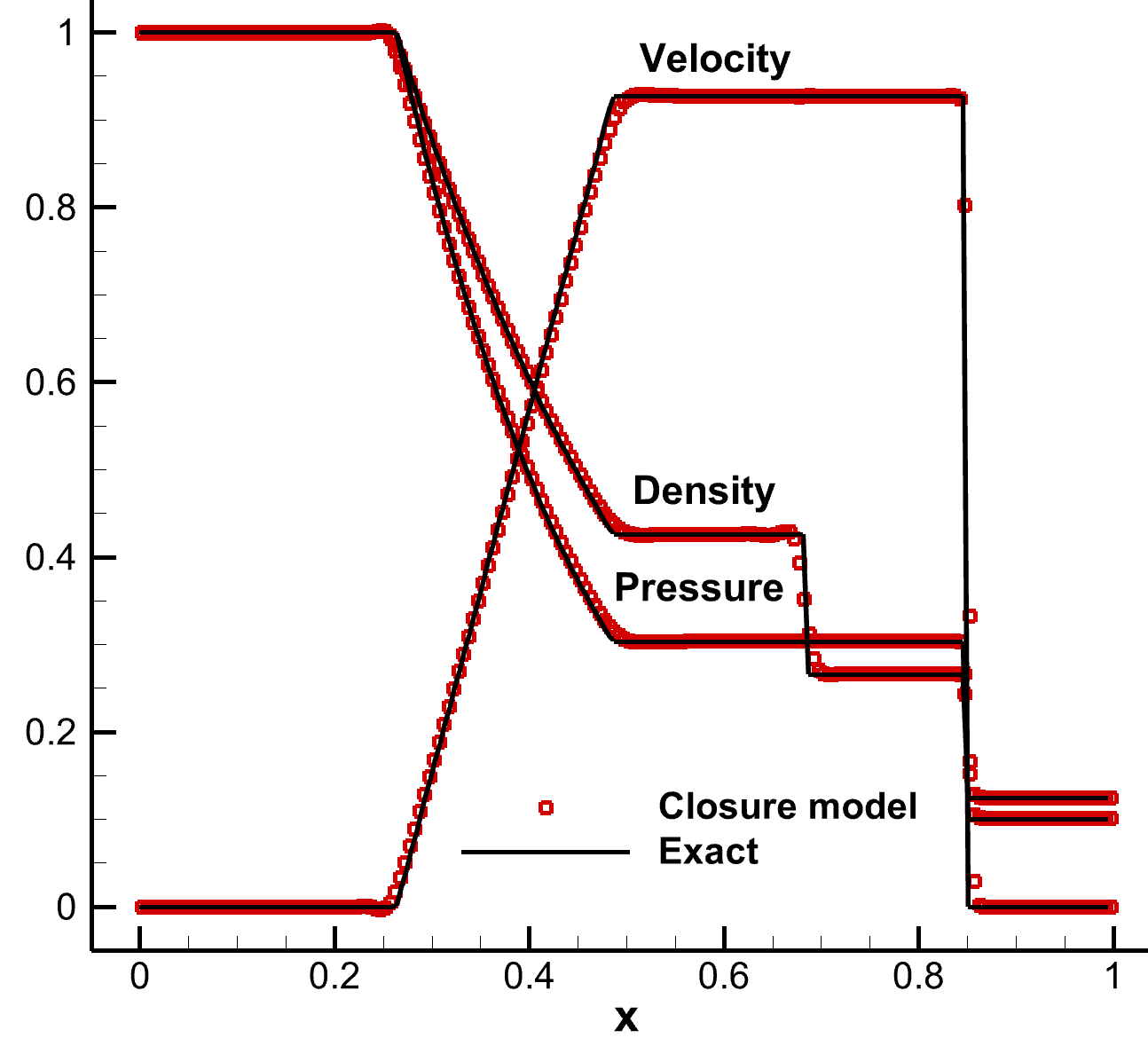}  
  \includegraphics[height=7cm]{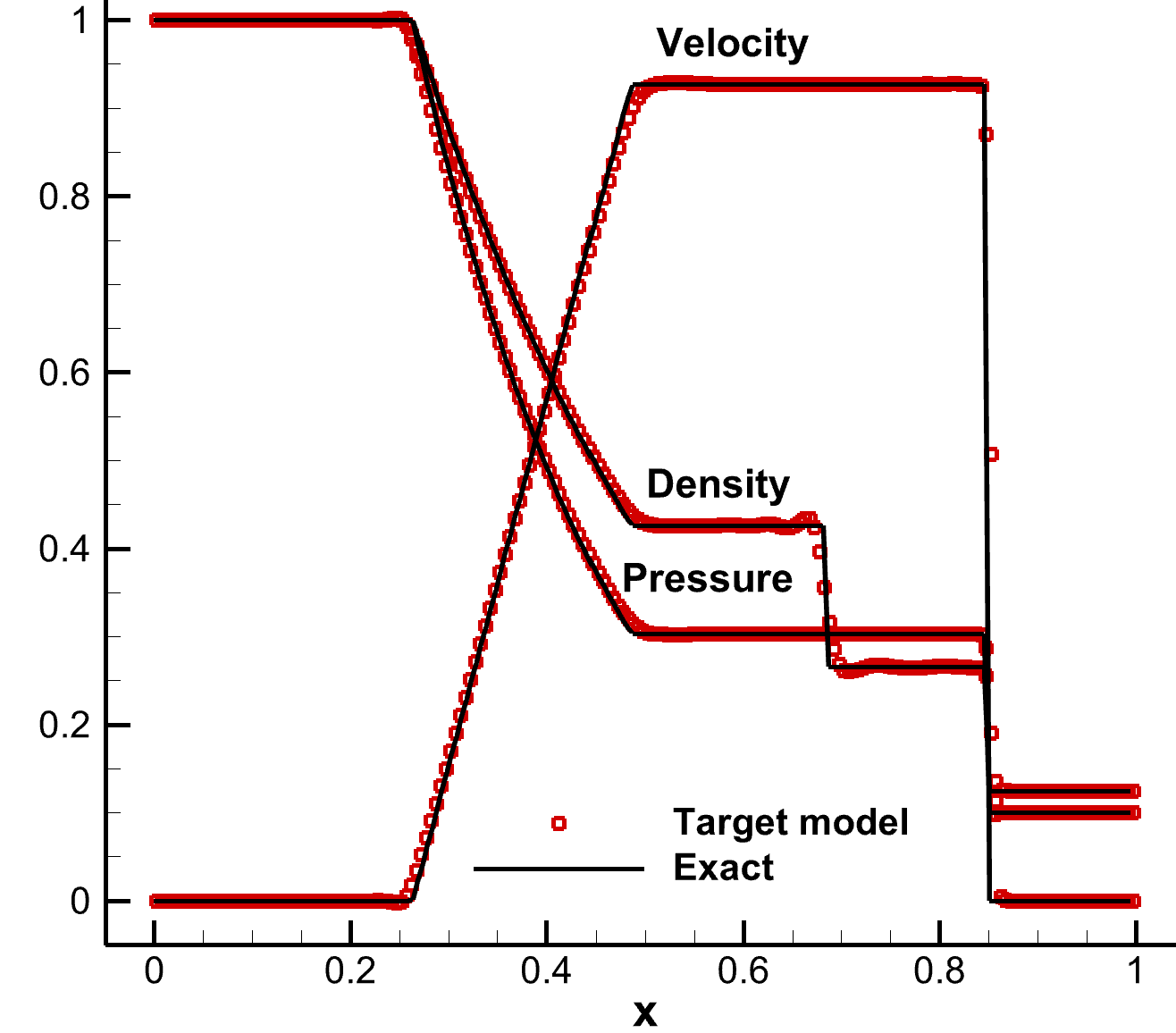}
  \renewcommand{\figurename}{Fig.}
  \caption{At $t = 0.2$, the comparison of predicted solutions for each physical quantity between the closure model and the target model using the ideal equation of state in the 1D Sod problem}
  \label{fig:Comparison chart of Sod problem}
\end{figure}

Figure \ref{fig:Comparison chart of Sod problem} shows the comparison of numerical solutions for $\rho$, $u$, and $p$ at $t = 0.2$ under two models. From the figures, it can be observed that the numerical results obtained using the closure model and coupling algorithm developed in this method show high consistency with those of the target model when applied to new problems. This observation highlights that, when tackling discontinuity problems involving shock waves, the coupled algorithm adeptly mitigates numerical divergence and oscillations caused by such shocks or other types of discontinuities. Furthermore, the closure model's robust generalization capabilities ensure the overall precision of the numerical solutions, thereby validating the efficacy of the proposed methodology.

\subsection{Test Case 3}
Similarly, to verify the model's applicability in high-dimensional problems, we apply it to a 2D periodic case for testing, with the initial conditions as follows:
\begin{equation}
\left\{
\begin{aligned}
\rho(x,0) &= 0.5,\\
u(x,0)  &= 0.3\sin(2\pi x),\\
v(x,0)  &= 0.3\sin(2\pi y),\\
p(x,0)&= 0.5
\end{aligned}
\right.
\end{equation}
The computational domain is $\Omega = [0, 1] \times [0, 1]$, and the final time is $T = 0.1$. The grid points are chosen as $N \times N = 400 \times 400$, and the fifth-order WENO-Z scheme is used for computation.

\begin{figure}[!htb]
  \centering
  \subfloat[Density (numerical solution given by the closure model)]{
      \centering
      \includegraphics[height=6.5cm]{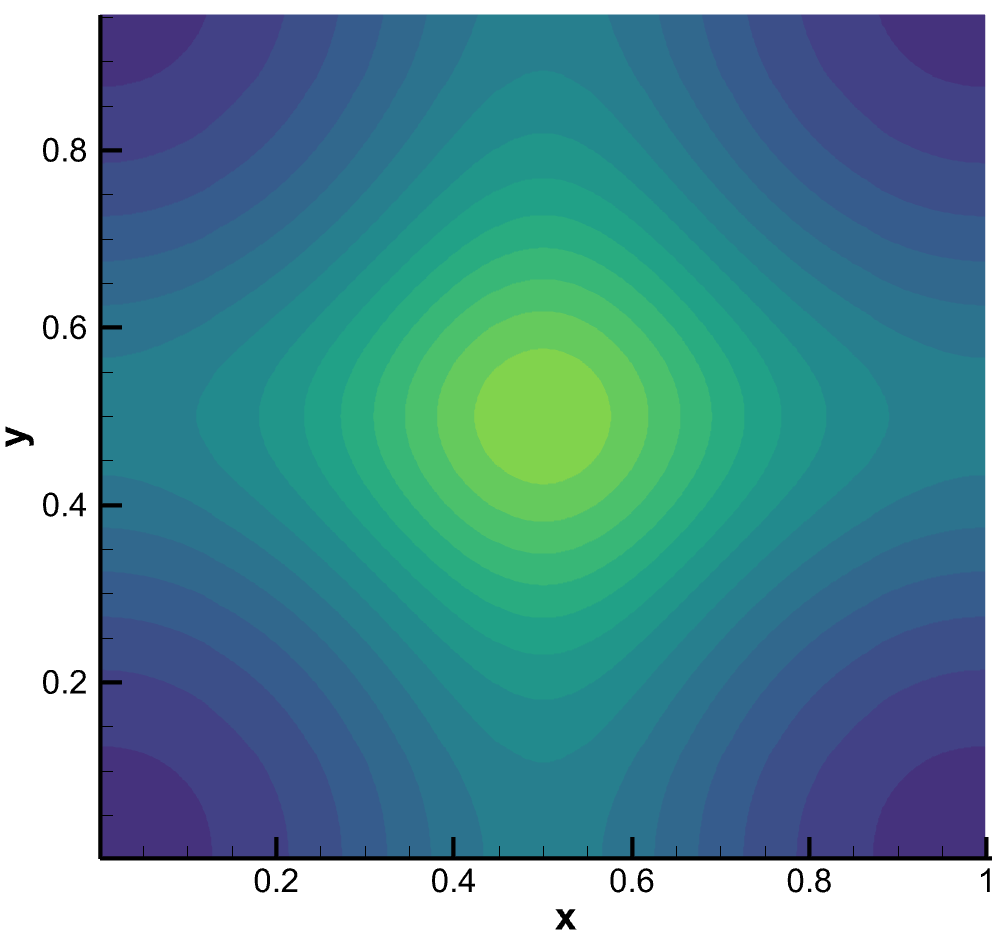}
  }
  \quad
  \subfloat[Density (numerical solution given by the target model)]{
      \centering
       \includegraphics[height=6.5cm]{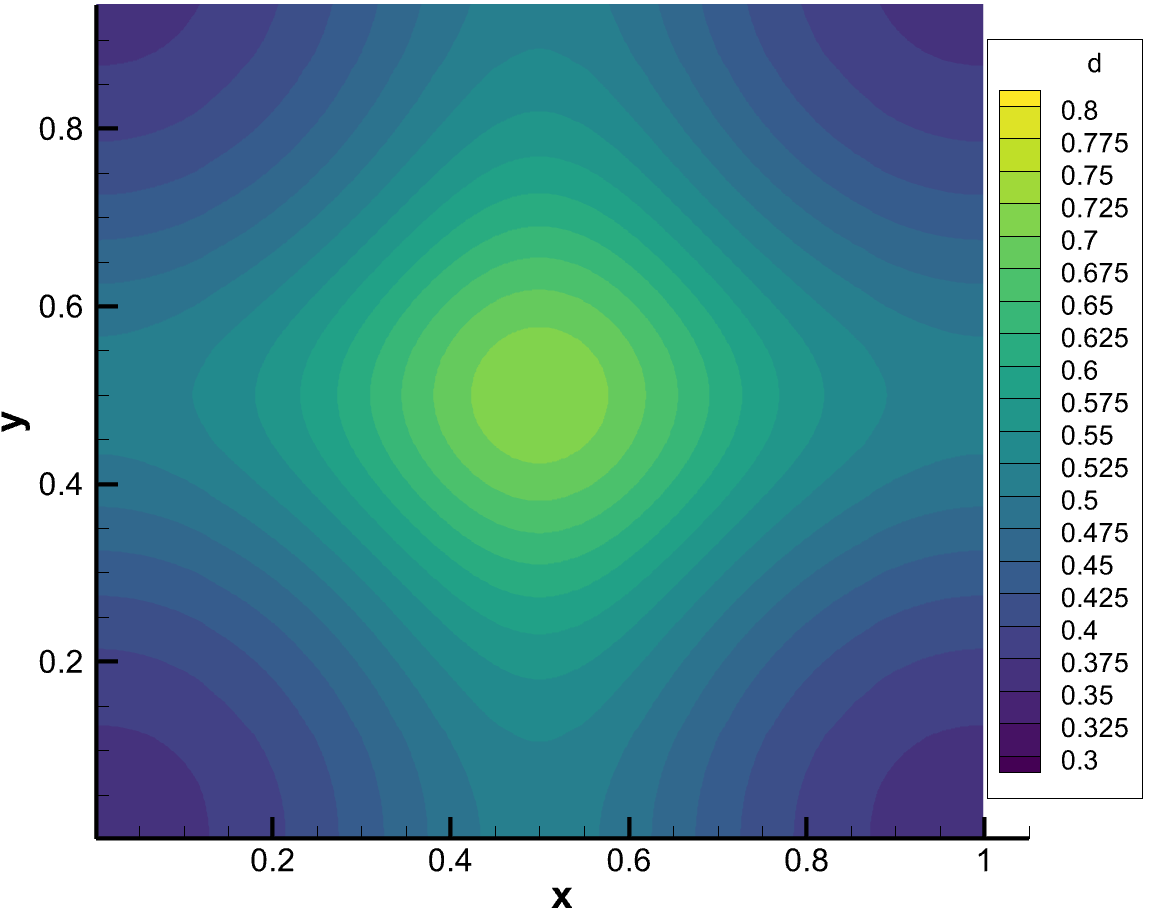}
  }
  \quad
  \subfloat[Velocity u (numerical solution given by the closure model)]{
      \centering
       \includegraphics[height=6.5cm]{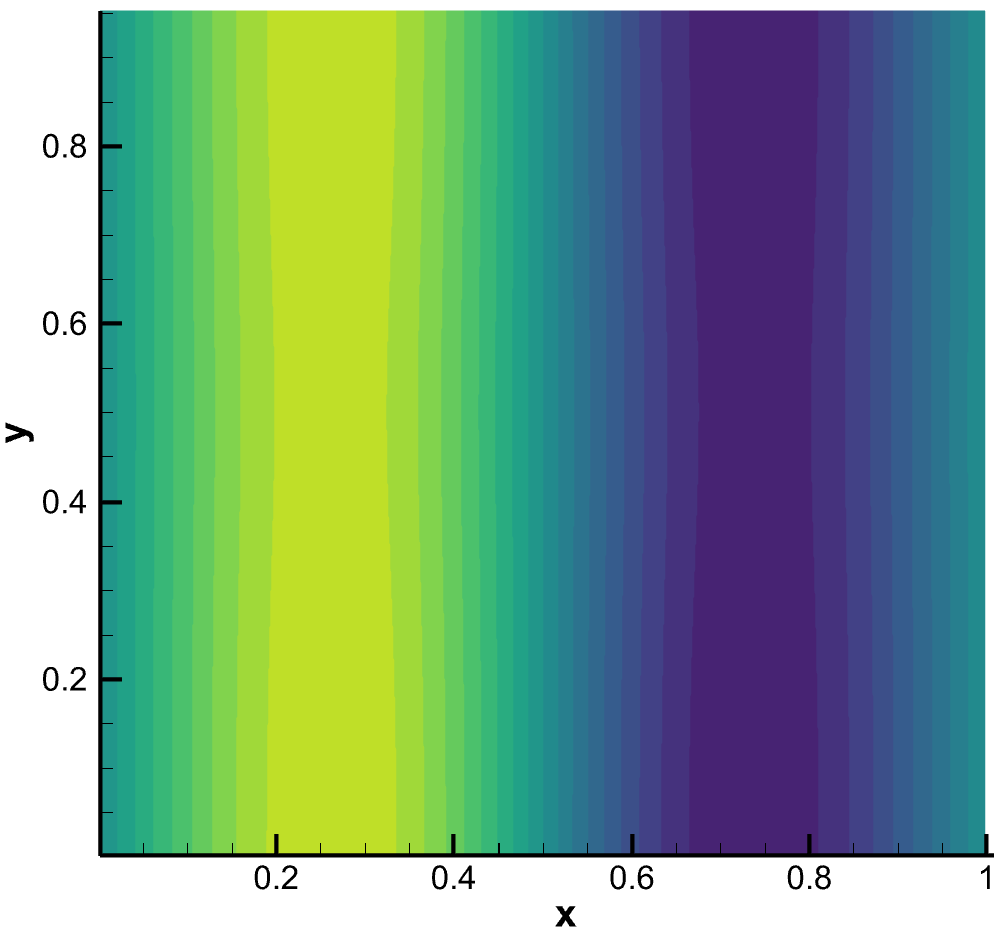}
  }
  \quad
  \subfloat[Velocity u (numerical solution given by the target model)]{
      \centering
      \includegraphics[height=6.5cm]{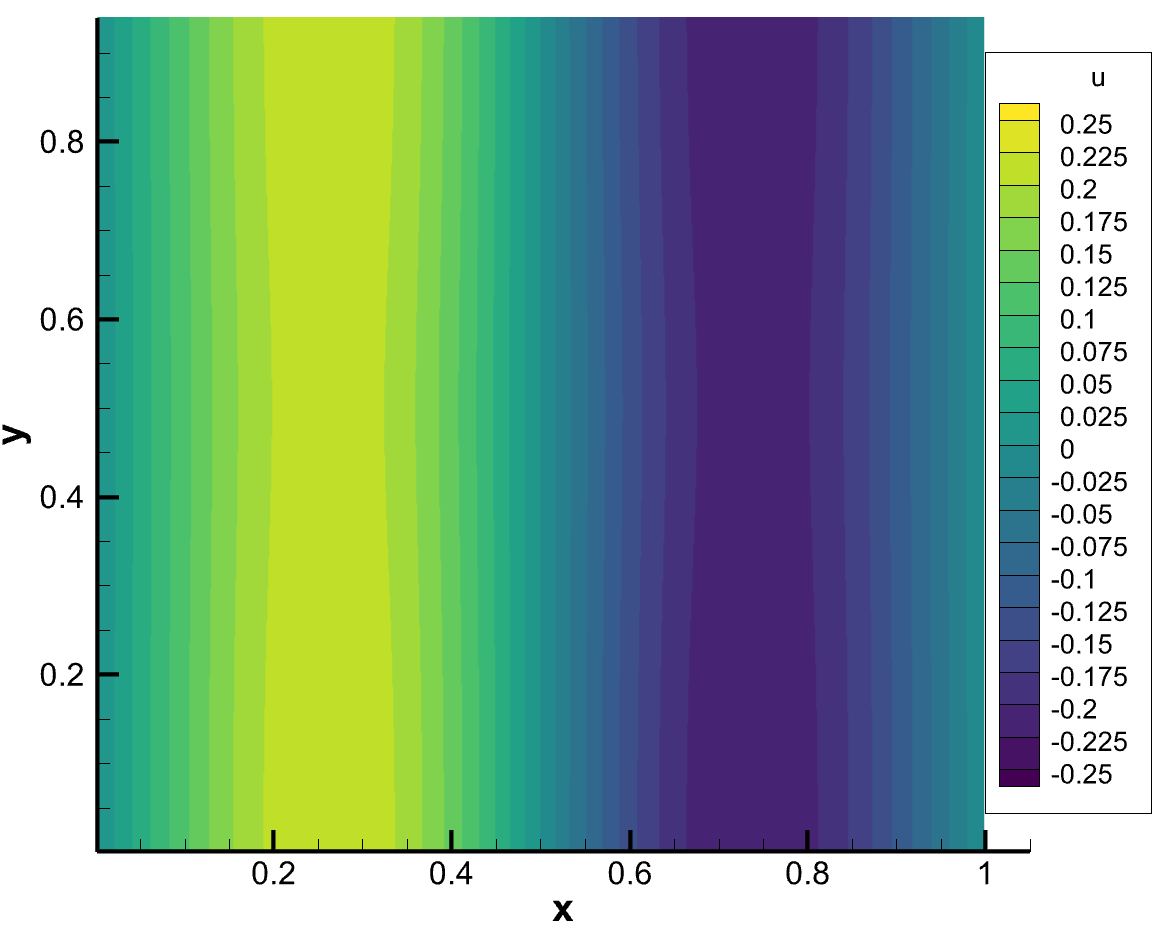}
  }
  \renewcommand{\figurename}{Fig.}
  \caption{At $t = 0.1$, the comparison of the predicted solutions for density and velocity u between the closure model of the ideal equation of state and the target model in the 2D smooth periodic test case(part 1)}
  \label{fig:2D smooth the predicted solutions for rho and u}
\end{figure}

\begin{figure}[!htb]
  \centering
  \subfloat[Velocity v (numerical solution given by the closure model)]{
      \centering
      \includegraphics[height=6.5cm]{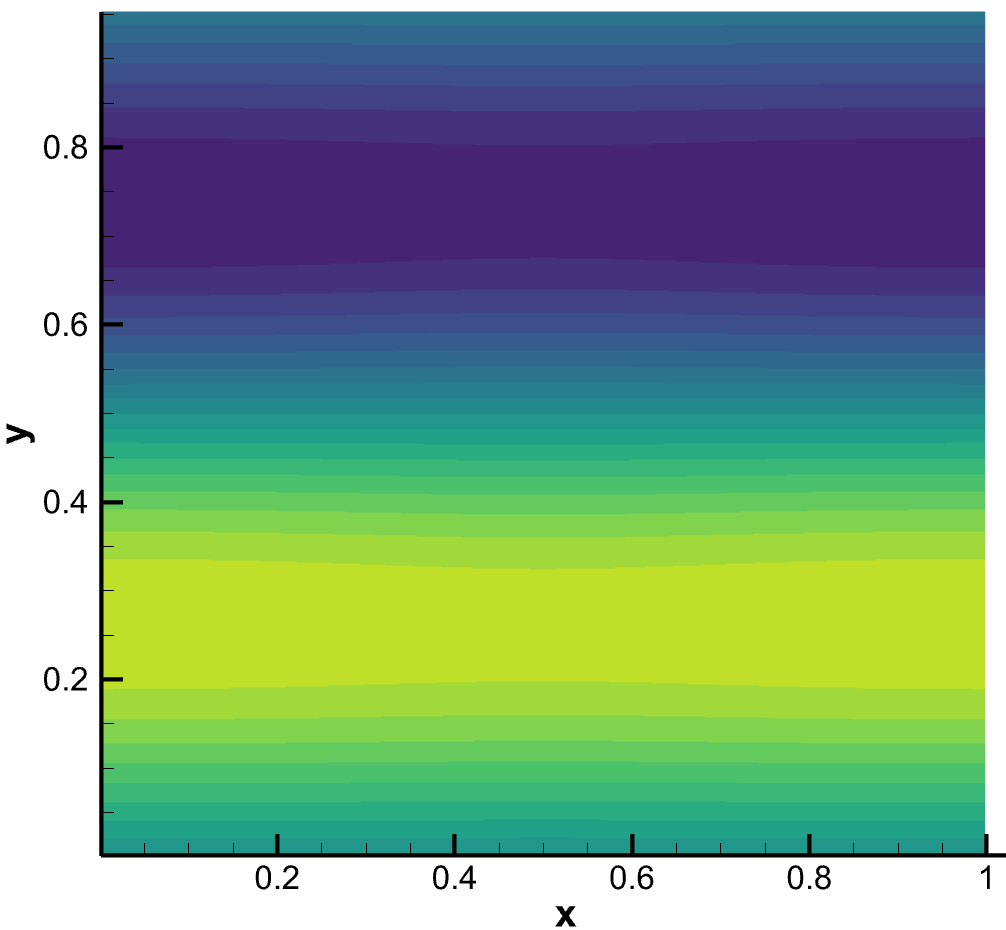}
  }
  \quad
  \subfloat[Velocity v (numerical solution given by the target model)]{
      \centering
       \includegraphics[height=6.5cm]{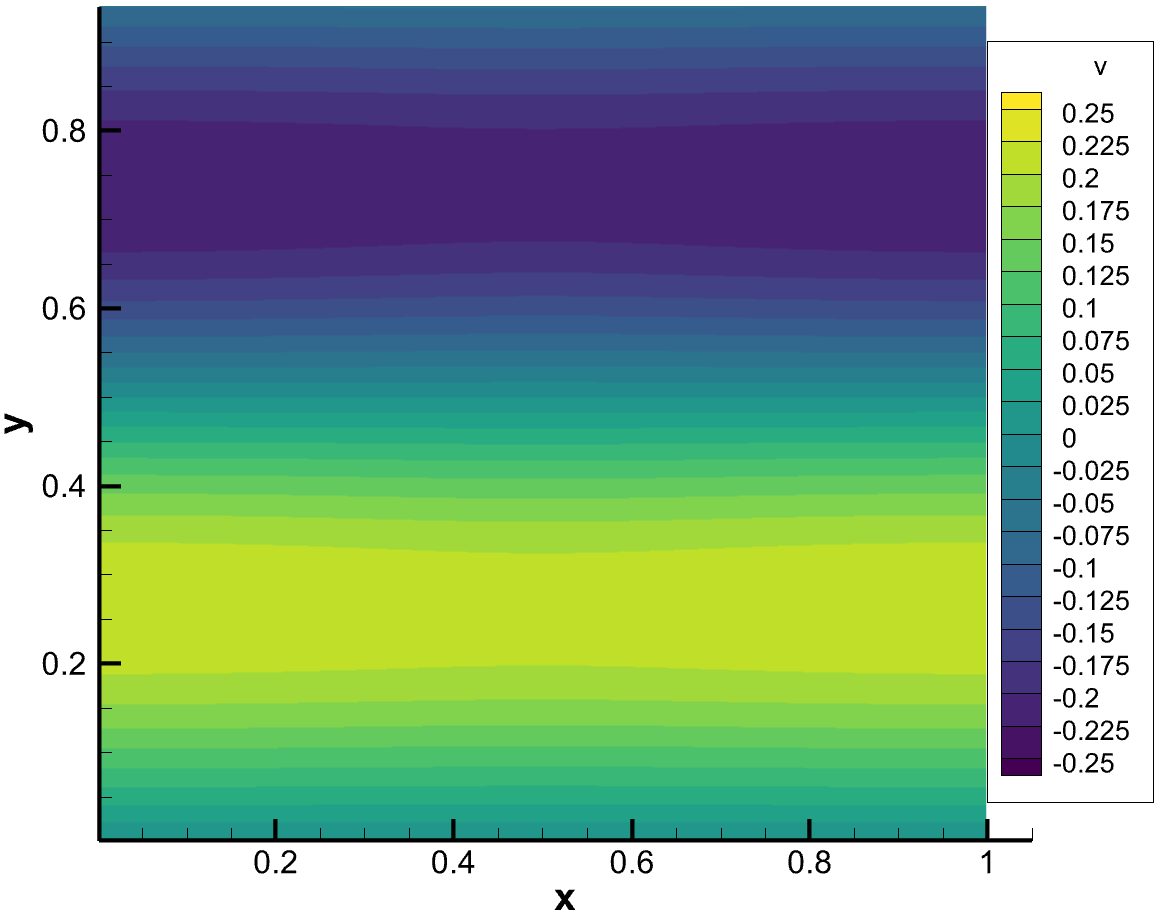}
  }
  \quad
  \subfloat[Pressure (numerical solution given by the closure model)]{
      \centering
       \includegraphics[height=6.5cm]{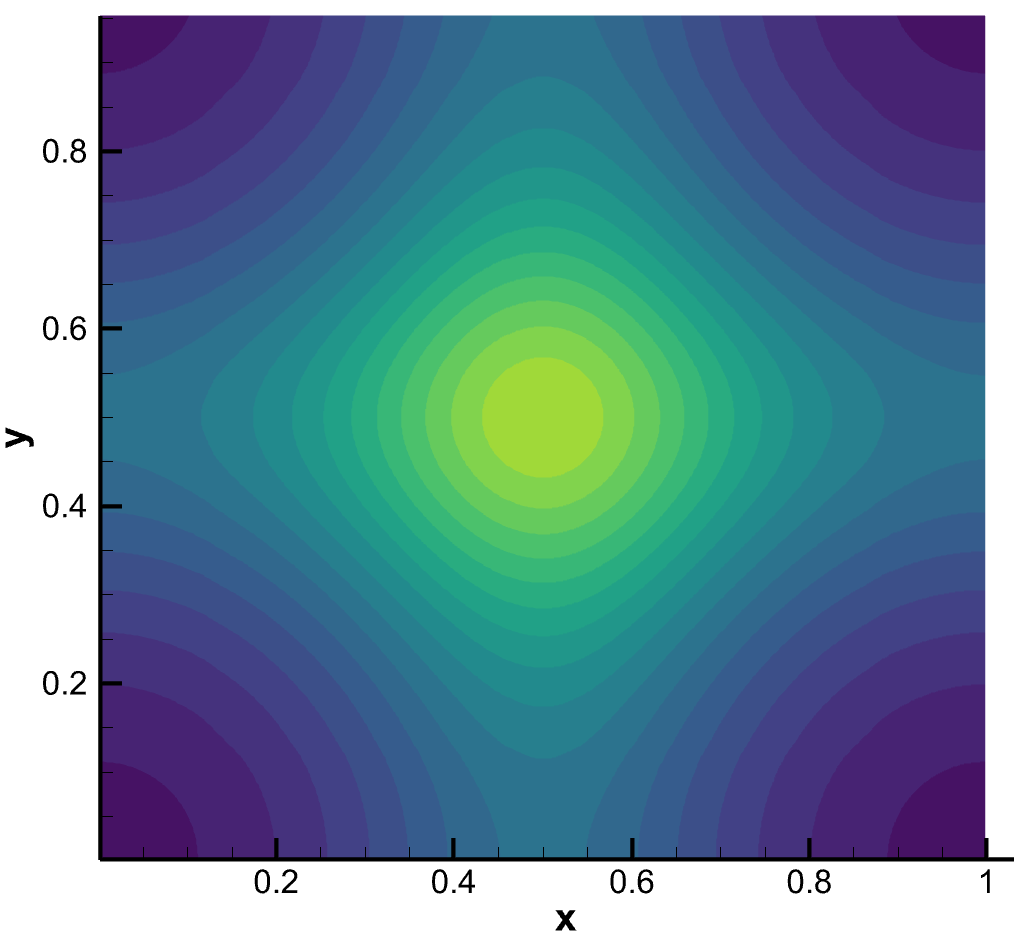}
  }
  \quad
  \subfloat[Pressure (numerical solution given by the target model)]{
      \centering
      \includegraphics[height=6.5cm]{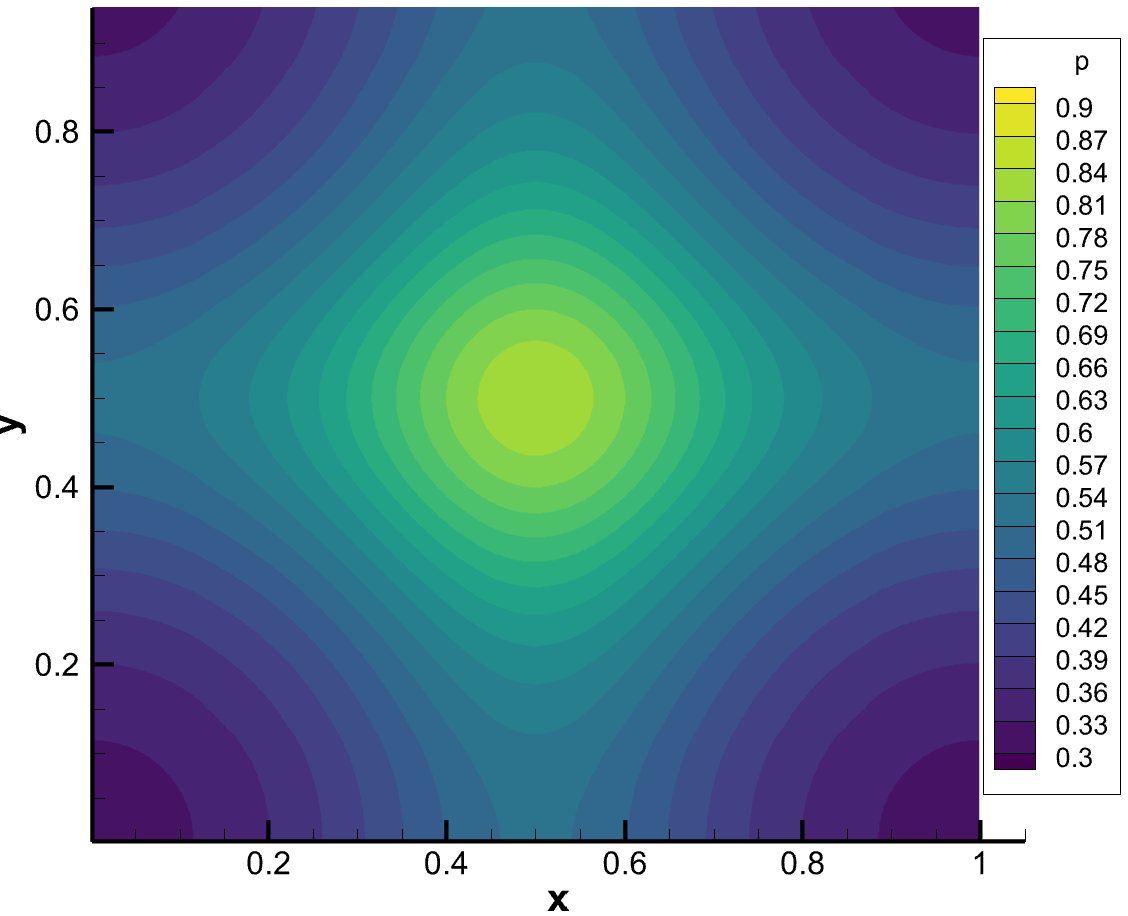}
  }
  \renewcommand{\figurename}{Fig.}
  \caption{At $t = 0.1$, the comparison of the predicted solutions for velocity v and pressure between the closure model of the ideal equation of state and the target model in the 2D smooth periodic test case(part 2)}
  \label{fig:2D smooth the predicted solutions for velocity v and pressure}
\end{figure}

Figures \ref{fig:2D smooth the predicted solutions for rho and u} and \ref{fig:2D smooth the predicted solutions for velocity v and pressure} show the comparison of numerical solutions for $\rho$, $u$, $v$, and $p$ at $t = 0.1$ under the two models. As can be seen from the figure, the numerical results obtained using the closure model and coupling algorithm developed in this method show high consistency with those of the target model when solving new problems. This demonstrates that, even in the context of two-dimensional problems, the closure model and coupling algorithm are capable of generating numerical solutions that closely match those of the target model, thereby showcasing remarkable accuracy and stability. Such findings further affirm the proposed method's efficacy and practical utility.

\subsection{Test Case 4}
To further verify the applicability and accuracy of the closure model, a 2D Riemann problem with the following initial settings is introduced for testing:
\begin{equation}
(\rho, u, v, p) = 
\begin{cases} 
(0.5, -0.5, 0.35, 0.5) & \text{if }\quad 0 \leq x \leq 0.5,\quad 0 \leq y \leq 0.5, \\
(1.0, 0.5, 0.35, 0.5) & \text{if }\quad 0 \leq x \leq 0.5, \quad 0.5 < y \leq 1, \\
(1.5, -0.5, -0.35, 0.5) & \text{if }\quad 0.5 < x \leq 1, \quad 0 \leq y \leq 0.5, \\
(0.5, 0.5, -0.35, 0.5) & \text{if }\quad 0.5 \leq x \leq 1, \quad 0.5 < y \leq 1.
\end{cases}
\end{equation}
The computational domain is $\Omega = [0, 1] \times [0, 1]$, and the final time is $T = 0.2$. The grid points are chosen as $N \times N = 400 \times 400$, and the fifth-order WENO-Z scheme is used for computation.

\begin{figure}[!htb]
  \centering
  \subfloat[Density (numerical solution given by the closure model)]{
      \centering
      \includegraphics[height=6.5cm]{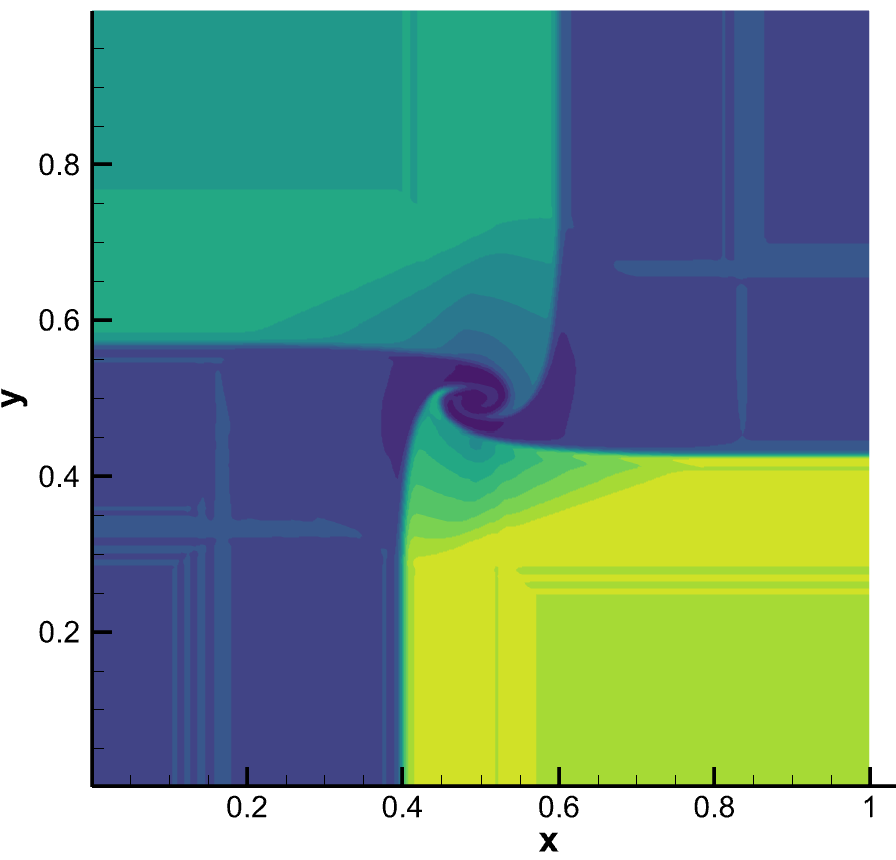}
  }
  \quad
  \subfloat[[Density (numerical solution given by the target model)]{
      \centering
       \includegraphics[height=6.5cm]{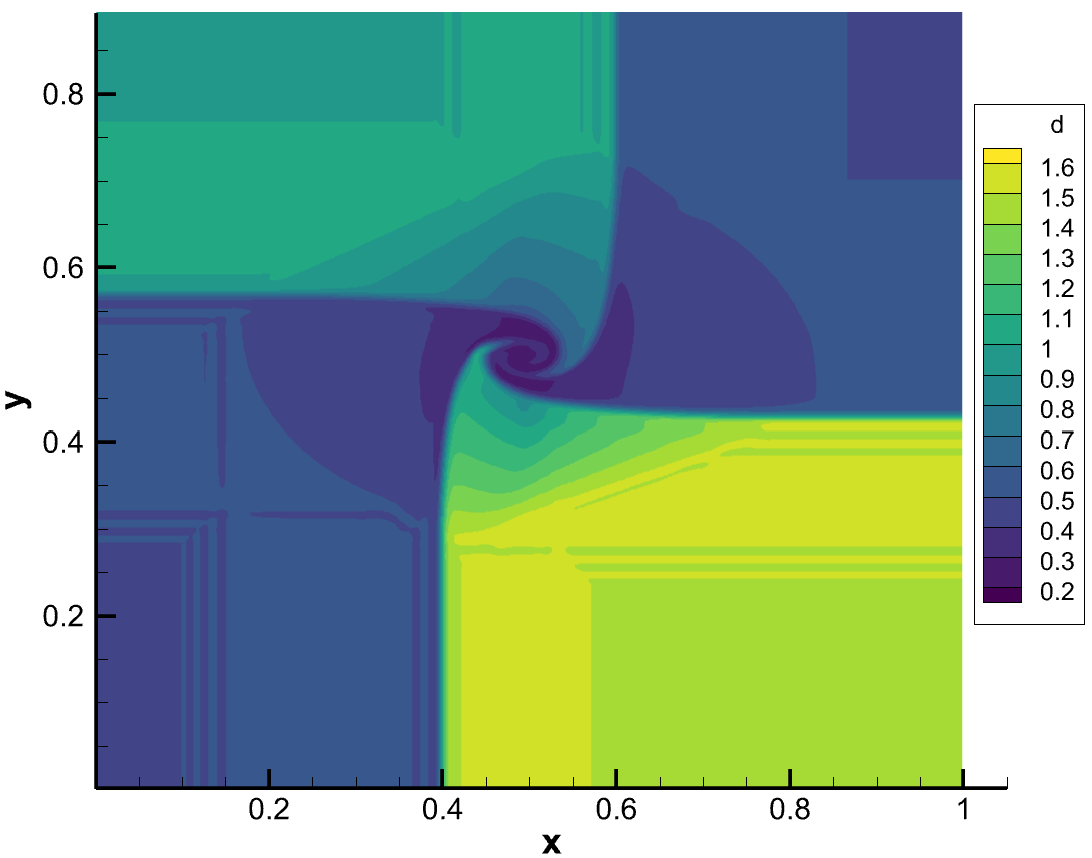}
  }
  \quad
  \subfloat[Velocity u (numerical solution given by the closure model)]{
      \centering
       \includegraphics[height=6.5cm]{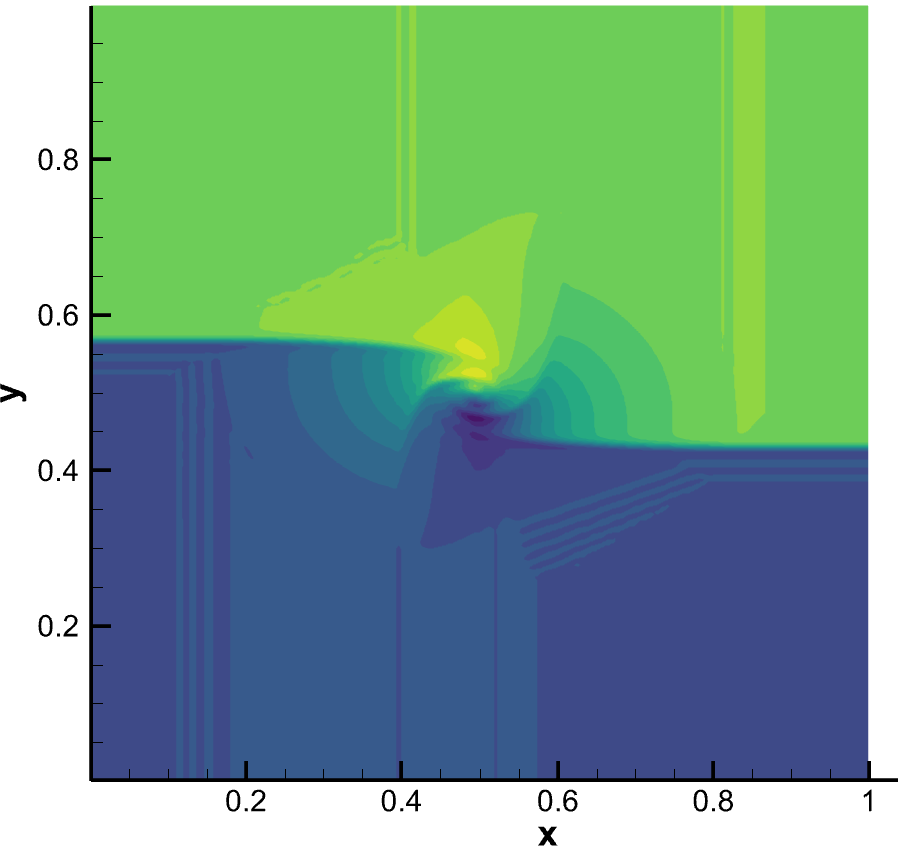}
  }
  \quad
  \subfloat[Velocity u (numerical solution given by the target model)]{
      \centering
      \includegraphics[height=6.5cm]{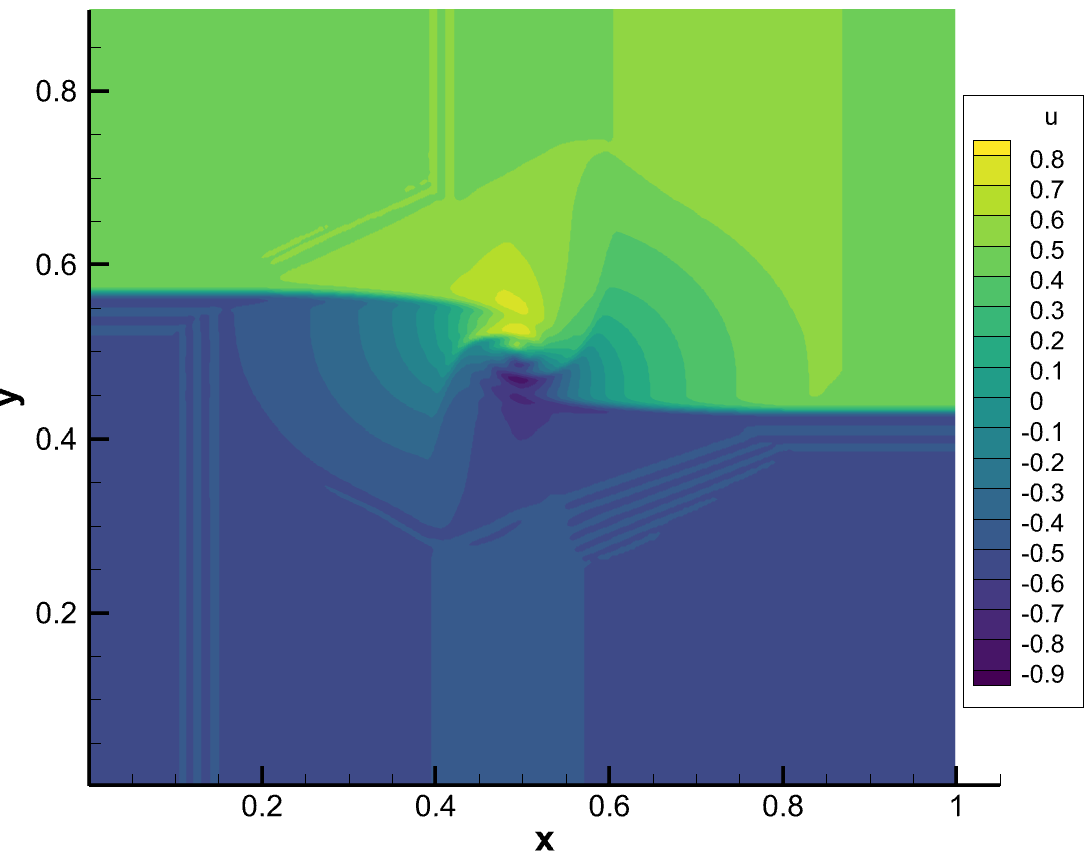}
  }
  \renewcommand{\figurename}{Fig.}
  \caption{At $t = 0.2$, the comparison of the predicted solutions for density and velocity u between the closure model of the ideal equation of state and the target model in the 2D Riemann problem(part 1)}
  \label{fig:2D Riemann the predicted solutions for density and velocity u}
\end{figure}

\begin{figure}[!htb]
  \centering
  \subfloat[Velocity v (numerical solution given by the closure model)]{
      \centering
      \includegraphics[height=6.5cm]{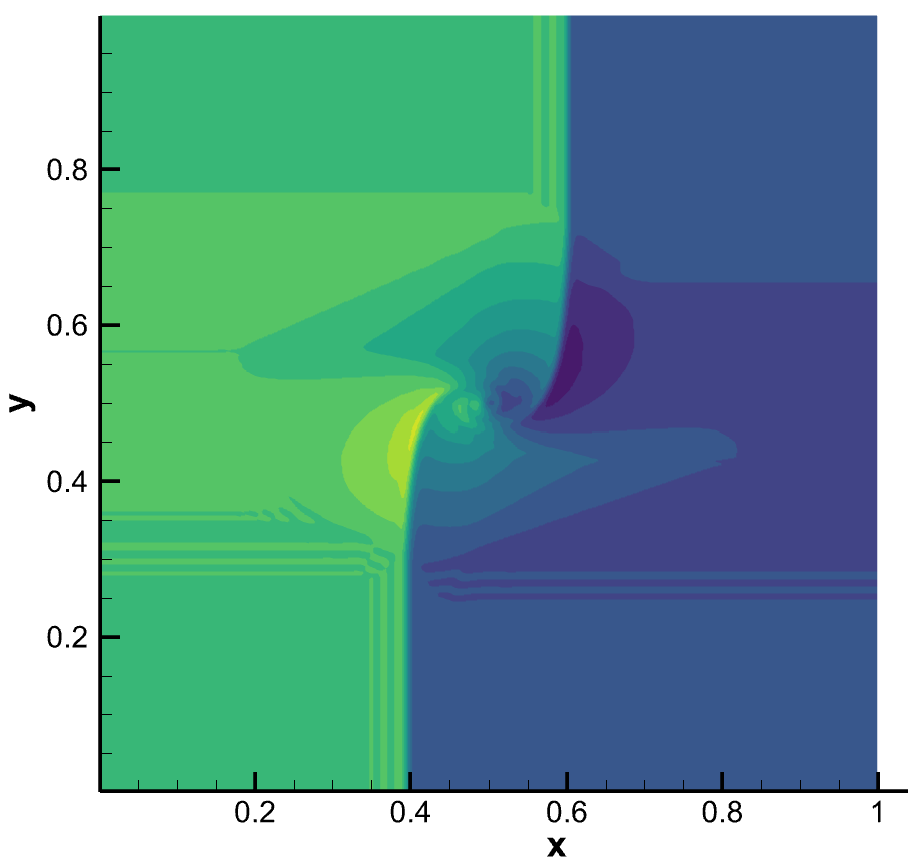}
  }
  \quad
  \subfloat[Velocity v (numerical solution given by the target model)]{
      \centering
       \includegraphics[height=6.5cm]{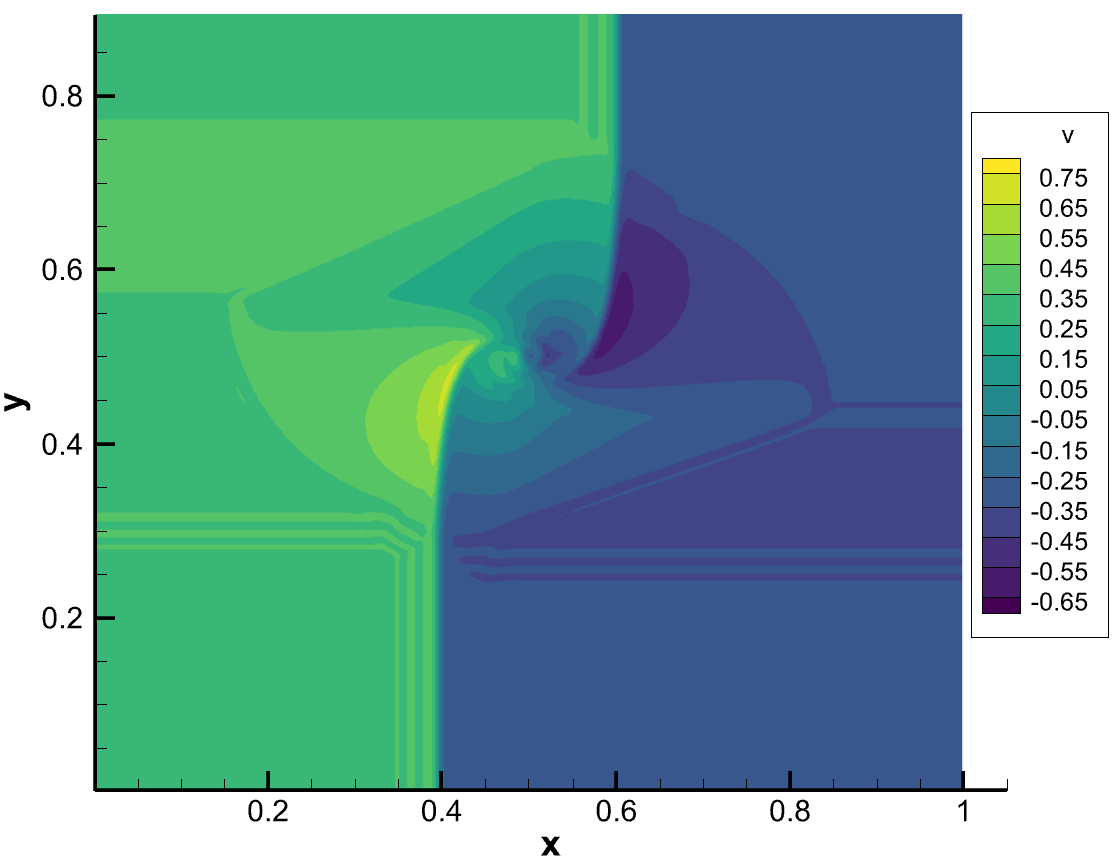}
  }
  \quad
  \subfloat[Pressure (numerical solution given by the closure model)]{
      \centering
       \includegraphics[height=6.5cm]{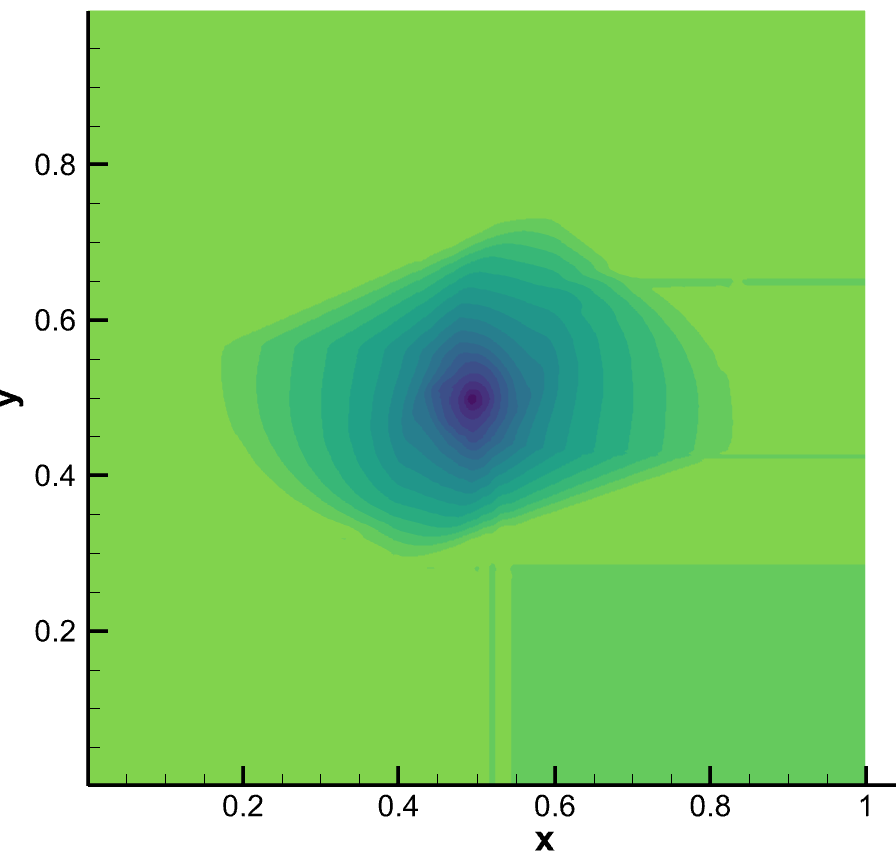}
  }
  \quad
  \subfloat[Pressure (numerical solution given by the target model)]{
      \centering
      \includegraphics[height=6.5cm]{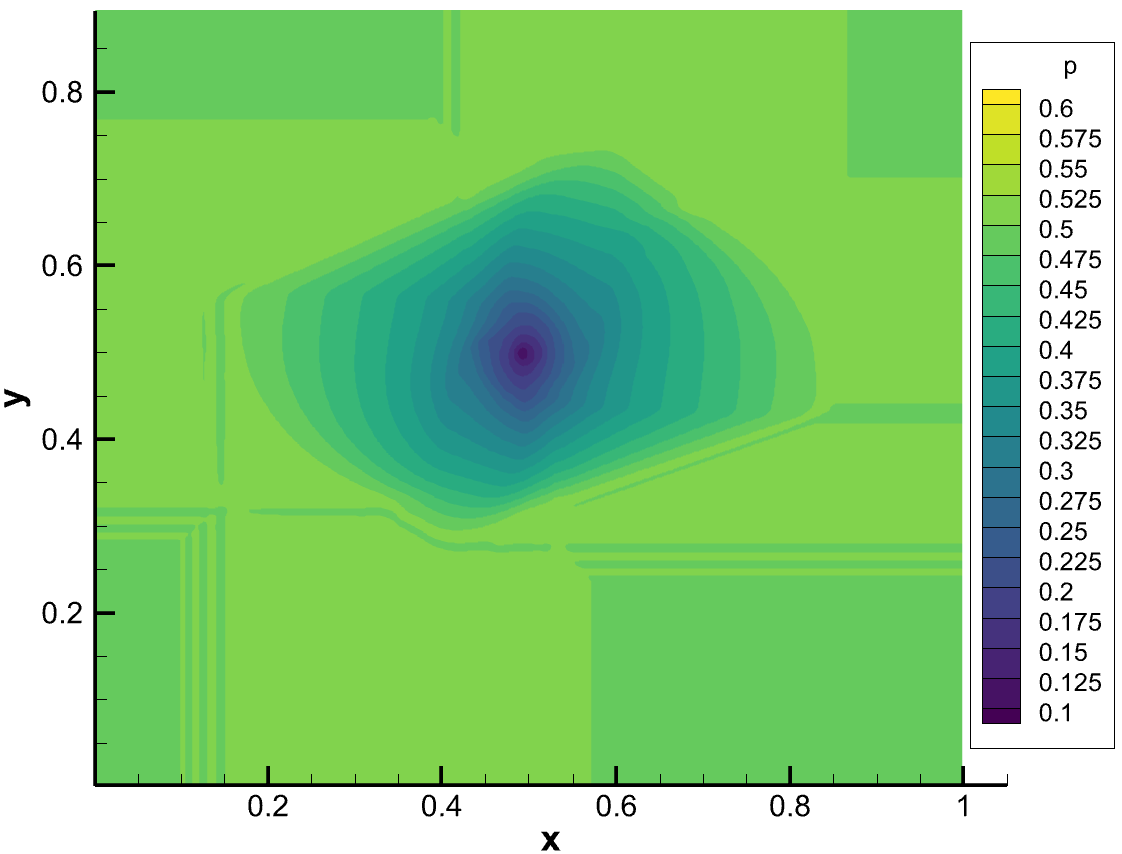}
  }
  \renewcommand{\figurename}{Fig.}
  \caption{At $t = 0.2$, the comparison of the predicted solutions for velocity v and pressure between the closure model of the ideal equation of state and the target model in the 2D Riemann problem(part 2)}
  \label{fig:2D Riemann the predicted solutions for velocity v and pressure}
\end{figure}

Figures \ref{fig:2D Riemann the predicted solutions for density and velocity u} and \ref{fig:2D Riemann the predicted solutions for velocity v and pressure} show the comparison of numerical solutions for $\rho$, $u$, $v$, and $p$ at $t = 0.2$ under the two models. From the figures, although there are slight differences in complex regions such as shock waves and contact discontinuities, the overall performance of the closure model and the coupling algorithm is highly consistent with the target model, fully demonstrating their reliability and accuracy in capturing complex flow field characteristics. 

\section{Data For N-A Equation of State}
The five datasets used for modeling the N-A equation of state were obtained from the numerical solutions of smooth flow fields under the following five initial condition settings.

\textbf{Case1:}
The problem adopts periodic boundary conditions, with time $ t \in [0, 0.47] $ and spatial variable $ x \in [-1.0, 1.0] $. The initial values of the physical quantities are given as:
\begin{equation}
\left\{
\begin{aligned}
\rho(x,0) &= 0.35+0.25\sin(\pi x),\\
u(x,0)  &= 1.0,\\
p(x,0)&=0.154-0.03\sin(\pi x)-0.1\sin^2(\pi x)
\end{aligned}
\right.
\end{equation}
In the constructed $Net_1$, we used a neural network with $10$ hidden layers, each containing $50$ neurons. In the $T \times X$ space, $8000$ points were randomly sampled as residual points for the RH loss and equation loss. Additionally, we have a total of $300$ conservation points at both $t_1=0$ and $t_2=0.47$. The data included only a small amount of initial and final-state information.

\textbf{Case2:}
The problem adopts periodic boundary conditions, with time $ t \in [0, 0.345] $ and spatial variable $ x \in [-1.0, 1.0] $. The initial values of the physical quantities are given as:
\begin{equation}
\left\{
\begin{aligned}
\rho(x,0) &= 0.75+0.5(1.0-\cos(\pi x)),\\
u(x,0)  &= 0.5\sqrt{1.4}(1.0-\cos(\pi x)),\\
p(x,0)&=0.6+0.4(1.0-\cos(\pi x))
\end{aligned}
\right.
\end{equation}
In the constructed $Net_1$, we used a neural network with $10$ hidden layers, each containing $50$ neurons. In the $T \times X$ space, $8000$ points were randomly sampled as residual points for the RH loss and equation loss. Additionally, we have a total of $300$ conservation points at both $t_1=0$ and $t_2=0.345$. The data included only a small amount of initial and final-state information.

\textbf{Case3:}
The problem adopts periodic boundary conditions, with time $ t \in [0, 0.5] $ and spatial variable $ x \in [-1.0, 1.0] $. The initial values of the physical quantities are given as:
\begin{equation}
\left\{
\begin{aligned}
\rho(x,0) &= 0.65+0.5(1.0-\cos(\pi x)),\\
u(x,0)  &= 0.35\sqrt{1.4}(1.0-\cos(\pi x)),\\
p(x,0)&=0.26+0.2(1.0-\cos(\pi x))
\end{aligned}
\right.
\end{equation}
In the constructed $Net_1$, we used a neural network with $10$ hidden layers, each containing $50$ neurons. In the $T \times X$ space, $8000$ points were randomly sampled as residual points for the RH loss and equation loss. Additionally, we have a total of $300$ conservation points at both $t_1=0$ and $t_2=0.5$. The data included only a small amount of initial and final-state information.

\textbf{Case4:}
The problem adopts periodic boundary conditions, with time $ t \in [0, 0.45] $ and spatial variable $ x \in [-1.0, 1.0] $. The initial values of the physical quantities are given as:
\begin{equation}
\left\{
\begin{aligned}
\rho(x,0) &= 0.7+0.5(1.0-\cos(\pi x)),\\
u(x,0)  &= 0.3\sqrt{1.4}(1.0-\cos(\pi x)),\\
p(x,0)&=0.216-0.09\cos(\pi x)-0.072\sin(\pi x)+0.015\sin(2\pi x)
\end{aligned}
\right.
\end{equation}
In the constructed $Net_1$, we used a neural network with $10$ hidden layers, each containing $50$ neurons. In the $T \times X$ space, $8000$ points were randomly sampled as residual points for the RH loss and equation loss. Additionally, we have a total of $300$ conservation points at both $t_1=0$ and $t_2=0.45$. The data included only a small amount of initial and final-state information.

\textbf{Case5:}
The problem adopts periodic boundary conditions, with time $ t \in [0, 0.3] $ and spatial variable $ x \in [-1.0, 1.0] $. The initial values of the physical quantities are given as:
\begin{equation}
\left\{
\begin{aligned}
\rho(x,0) &= 0.15+0.5(1.0-\cos(\pi x),\\
u(x,0)  &= 0.25\sqrt{1.4}(1.0-\cos(\pi x)),\\
p(x,0)&= 0.126+0.42(1.0-\cos(\pi x))
\end{aligned}
\right.
\end{equation}
In the constructed $Net_1$, we used a neural network with $10$ hidden layers, each containing $50$ neurons. In the $T \times X$ space, $8000$ points were randomly sampled as residual points for the RH loss and equation loss. Additionally, we have a total of $300$ conservation points at both $t_1=0$ and $t_2=0.3$. The data included only a small amount of initial and final-state information.

\bibliographystyle{plainnat}
\bibliography{example}

\end{document}